\documentclass[compsoc]{IEEEtran}

\IEEEoverridecommandlockouts                              % This command is only needed if 
\usepackage[superscript]{cite}
\usepackage{amsmath,amssymb,amsfonts}
\usepackage{mathtools}
\usepackage{algorithmic}
\usepackage{graphicx}
\usepackage{textcomp}
\usepackage{float}
\usepackage[hyphens]{url}
\usepackage{hyperref}
\usepackage[hyphenbreaks]{breakurl}
\usepackage[capitalise]{cleveref}
\usepackage[table,xcdraw]{xcolor}
\usepackage{tikz,pgf} 

\usepackage[autostyle, english = american]{csquotes}  % to fix opening quotation marks
\MakeOuterQuote{"}

\usepackage[caption=false]{subfig}

\usepackage{booktabs}

\usepackage[utf8]{inputenc}
\usepackage{pgfplots}
\usepgfplotslibrary{groupplots,dateplot}
\usetikzlibrary{patterns,shapes.arrows}
\pgfplotsset{compat=newest}

\usetikzlibrary{external}
\usetikzlibrary{calc}

\title{\LARGE \bf
Toward an Idiomatic Framework for Cognitive Robotics}
\author{Malte R. Damgaard*, Rasmus Pedersen, Thomas Bak
\thanks{All authors are with Department of Electronic Systems, Automation and Control, Aalborg University,     Denmark, *Correspondence: {\tt\small  mrd@es.aau.dk}.}%
}

\begin{document}

\maketitle
%\thispagestyle{empty}
%\pagestyle{empty}
% to add page numbers
\thispagestyle{plain}
\pagestyle{plain}
%%%%%%%%%%%%%%%%%%%%%%%%%%%%%%%%%%%%%%%%%%%%%%%%%%%%%%%%%%%%%%%%%%%%%%%%%%%%%%%%
\begin{abstract}
%This paper present an updated version of the "Cognitive Hourglass" model first presented in \cite{RosenbloomDemskiUstun+2017+1+103} based on recent progress in the machine learning community, and is intended as a discussion of how this updated model potentially can influence the the research field of cognitive robotics ...
Inspired by the "Cognitive Hourglass" model presented by the researchers behind the cognitive architecture called Sigma\cite{RosenbloomDemskiUstun+2017+1+103}, we propose a framework for developing cognitive architectures aimed at cognitive robotics. The purpose of the proposed framework is foremost to ease the development of cognitive architectures by encouraging and mitigating cooperation and re-use of existing results. This is done by proposing a framework dividing the development of cognitive architectures into a series of layers that can be considered partly in isolation, and some of which directly relate to other research fields. Finally, we give introductions to and review some topics essential to the proposed framework.
\end{abstract}

\section{Introduction}\label{sec:introduction}
%%%%%%%%%%%%%%%%%%%%%% WHY is the topic of interest%%%%%%%%%%%%%%%%%%%%%% 
% why do we need idiomatic cognitive robotics?
% developing specialized algorithms for specific parts of cognition is inefficient unless it is for a very important part of cognition, and potentially hinders cooperation!

% The research field of cognitive robotics has sprung out of increasing interest for getting robots out of their factory cages to perform and automate daily tasks in less structured and dynamic environments in close vicinity to and possibly even in direct interaction with humans. The increased complexity of the required tasks that should be performed combined with the uncertainty in such environments requires that robots are endowed with the capacity to reason about and plan solutions for complex goals, and to enact those plans while being reactive to unexpected changes in their environment \cite{electronics10070793}.
Research in cognitive robotics originates from a need to perform and automate tasks in dynamic environments and in close or direct interaction with humans. Uncertainty about the environment and complexity of the tasks require robots with the ability to reason and plan while being reactive to changes in their environment. To achieve such behavior, robots cannot rely on predefined rules of behavior \cite{Haazebroek} and inspiration is taken from cognitive architectures.

%%%%%%%%%%%%%%%%%%%%%% WHAT (1) is the background on the previous solutions, if any? %%%%%%%%%%%%%%%%%%%%%% 
Cognitive architectures provide a model for information processing that can capture robot functionalities. In combination with acquired sensory data, they can potentially generate intelligent autonomous behavior \cite{electronics10070793}. Cognitive architectures dates back to the 1950s \cite{40yearsOfCog} with a grand goal of implementing a full working cognitive system \cite{RosenbloomDemskiUstun+2017+1+103}. From this considerable challenge, an abundance of architectures have evolved, and a recent survey suggests that the number of existing architectures has reached several hundred \cite{40yearsOfCog}. Some are aimed towards robotics application, e.g., Robo-Soar \cite{LAIRD1991113}, CARACaS \cite{CARACaS}, and RoboCog \cite{RoboCog}. Unfortunately, most of these architectures take wildly different approaches to model cognition and are implemented in different programming languages. Furthermore, most of these architectures are constructed from a diverse set of specialized modules, e.g., special purpose speech, vision, and simultaneous localization and mapping modules, interacting with each other and have taken decades to develop \cite{RosenbloomDemskiUstun+2017+1+103}. Thereby, making it non-trivial to expand upon, combine, and re-use parts of these architectures. Again, this makes it potentially harder for other researchers and practitioners to contribute to or adopt the architectures for specific problems and applications. In line with the arguments for developing an interface layer for artificial intelligence put forward by other researches\cite{6812892}, we argue that a unifying and standardized framework for developing new cognitive architectures aimed at cognitive robotics could potentially remedy these issues and ease the development of cognitive robotics.

% With roots in these architectures many advanced robotics applications have been demonstrated. \textcolor{red}{ GIVE EXAMPLES ...} 

% Like the aforementioned examples, usually these applications are developed by the development teams behind the used architectures possessing knowledge and architectural insights only obtainable by other researchers within cognitive robotics after considerable efforts. 

% Presumably, due to the effort required to utilize and build on top of existing cognitive architectures much research in cognitive robotics still focus on relatively narrow cognitive capabilities such as \textcolor{red}{ GIVE EXAMPLES ...}.

%%%%%%%%%%%%%%%%%%%%%% WHAT (2) is the background on potential solutions? %%%%%%%%%%%%%%%%%%%%%%
In recent years a community consensus has begun to emerge about a standard model of humanlike minds, i.e., computational entities whose structures and processes are substantially similar to those found in human cognition \cite{Laird_Lebiere_Rosenbloom_2017}. While this "Standard Model of the Mind" spans key aspects of structure and processing, memory and content, learning, and perception and motor, it is agnostic to the best practice for modeling and implementing these things \cite{Laird_Lebiere_Rosenbloom_2017}. 
In line with the idea proposed by the researchers behind the cognitive architecture Sigma \cite{RosenbloomDemskiUstun+2017+1+103}, we argue that the evolution of the scientific field of cognitive robotics could benefit from anchoring new implementations around a common theoretical elegant base separating a specific model of a part of cognition from the algorithm that implements it. Furthermore, this theoretical base could allow new functionalities to evolve hierarchically just like software libraries build on top of each other. Thereby, allowing the discussions and development to flourish at different levels of abstractions, and allow for direct synergy with other research fields. % such as deep learning, which have had considerable progress in recent years.

% In [3] the authors presents a Cognitive Hourglass model with the so-called "graphical architecture" at its waist as such a potential theoretical elegant base. 
%In \cite{RosenbloomDemskiUstun+2017+1+103} the authors present a Cognitive Hourglass model for their cognitive architecture called "Sigma" based on the following four desiderata:
To explain the cognitive architecture Sigma \cite{RosenbloomDemskiUstun+2017+1+103}, the authors present a Cognitive Hourglass model based on the following four desiderata:

\begin{itemize}
    \item Grand Unification, spanning all of cognition,
    \item Generic Cognition, spanning both natural and artificial cognition,
    \item Functional Elegance, achieving generically cognitive grand unification with simplicity and theoretical elegance,
    \item Sufficient Efficiency, efficient enough to support the anticipated uses in real-time.
\end{itemize}

While Grand Unification and Sufficient Efficiency aligns well with the needs of cognitive robotics, the need for Generic Cognition and Functional Elegance is subtle for cognitive robotics. Although the end goal of cognitive robotics might only be to develop functional artificial intelligence, building %this artificial intelligence 
upon something that potentially is also able to model natural intelligence would allow artificial intelligence to more easily benefit from insights obtained by the modeling of natural intelligence and vice versa. Similarly, Functional Elegance is not a goal of cognitive robotics per se. Still, it could allow researchers and practitioners working on different levels of cognition to obtain a common reference point and understanding at a basic level, potentially easing co-operation and re-use of results and innovative ideas. 

In an attempt to obtain all four of these desiderata %the authors of \cite{RosenbloomDemskiUstun+2017+1+103} place the so-called "graphical architecture" at the waist of their Cognitive Hourglass model as such a potential theoretical elegant base glueing everything together just like the Internet Protocol (IP) in the Internet-hourglass model \cite{ipHourglass}. 
the so-called "graphical architecture" is placed at the waist of Sigma's Cognitive Hourglass model as such a potential theoretical elegant base glueing everything together just like the Internet Protocol (IP) in the Internet-hourglass model \cite{ipHourglass}. 
Functional elegance is obtained by recognizing and developing general architectural fragments, and based on these defining idioms that can be re-used in modelling different parts of cognition. Having defined sufficiently general idioms, the hope is to be able to develop full models of cognition from a limited set of such idioms and thereby achieve functional elegance, while at the same time achieving the three other desiderata \cite{RosenbloomDemskiUstun+2017+1+103}. 

%%%%%%%%%%%%%%%%%%%%%% WHAT (3) was attempted in the present effort? %%%%%%%%%%%%%%%%%%%%%% 
With roots in the given desiderata, Sigma's Cognitive Hourglass model in many ways could constitute a unifying and standardized framework for cognitive robotics. However, as we will elaborate on in \cref{sec:sigma_model} the model commits to specific architectural decisions, which hinders the utilization of new technology and ideas. E.g., their commitment to the sum-product algorithm prevents the use of new algorithms for efficient probabilistic inference. %Therefore, in this paper, we propose a new generalized Cognitive Hourglass model based on recent advancements within the machine learning community that makes no such commitments.
Therefore, based on the observation that the layers of Sigma's Cognitive Hourglass model conceptually can be divided into more and generalized layers, in this paper, we propose a new generalized Cognitive Hourglass model based on recent advancements within the machine learning community that makes no such commitments.
The presented Generalized Cognitive Hourglass model constitutes a new framework for guiding and discussing the future development of cognitive robotics. As such, with this paper we do not intend to construct a new specific cognitive architecture our-self. Instead, our framework should be viewed as a space of systems, and our intend with this framework and manuscript is 
\begin{enumerate}
    \item to provide a framework for other researchers to expand upon,
    \item to ease the development of cognitive architectures for robotics by encouraging and mitigating cooperation and re-use of existing results,
    %\item to provide researchers in cognitive architectures and cognitive robotics with a broader framework
    %\item to make other researchers in the field of cognitive architectures and cognitive robotics aware about how their own work fits into a broader picture, 
    %\item secondly to inspire new work to be analysed and streamlined in the light of this framework,
    \item and finally to highlight some of the current state-of-the-art technology available to progress this research field.
\end{enumerate}
%By doing so we hope to mitigate better cooperation and re-use of results in the community.

%%%%%%%%%%%%%%%%%%%%%% WHAT (4) will be presented in this paper? %%%%%%%%%%%%%%%%%%%%%% 
In \cref{sec:sigma_model} we briefly introduce the Sigma's Cognitive Hourglass model in more detail. Based on this, we present our Generalized Cognitive Hourglass model as a framework for developing new cognitive architectures aimed at cognitive robotics in \cref{sec:our_model}. In \cref{sec:probabilistic_programs} we give a brief introduction to probabilistic programs since the presented framework is built around them. Explaining the functionality of probabilistic programs with conventional methods can be difficult, therefore in \cref{sec:gfg} we present a new graphical representation of probabilistic programs which we call "generative flow graphs". We do so in the hope that it will ease the dissemination of new models of parts of cognition developed within the proposed framework. Being fundamental to achieving functional elegance within the proposed framework, we formally introduce the concept of probabilistic programming idioms in \cref{sec:ppl_idioms} and explain how "generative flow graphs" can aid the identification of such idioms. In \cref{sec:inference} we discuss the intrinsic problem of performing approximate inference in complex probabilistic programs and present some modern algorithms to tackle this problem for cognitive robotics. As probabilistic programming languages form the foundation of the present framework, we give a brief survey of probabilistic programming languages relevant to the framework in \cref{sec:ppl}. Finally, in \cref{sec:our_examples} we shortly present some preliminary work to support the framework presented.

% \label{sec:our_examples}
% Sigma model
% Our framework
% Probabilistic programs
%% Generative flow graphs

%% Probabilistic Programming Idioms \cref{sec:ppl_idioms} 
% INFERENCE ALGORITHMS \cref{sec:inference}
%% VI \cref{sec:VI}
%% SVI \cref{sec:SVI}
%% MSG-Pas \cref{sec:msg_passing}
%% Stochastic msg-pas \cref{sec:SMP}
% Probabilistic programming languages \cref{sec:ppl}

% In the next section, we present a model with a structure very similar to the advantageous structure of \cref{fig:sigma_model} albeit without such restrictive commitments.
% For this reason, the reminding sections are devoted to briefly introducing and reviewing some important topics related to the layers of our framework.
% For this reason, we will in the following section present what we have chosen to call Generative Flow Graphs; a representation that also helps us identify possible graphical idioms.
% In the next section we will discuss how generative flow graphs can be used as a tool for discovering probabilistic programming idioms.
% For a given probabilistic program to be useful, we want to answer queries about specific variables. To do so we need to be able to perform inference over these probabilistic programs, which is the topic of the next section.

\section{Sigma's Cognitive Hourglass Model}\label{sec:sigma_model}
% write about efficiency

%\begin{figure*}
%    \centering
%    \subfloat[Loose re-drawing of figures of the cognitive Hourglass model presented in \cite{RosenbloomDemskiUstun+2017+1+103}. Layers with dashed borders are not recognized as distinguishable layers by the authors of \cite{RosenbloomDemskiUstun+2017+1+103}. \label{fig:sigma_model}]{
%        \resizebox{0.45\linewidth}{!}{
%            \input{tikz/sigma_model.tikz}
%        }
%    } 
%    \subfloat[Dashed borders indicates layers that are not strictly necessary, but helps.\label{fig:overall_idea}]{
%        \resizebox{0.525\linewidth}{!}{
%            \input{tikz/overall_idea.tikz}
%        }
%    }
%    \caption{}
%    %\label{fig:slam_graph_idiom_new}
%\end{figure*}

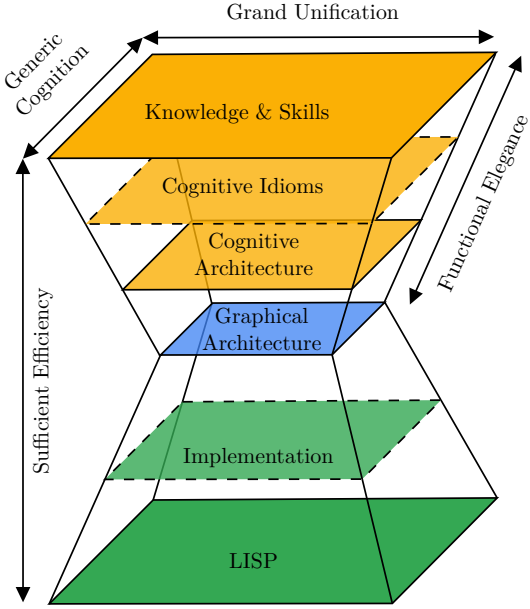
\begin{figure}[!t]
\centering
\resizebox{0.925\linewidth}{!}{
    \tikzset{every picture/.style={line width=0.75pt}} %set default line width to 0.75pt        

\begin{tikzpicture}[x=0.75pt,y=0.75pt,yscale=-1,xscale=1]
%uncomment if require: \path (0,394); %set diagram left start at 0, and has height of 394

%Straight Lines [id:da48214511493186607] 
\draw    (132.46,186.63) -- (115.4,243.81) ;
%Straight Lines [id:da7587216853874668] 
\draw    (115.4,243.81) -- (98.33,301) ;
%Shape: Polygon [id:ds17701567924960182] 
\draw  [fill={rgb, 255:red, 52; green, 168; blue, 83 }  ,fill opacity=1 ] (98.33,301) -- (298.5,299.8) -- (237.5,361.25) -- (135.03,361.38) -- (37.83,361.5) -- cycle ;
%Shape: Polygon [id:ds42099163838915454] 
\draw  [fill={rgb, 255:red, 52; green, 168; blue, 83 }  ,fill opacity=0.8 ][dash pattern={on 4.5pt off 4.5pt}] (115.4,243.81) -- (265.81,242.84) -- (219.98,288.84) -- (70.23,289.03) -- cycle ;
%Straight Lines [id:da7374997790736944] 
\draw    (97.83,41.5) -- (109.38,89.88) ;
%Straight Lines [id:da8768124193930502] 
\draw    (109.38,89.88) -- (120.92,138.25) ;
%Straight Lines [id:da5894715536228128] 
\draw    (37.33,102) -- (58.96,140.29) ;
%Straight Lines [id:da5240411687244109] 
\draw    (58.96,140.29) -- (80.58,178.58) ;
%Straight Lines [id:da5062019309669883] 
\draw    (95.25,31.74) -- (293.75,31.26) ;
\draw [shift={(296.75,31.25)}, rotate = 539.86] [fill={rgb, 255:red, 0; green, 0; blue, 0 }  ][line width=0.08]  [draw opacity=0] (8.93,-4.29) -- (0,0) -- (8.93,4.29) -- cycle    ;
\draw [shift={(92.25,31.75)}, rotate = 359.86] [fill={rgb, 255:red, 0; green, 0; blue, 0 }  ][line width=0.08]  [draw opacity=0] (8.93,-4.29) -- (0,0) -- (8.93,4.29) -- cycle    ;
%Straight Lines [id:da411216310670226] 
\draw    (90.14,33.88) -- (24.61,100.12) ;
\draw [shift={(22.5,102.25)}, rotate = 314.69] [fill={rgb, 255:red, 0; green, 0; blue, 0 }  ][line width=0.08]  [draw opacity=0] (8.93,-4.29) -- (0,0) -- (8.93,4.29) -- cycle    ;
\draw [shift={(92.25,31.75)}, rotate = 134.69] [fill={rgb, 255:red, 0; green, 0; blue, 0 }  ][line width=0.08]  [draw opacity=0] (8.93,-4.29) -- (0,0) -- (8.93,4.29) -- cycle    ;
%Shape: Polygon [id:ds3050324501344104] 
\draw  [fill={rgb, 255:red, 66; green, 133; blue, 244 }  ,fill opacity=0.77 ] (132.88,186.31) -- (233.13,185.88) -- (202.46,216.44) -- (102.63,216.56) -- cycle ;
%Straight Lines [id:da3384253317217001] 
\draw    (297.5,41.25) -- (275.9,89.56) ;
%Straight Lines [id:da39018791696940314] 
\draw    (275.9,89.56) -- (254.31,137.87) ;
%Straight Lines [id:da6225276651300811] 
\draw    (311.28,44.99) -- (248.93,184.45) ;
\draw [shift={(247.71,187.19)}, rotate = 294.09000000000003] [fill={rgb, 255:red, 0; green, 0; blue, 0 }  ][line width=0.08]  [draw opacity=0] (8.93,-4.29) -- (0,0) -- (8.93,4.29) -- cycle    ;
\draw [shift={(312.5,42.25)}, rotate = 114.09] [fill={rgb, 255:red, 0; green, 0; blue, 0 }  ][line width=0.08]  [draw opacity=0] (8.93,-4.29) -- (0,0) -- (8.93,4.29) -- cycle    ;
%Straight Lines [id:da30838703999159756] 
\draw    (213.97,178.21) -- (202.46,216.44) ;
%Straight Lines [id:da775992946862964] 
\draw    (120.92,138.25) -- (132.46,186.63) ;
%Straight Lines [id:da09956950748197024] 
\draw    (80.58,178.58) -- (102.21,216.88) ;
%Straight Lines [id:da3191248841619554] 
\draw    (254.31,137.87) -- (232.71,186.19) ;
%Shape: Polygon [id:ds9964719674747267] 
\draw  [fill={rgb, 255:red, 251; green, 176; blue, 5 }  ,fill opacity=0.8 ] (120.92,138.25) -- (254.31,137.87) -- (213.97,178.21) -- (80.58,178.58) -- cycle ;
%Straight Lines [id:da3402262896773103] 
\draw    (225.49,139.98) -- (213.97,178.21) ;
%Shape: Polygon [id:ds18327409351060098] 
\draw  [fill={rgb, 255:red, 251; green, 176; blue, 5 }  ,fill opacity=0.8 ][dash pattern={on 4.5pt off 4.5pt}] (109.38,89.88) -- (275.9,89.56) -- (225.49,139.98) -- (58.96,140.29) -- cycle ;
%Straight Lines [id:da6073975425902616] 
\draw    (237,101.75) -- (225.49,139.98) ;
%Shape: Polygon [id:ds42309497010640174] 
\draw  [fill={rgb, 255:red, 251; green, 176; blue, 5 }  ,fill opacity=1 ] (97.83,41.5) -- (298,40.3) -- (237,101.75) -- (37.33,102) -- cycle ;
%Straight Lines [id:da27597256435756745] 
\draw    (22.5,105.25) -- (22.6,357.6) ;
\draw [shift={(22.6,360.6)}, rotate = 269.98] [fill={rgb, 255:red, 0; green, 0; blue, 0 }  ][line width=0.08]  [draw opacity=0] (8.93,-4.29) -- (0,0) -- (8.93,4.29) -- cycle    ;
\draw [shift={(22.5,102.25)}, rotate = 89.98] [fill={rgb, 255:red, 0; green, 0; blue, 0 }  ][line width=0.08]  [draw opacity=0] (8.93,-4.29) -- (0,0) -- (8.93,4.29) -- cycle    ;
%Straight Lines [id:da11430097573251707] 
\draw    (102.63,216.56) -- (70.23,289.03) ;
%Straight Lines [id:da7642566934077504] 
\draw    (70.23,289.03) -- (37.83,361.5) ;
%Straight Lines [id:da26523788901371326] 
\draw    (233.13,185.88) -- (265.81,242.84) ;
%Straight Lines [id:da8326787953859103] 
\draw    (265.81,242.84) -- (298.5,299.8) ;
%Straight Lines [id:da22645008459378868] 
\draw    (202.46,216.44) -- (219.98,288.84) ;
%Straight Lines [id:da5810586445385235] 
\draw    (219.98,288.84) -- (237.5,361.25) ;

% Text Node
\draw  [draw opacity=0]  (21.25,161) -- (47.25,161) -- (47.25,302) -- (21.25,302) -- cycle  ;
\draw (34.25,231.5) node  [rotate=-270] [align=left] {\begin{minipage}[lt]{93.08pt}\setlength\topsep{0pt}
\begin{center}
Sufficient Efficiency
\end{center}

\end{minipage}};
% Text Node
\draw (34.5,51.5) node  [rotate=-315.03] [align=left] {\begin{minipage}[lt]{68pt}\setlength\topsep{0pt}
\begin{center}
Generic Cognition
\end{center}

\end{minipage}};
% Text Node
\draw (196.75,16.5) node   [align=left] {\begin{minipage}[lt]{83.64pt}\setlength\topsep{0pt}
Grand Unification
\end{minipage}};
% Text Node
\draw (295.13,119.5) node  [rotate=-294.16] [align=left] {\begin{minipage}[lt]{97.75pt}\setlength\topsep{0pt}
Functional Elegance
\end{minipage}};
% Text Node
\draw (162.9,119.25) node   [align=left] {\begin{minipage}[lt]{89.35pt}\setlength\topsep{0pt}
Cognitive Idioms
\end{minipage}};
% Text Node
\draw (162.45,76.15) node   [align=left] {\begin{minipage}[lt]{103.36pt}\setlength\topsep{0pt}
Knowledge \& Skills
\end{minipage}};
% Text Node
\draw (156.9,157.68) node   [align=left] {\begin{minipage}[lt]{89.35pt}\setlength\topsep{0pt}
\begin{center}
Cognitive\\Architecture
\end{center}

\end{minipage}};
% Text Node
\draw (161.57,201.25) node   [align=left] {\begin{minipage}[lt]{89.31pt}\setlength\topsep{0pt}
\begin{center}
Graphical\\Architecture
\end{center}

\end{minipage}};
% Text Node
\draw (159.57,276.05) node   [align=left] {\begin{minipage}[lt]{89.31pt}\setlength\topsep{0pt}
\begin{center}
Implementation
\end{center}

\end{minipage}};
% Text Node
\draw (156.77,336.05) node   [align=left] {\begin{minipage}[lt]{89.31pt}\setlength\topsep{0pt}
\begin{center}
LISP
\end{center}

\end{minipage}};

\end{tikzpicture}
}
\caption{Loose re-drawing of figures of the cognitive hourglass model presented in \cite{RosenbloomDemskiUstun+2017+1+103}. Layers with dashed borders are not recognized as distinguishable layers by the authors of \cite{RosenbloomDemskiUstun+2017+1+103}. }\label{fig:sigma_model}
\end{figure}

% As stated, the authors of \cite{RosenbloomDemskiUstun+2017+1+103} base their Cognitive Hourglass Model on four desiderata. 
\cref{fig:sigma_model} illustrates how the dimensions of the model Sigma's Cognitive Hourglass Model relate to the four desiderata. The top layer of the hourglass represents all the knowledge and skills implemented by the cognitive system. This includes high level cognitive capabilities such as reasoning, decision making, and meta cognition, as well as low level cognitive capabilities such as perception, attention, and the formation of knowledge and memory that could potentially be inspired by human cognition. But it also includes artificial cognitive capabilities such as, e.g., the creation of a grid-maps common in robotics. As such, the extend of this layer corresponds to the achievable extend of \textit{Grand Unification} and \textit{Generic Cognition}. % The "cognitive architecture" layer conceptually provides a language in which all of the knowledge and skills in the top layer can be embodied and learned. As an intermediate layer, cognitive idioms provide design patterns, libraries, and services that ease the implementation of knowledge and skills. Besides providing a language for high-level knowledge and skills, the cognitive architecture also includes special architectural constructs such as the "cognitive cycle" and a tri-level control structure. 

The "cognitive architecture" layer defines central architectural decisions such as the utilization of the cognitive cycle and tri-level control structure for information processing, and the division of memory into a perceptual buffer, working memory and long-term memory. But also defines other architectural concepts such as "functions", "structures", "affect/emotion", "surprise", and "attention". Thereby, the cognitive architecture induces what can be considered an "cognitive programming language" in which all of the knowledge and skills in the top layer can be embodied and learned. As an intermediate layer, cognitive idioms provide design patterns, libraries, and services that ease the implementation of knowledge and skills. 

Below the Cognitive architecture and at the waist of the model, they have the graphical architecture constituting a small elegant core of functionality. \textit{Functional elegance} is thereby obtained by compilation of knowledge and skills through a series of layers into a common representation in the graphical architecture. This graphical architecture primarily consists of probabilistic inference over graphical models, more specifically factor graphs, utilizing the sum-product algorithm \cite{910572} plus the following extensions:
\begin{enumerate}
    \item each variable node is allowed to correspond to one or more function variables,
    \item special purpose factor nodes,
    \item and the possibility of limiting the direction of influence along a link in the graph.
\end{enumerate}
Of these extensions, the two first are merely special-purpose optimizations for the inference algorithm, i.e., a part of the implementation layer in \cref{fig:sigma_model}. According to the authors the third extension has "a less clear status concerning factor graph semantics"\cite{RosenbloomDemskiUstun+2017+1+103}. Finally, the graphical architecture is implemented in the programming language LISP. In this model, \textit{sufficient efficiency} is achieved as the cumulative efficiency of all layers. I.e., an efficient implementation in LISP is futile if models of knowledge and skills are inefficient for a given task.

To summarize, the model shown in \cref{fig:sigma_model} commits to multiple more or less restrictive decisions such as the utilization of factor graphs and the sum-product algorithm at its core, the "cognitive cycle", the tri-level control structure, and LISP as the exclusive implementation language. While these commitments may be suitable for the specific cognitive architecture Sigma mainly targeted human-like intelligens, they would hinder the exploration of new ideas and the utilization of new technologies, making this model less suitable as a general framework. % In the next section, we present a model with a structure very similar to the advantageous structure of \cref{fig:sigma_model} albeit without such restrictive commitments.

%\section{Results}
\section{Generalized Cognitive Hourglass Model}\label{sec:our_model}

%\begin{figure}[tb]
%\centering
%\resizebox{1.0\linewidth}{!}{
%    \input{tikz/overall_idea.tikz}
%}
%\caption{Dashed borders indicates layers that are not strictly necessary, but helps. }
%\label{fig:overall_idea}
%\end{figure}

While Sigma's Cognitive Hourglass model has an advantageous structure with roots in highly appropriate desiderata, it is not suitable as a general framework, due to some exclusive structural commitments. %as expatiated on in \cref{sec:sigma_model}. 
We argue that these structural commitments are mostly artefacts of the limited expressibility of factor graphs and the sum-product algorithm. % at the core of the model.

Consider, for instance, the "cognitive cycle" dividing processing into an elaboration and adaption phase. The elaboration phase performs inference over the factor graph, while the adaption phase modifies the factor graph before further inference. We argue that the need for this two-phase division of processing is caused by the need for the sum-product algorithm to operate on a static factor graph. This cognitive cycle further makes the tri-level control structure necessary to make cognitive branching and recursion possible. 
Similarly, we argue that the third extension of the factor graph semantics employed in the cognitive architecture Sigma is nothing more than a simple control flow construct over the information flow in the graphical model and inference algorithm. It is straightforward to imagine how other control flow constructs such as recursion, loops, and conditionals could also be advantageous in modeling cognition.

Basically, we believe that special-purpose implementations of architectural constructs such as the two-phase "cognitive cycle" employed by Sigma and similar cognitive architectures have previously been necessary due to the limitations of the available modeling tools. The flexibility of probabilistic programs provided by the possibility of incorporating I/O operations, loops, branching, and recursion into a probabilistic model, should permit representing such constructs as either Probabilistic Programming or Cognitive Idioms instead. 

\begin{figure}[t]
\centering
\resizebox{1.1\linewidth}{!}{
    \input{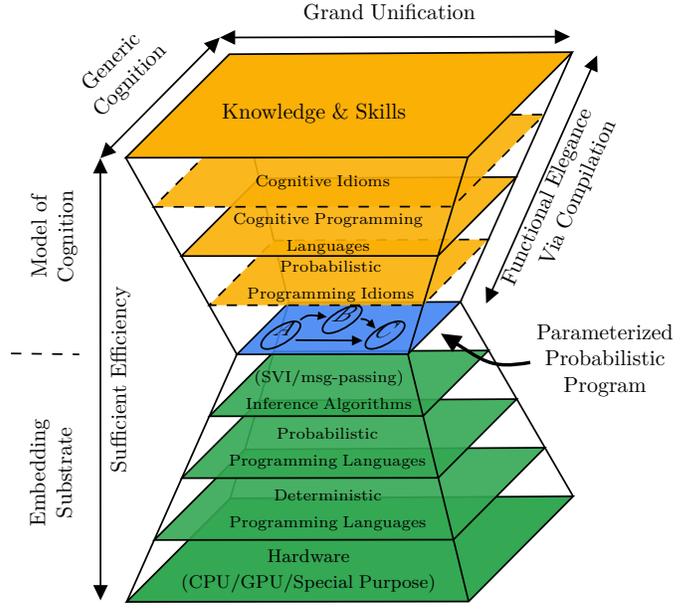}
}
\caption{Our proposal for a generalized cognitive hourglass model. Dashed borders indicates layers that are not necessarily recognized as distinguishable layers, but could help in the development of the other layers. }
\label{fig:overall_idea}
\end{figure}

In \cref{fig:overall_idea} we present our proposal for a more general cognitive hourglass model having probabilistic programs at its waist as the theoretical modeling base.
Just like the model in \cref{sec:sigma_model}, our model is composed of a series of layers that expands away from the waist of the hourglass. On top of the pure probabilistic program, we might be able to recognize program fragments that are sufficiently general to be considered idioms. From these idioms, it might be possible to construct dedicated programming languages for expressing cognitive behavior, knowledge, and skills, such as the "Cognitive Language" employed in the cognitive architecture Sigma\cite{RosenbloomDemskiUstun+2017+1+103}. In this framework, functional elegance above the probabilistic program is obtained via compilation of knowledge and skills through appropriate cognitive programming languages into probabilistic programs.
Below the probabilistic program's different inference algorithms can carry out the necessary inference in the probabilistic program. Different versions of these inference algorithms can potentially be implemented in different probabilistic languages. Both the probabilistic program and probabilistic programming language can be situated in standard deterministic programming languages. Furthermore, one needs not even use the same deterministic programming language for both\cite{vandemeent2018introduction}. Finally, the deterministic programming languages allow us to execute a model of cognition on different types of hardware doing the actual computations. When comparing Sigma's Hourglass model to the Generalized Hourglass model the complexity might seem to have increased. However, this is not the case. The Generalized Hourglass model simply highlights some of the components implicit in Sigma's Hourglass model.

We expect that this model is sufficiently general to be considered a framework for research in, and development of, cognitive robotics. In fact, as probabilistic programs can be considered an extension of deterministic programs, it should even be possible to situate both emergent, symbolic, and hybrid approaches to cognitive architectures in this framework, thereby covering the full taxonomy considered in\cite{40yearsOfCog}. In our framework constructs such as the cognitive cycle and tri-level control structure could potentially be expressed as probabilistic programming idioms rather than special-purpose architectural implementations. Similarly, incorporation of results from other research areas such as deep learning is only limited to the extent that a given probabilistic programming language and corresponding inference algorithms can incorporate essential tools used in these research areas, i.e., automatic differentiation for deep learning. Furthermore, this framework gives a satisfying view on the foundational hypothesis in artificial intelligence about substrate independence \cite{Laird_Lebiere_Rosenbloom_2017}, by cleanly separating the model of cognition, i.e. the probabilistic program and everything above it, from the organic or inorganic substrate that it exists on, i.e., everything below the probabilistic program.

While the above might sound promising, the choice of probabilistic programs as a focal point also has important ramifications. In general, we cannot guarantee the existence of an analytic solution for all models, and even if a solution exists it might be computationally intractable \cite{vandemeent2018introduction}. Therefore, we have to endure approximate solutions. Though this might sound restrictive, this is also the case for most other complex real-world problems. In fact, it can be considered a form of bounded rationality consistent with the concept of "satisficing" stating that an organism confronted with multiple goals does not have the senses nor the wits to infer an "optimal" or perfect solution, and thus will settle for the first solution permitting satisfaction at some specified level of all of its needs \cite{SimonH.A1956Rcat}. The second important ramification is that the model with its roots in probabilistic and deterministic programming languages is only applicable to the extent to which the hypothesis that artificial cognition can be grounded in such programming languages is valid. However, this is currently a widely accepted hypothesis.

It is important to stress that the layers of the proposed framework are not independent. On the contrary, as the technological possibilities and community knowledge evolves, changes in one layer might open new possibilities in the layers above. Similarly, the need for new features in one layer might guide the research directions and development of the layers below. However, this structure is exactly what would allow further discussions and development in cognitive robotics to evolve at different levels of abstractions, and benefit from other research fields related to the layers below probabilistic programs. In the layers above probabilistic programs, the development and identification of both probabilistic programming idioms, cognitive programming languages, and cognitive idioms mitigate cooperation and re-use of existing results. The framework thus minimizes the burden of developing new cognitive architectures by allowing researchers to focus their energy on specific layers, or parts thereof, in the hourglass model rather than dealing with all the details of a cognitive architecture. The extent to which the burden of development is reduced thus depends upon the technology available in each of the layers of the hourglass. 

% For this reason, the reminding sections are devoted to briefly introducing and reviewing some important topics related to the layers of our framework.

% We know that given sufficient time some inference algorithms will find an exact solution ... i.e. we can simulate and test if a model of cognition is good even though we might not be able to execute it is real-time

%\subsection{Related Work}\label{sec:RW}
Even though our generalized cognitive hourglass model has roots in the ideas behind the cognitive architecture Sigma, other evidence for the appropriability of the framework for cognitive or developmental robotics can also be found in the literature. In recent years, the power of utilizing probabilistic models for cognitive robotics has been demonstrated in a lot of studies. For example, multi-layered multimodal latent Dirichlet allocation (mMLDA) models have been used for learning and representing the hierarchical structure of concepts\cite{6696672}. A complex probabilistic graphical model and an online inference algorithm for simultaneous lexical and spatial concept acquisition have also been demonstrated \cite{Taniguchi2020}. Based on this model, another probabilistic graphical model for navigation utilizing the learned concepts has also been proposed\cite{doi:10.1080/01691864.2020.1817777}. The interaction between multiple probabilistic graphical models, mainly mMLDA, integrated as a cognitive architecture for robot learning of action and language has also been studied\cite{10.3389/frobt.2019.00131}. Related to this research two frameworks called SERKET\cite{10.3389/fnbot.2018.00025} and its extension Neuro-SERKET\cite{10.1007/s00354-019-00084-w} has been proposed with the goal of connecting multiple probabilistic graphical models on a large scale to construct cognitive architectures for robotics. Thereby, both of these frameworks indirectly acknowledge the advantages of constructing large-scale cognitive models by composing them from smaller fundamental entities. In both frameworks, these fundamental entities are called "modules". In SERKET these "modules" are limited to fragments of hierarchical probabilistic graphical models with head-to-tail connections. Neuro-SERKET extends the expressibility of SERKET by allowing "modules" to be general probabilistic generative models with special attention to deep generative models. Connections between these "modules" are allowed in both a head-to-tail and head-to-head fashion, but also in a tail-to-tail fashion via a product of expert approximation of such connections. In both frameworks exact message-passing is used to perform inference on models with discrete and finite variables, otherwise sampling importance resampling is used. As such both of these frameworks can be considered special cases of our framework, with "modules" and their connections somehow resembling what we have chosen to call "probabilistic programming idioms". The difference is that "modules" in SERKET and Neuro-SERKET are supposed to be fully defined and self-contained, whereas our definition of "probabilistic programming idioms" allows for nesting and e.g. class definitions with abstract methods as we will exemplify in \cref{sec:our_examples}. Furthermore, being based solely on probabilistic graphical models, SERKET and Neuro-SERKET currently do not seem to incorporate logic into their "modules". %Some of the most influential work arguing that a combination of logic, especially first-order, and pure probabilistic graphical models are necessary to implement a sufficient interface between artificial intelligence/models of cognition and the embedding substrate is perhaps the work related to Markov Logic and the system called Alchemy \cite{6812892}. 
Some of the most influential work arguing that a combination of logic, especially first-order, and pure probabilistic graphical models is necessary to compose a sufficiently general interface layer between artificial intelligence and the algorithms that implements it, i.e. the waist of the presented framework, is perhaps the work related to Markov Logic and the system called Alchemy \cite{6812892}. Like SERKET and Neuro-SERKET, Alchemy may also be considered one instance of our framework, limiting the probabilistic programs at the waist to Markov logic, and utilizing a combination of Markov chain Monte Carlo and lifted belief propagation for inference \cite{6812892}.

\section{Preliminaries}\label{sec:Preliminaries}
In this paper, we do not distinguish between probability density functions and probability mass functions, and jointly denote them as probability functions. The symbol $\int $ is used to denote both integrals and summations depending on the context. In general, we use $z$ to denote latent random variables, $x$ to denote observed random variables, $p(...)$ to denote "true" probability functions, $q(...)$ to denote approximations to "true" probability functions, $\theta$ to denote parameters of "true" probability functions, $p(...)$, and $\phi$ to denote parameters of approximations to "true" probability functions, $q(...)$. When a probability function directly depends on a parameter we write the parameter in a subscript before the parenteses, e.g., $p_{\theta}(...)$ and $q_{\phi}(...)$. We use line over a value, parameter, or random variable to denote that it is equal to a specific value, e.g., $\overline{z} = 1,432$. We use a breve over a parameter or random variable to denote that it should be considered a fixed parameter or random variable within that equation, e.g., $\breve{\theta}$ or $\breve{z}$. For parameters this means that they attain a specific value, $\overline{\theta}$, i.e., $\breve{\theta}$ means that $\theta = \overline{\theta}$. For random variables it means that the probability functions that this variable is associated with is considered fixed within a given equation. We use capital letters to denote sets, e.g., $A=\left\{1,...,\overline{n}\right\}$. We use a superscript with curly brackets to denote indexes. E.g. $z^{\left\{i\right\}}$ would denote the i'th latent random variable. Similarly, we use a superscript with curly brackets and two numbers separated by a semicolon to denote a set of indexes values, i.e., $z^{\left\{1;\overline{n}\right\}} = z^{\left\{A\right\}} = \left\{ z^{\left\{1\right\}},..., z^{\left\{\overline{n}\right\}}\right\}$. We use a backslash, $\setminus$, after a set followed by a value, random variable, or parameter to denote the exclusion of that value, random variable, or parameter from that set, i.e., $z^{\left\{A\right\}}\setminus z^{\left\{\overline{n}\right\}} = \left\{ z^{\left\{1\right\}},..., z^{\left\{\overline{n}-1\right\}}\right\}$. We use capital $C$ to denote a collection of latent random variables, observed random variables, and parameters. Furthermore, we will specify such a collection by enclosing variables and parameters with curly brackets around and with a semi-colon separating latent random variables, observed random variables, and parameters in that order, e.g., $C=\left\{Z;X;\Theta\right\}$. We will use $\text{Pa}$, $\text{Ch}$, $\text{An}$, and $\text{De}$ as abbreviations for parent, child, ancestors, and descendants, respectively, and use, e.g., $\text{Pa}\Theta(C)$ to denote the set of parameters parent to the collection $C$, and $\text{Ch}X(Z)$ to denote the set of observed variables that are children of the latent random variable $Z$.

% $\text{Pa}$ parent i.e., $\text{Pa}\theta(C)$, $\text{Pa}Z(C)$, and $\text{Pa}X(C)$\\
% $\text{Ch}$ child i.e., $\text{Ch}\theta(C)$, $\text{Ch}Z(C)$, and $\text{Ch}X(C)$\\
% $\text{An}$ ancestors i.e., $\text{An}\theta(C)$, $\text{An}Z(C)$, and $\text{An}X(C)$\\
% $\text{De}$ descendents i.e., $\text{De}\theta(C)$, $\text{De}Z(C)$, and $\text{De}X(C)$\\

\section{Probabilistic Programs}\label{sec:probabilistic_programs}
% we need to be able to represent Control flow constructs in our models --> Current Graphical models does not support control flow constructs -> Generative Control Flow Graphs
% Comment on the extreme cases: Pure logic in one end, vs end-to-end learning in the other, with most cognitive architectures somewhere in between. symbolic vs sub-symbolic vs emergent
% "Three major paradigms are currently recognized: symbolic (also referred to as cognitivist), emergent (connectionist) and hybrid." [40 years of ...]
% "Symbolic systems represent concepts using symbols that can be manipulated using a predefined instruction set. Such instructions can be implemented as if-then rules applied to the symbols representing the facts known about the world"
% "The emergent approach resolves the adaptability and learning issues by building massively parallel models, analogous to neural networks, where information flow is represented by a propagation of signals from the input nodes."

At the heart of our framework, we have chosen to place probabilistic programs. % To fully appreciate this choice we have to get acquainted with the expressibility that they possess. 
One definition of probabilistic programs is as follows: \\

\textit{“Probabilistic programs are usual functional or imperative programs with two added constructs: (1) the ability to draw values at random from distributions, and (2) the ability to condition values of variables in a program via observations.”} \cite{10.1145/2593882.2593900}\\

With these two constructs, any functional or imperative program can be turned into a simultaneous representation of a joint distribution, $p_{\Theta}(Z,X)$, and conditional distribution, $p_{\Theta}(X|Z)$, where $X$ represent the conditioned/observed random variables, $Z$ the unconditioned/latent random variables, and $\Theta$ represents other parameters in the program that are not given a probabilistic treatment. Thereby allowing us to integrate classical control constructs familiar to any programmer such as if/else statement, loops, and recursions into probabilistic models. As such probabilistic programs can express exactly the same functionality as any deterministic programs can and even more. These two constructs are usually provided as extensions to a given programming language through special \textit{sample} and \textit{observe} functions or keywords\cite{vandemeent2018introduction}. Thus it would be natural to represent such probabilistic programs by pseudo-code. However, based on experiences it can be hard to follow the generative flow of random variables in such pseudo-code. Alternatively, such generative flows have classically been represented by directed graphical models \cite{10.5555/1795555}. Unfortunately, we have also found that the semantics of classical directed graphical models neither provide an appropriate presentation. % For this reason, we will in the following section present what we have chosen to call Generative Flow Graphs; a representation that also helps us identify possible graphical idioms.

\subsection{Generative Flow Graphs}\label{sec:gfg}
% An generative selection implies that only one sample of the potentially is used. Mathematically, we can represent a generative selection by a case function
% Comment on why Influence links might not be necessary, however is kept for now, since we are not aware of how information should be propagated back through such a link.\\
% comment on causal cycles, and how they can be unfolded by discrete time steps and thus can be viewed as low level graph idiom, and use this to argue that the reasoning behaviour in the case of such cycles can be viewed as part cognitive model rather than a special feature of the graphical model, and thus we do not need a special symbol to represent such cycles.\\

% Most discussions and presentations of complex topics can benefit from graphical illustrations. 
We have found that combining the semantics of classical directed graphical models with the semantics of flowcharts into a hybrid representation is a good visual representation. Directed graphical models represent the conditional dependency structure of a model and flowcharts represent the steps in an algorithm or workflow. The hybrid representation illustrates the order in which samples of random variables in a probabilistic program are generated and how these samples influence the distributions used to generate other samples. For this reason, we denote this hybrid representation by the name \textit{Generative Flow Graph}.

%\begin{figure}
%     \centering
%     \begin{subfigure}[tb]{1.0\linewidth}
%         \centering
%         \resizebox{1.0\linewidth}{!}{
%            \input{tikz/slam_graph_idiom.tikz}
%         }
%         \caption{Graph idiom for the classical simultaneous localization and mapping (SLAM) problem \cite{1638022}. $z^t_s$ is the state at time $t$, $z^{t}_a$ is the action at time $t$, $z^{i}_{map}$ is the $i$'th pixel in a grid map, and $x^t_{p}$ is the perceived information at time $t$. \textcolor{red}{ADD parentheses to indices}}
%         \label{fig:slam_graph_idiom}
%     \end{subfigure}
%     \hfill
%     \begin{subfigure}[tb]{1.0\linewidth}
%         \centering
%         \resizebox{1.0\linewidth}{!}{
%            \input{tikz/markov_decision_process_graph_idiom.tikz}
%         }
%         \caption{Graph idiom for a Markov Decision Process \cite{DBLP:journals/corr/abs-1805-00909}. $z^t_s$ is the state at time $t$, $z^t_a$ is the action at time $t$, and $x^t_{O}$ is a "observed" optimality variable at time $t$.}
%         \label{fig:markov_decision_process_graph_idiom}
%     \end{subfigure}
%        \caption{Examples of two directed graphical models developed in different research areas.}
%        \label{fig:example_of_graph_idioms}
%\end{figure}

\begin{figure}
     \centering
      \subfloat[Graph idiom for the classical simultaneous localization and mapping (SLAM) problem \cite{1638022}. $z^{\left\{t\right\}}_s$ is the state at time $t$, $z^{\left\{t\right\}}_a$ is the action at time $t$, $z^{\left\{i\right\}}_{map}$ is the $i$'th pixel in a grid map, and $x^{\left\{t\right\}}_{p}$ is the perceived information at time $t$. \label{fig:slam_graph_idiom}]{
            \resizebox{1.0\linewidth}{!}{
            \input{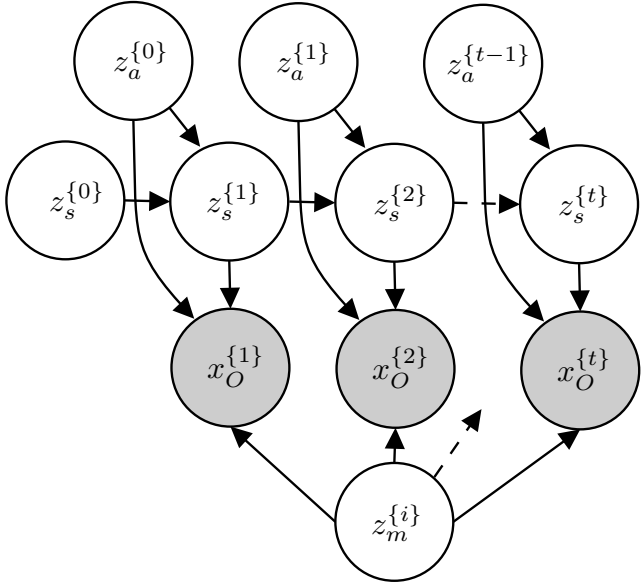}
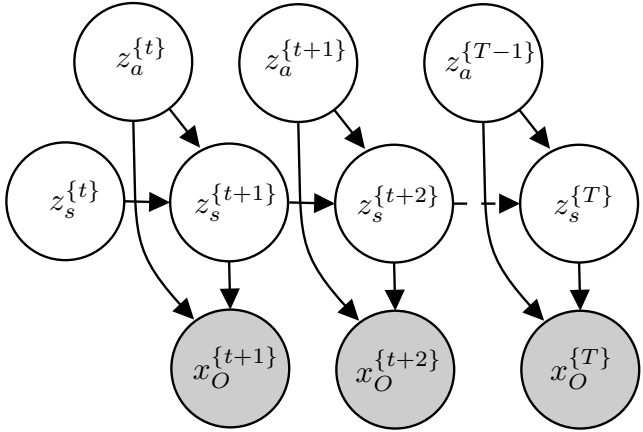
      }}\qquad 
      \subfloat[Graph idiom for a Markov Decision Process \cite{DBLP:journals/corr/abs-1805-00909}. $z^{\left\{t\right\}}_s$ is the state at time $t$, $z^{\left\{t\right\}}_a$ is the action at time $t$, and $x^{\left\{t\right\}}_{O}$ is a "observed" optimality variable at time $t$.
         \label{fig:markov_decision_process_graph_idiom}]{
         \resizebox{1.0\linewidth}{!}{
            \tikzset{every picture/.style={line width=0.75pt}} %set default line width to 0.75pt        

\begin{tikzpicture}[x=0.75pt,y=0.75pt,yscale=-1,xscale=1]
%uncomment if require: \path (0,191); %set diagram left start at 0, and has height of 191

%Shape: Circle [id:dp32224684580365226] 
\draw  [color={rgb, 255:red, 0; green, 0; blue, 0 }  ,draw opacity=1 ] (70.83,85) .. controls (70.83,71.19) and (82.03,60) .. (95.83,60) .. controls (109.64,60) and (120.83,71.19) .. (120.83,85) .. controls (120.83,98.81) and (109.64,110) .. (95.83,110) .. controls (82.03,110) and (70.83,98.81) .. (70.83,85) -- cycle ;
%Shape: Circle [id:dp5130976465228099] 
\draw   (30.17,25.33) .. controls (30.17,11.53) and (41.36,0.33) .. (55.17,0.33) .. controls (68.97,0.33) and (80.17,11.53) .. (80.17,25.33) .. controls (80.17,39.14) and (68.97,50.33) .. (55.17,50.33) .. controls (41.36,50.33) and (30.17,39.14) .. (30.17,25.33) -- cycle ;
%Straight Lines [id:da5741409379371931] 
\draw    (70.5,45.25) -- (81.64,59.39) ;
\draw [shift={(83.5,61.75)}, rotate = 231.77] [fill={rgb, 255:red, 0; green, 0; blue, 0 }  ][line width=0.08]  [draw opacity=0] (8.93,-4.29) -- (0,0) -- (8.93,4.29) -- cycle    ;
%Straight Lines [id:da24524762453845117] 
\draw    (51.33,84.5) -- (67.83,84.92) ;
\draw [shift={(70.83,85)}, rotate = 181.47] [fill={rgb, 255:red, 0; green, 0; blue, 0 }  ][line width=0.08]  [draw opacity=0] (8.93,-4.29) -- (0,0) -- (8.93,4.29) -- cycle    ;
%Shape: Circle [id:dp4303296282411586] 
\draw  [fill={rgb, 255:red, 155; green, 155; blue, 155 }  ,fill opacity=0.5 ] (71.33,155.5) .. controls (71.33,141.69) and (82.53,130.5) .. (96.33,130.5) .. controls (110.14,130.5) and (121.33,141.69) .. (121.33,155.5) .. controls (121.33,169.31) and (110.14,180.5) .. (96.33,180.5) .. controls (82.53,180.5) and (71.33,169.31) .. (71.33,155.5) -- cycle ;
%Straight Lines [id:da5093863137022343] 
\draw    (95.83,110) -- (96.26,127.5) ;
\draw [shift={(96.33,130.5)}, rotate = 268.6] [fill={rgb, 255:red, 0; green, 0; blue, 0 }  ][line width=0.08]  [draw opacity=0] (8.93,-4.29) -- (0,0) -- (8.93,4.29) -- cycle    ;
%Curve Lines [id:da14037637841007466] 
\draw    (55.17,50.33) .. controls (56.72,100.23) and (52.67,104.59) .. (79.33,132.98) ;
\draw [shift={(81,134.75)}, rotate = 226.47] [fill={rgb, 255:red, 0; green, 0; blue, 0 }  ][line width=0.08]  [draw opacity=0] (8.93,-4.29) -- (0,0) -- (8.93,4.29) -- cycle    ;
%Shape: Circle [id:dp9230994792832592] 
\draw  [color={rgb, 255:red, 0; green, 0; blue, 0 }  ,draw opacity=1 ] (1.33,84.5) .. controls (1.33,70.69) and (12.53,59.5) .. (26.33,59.5) .. controls (40.14,59.5) and (51.33,70.69) .. (51.33,84.5) .. controls (51.33,98.31) and (40.14,109.5) .. (26.33,109.5) .. controls (12.53,109.5) and (1.33,98.31) .. (1.33,84.5) -- cycle ;
%Shape: Circle [id:dp2033495092254176] 
\draw  [color={rgb, 255:red, 0; green, 0; blue, 0 }  ,draw opacity=1 ] (140.83,85.5) .. controls (140.83,71.69) and (152.03,60.5) .. (165.83,60.5) .. controls (179.64,60.5) and (190.83,71.69) .. (190.83,85.5) .. controls (190.83,99.31) and (179.64,110.5) .. (165.83,110.5) .. controls (152.03,110.5) and (140.83,99.31) .. (140.83,85.5) -- cycle ;
%Shape: Circle [id:dp008428111837865515] 
\draw   (100.17,25.83) .. controls (100.17,12.03) and (111.36,0.83) .. (125.17,0.83) .. controls (138.97,0.83) and (150.17,12.03) .. (150.17,25.83) .. controls (150.17,39.64) and (138.97,50.83) .. (125.17,50.83) .. controls (111.36,50.83) and (100.17,39.64) .. (100.17,25.83) -- cycle ;
%Straight Lines [id:da17140985921913754] 
\draw    (140.5,45.75) -- (151.64,59.89) ;
\draw [shift={(153.5,62.25)}, rotate = 231.77] [fill={rgb, 255:red, 0; green, 0; blue, 0 }  ][line width=0.08]  [draw opacity=0] (8.93,-4.29) -- (0,0) -- (8.93,4.29) -- cycle    ;
%Straight Lines [id:da8865334741896262] 
\draw    (121.33,85) -- (137.83,85.42) ;
\draw [shift={(140.83,85.5)}, rotate = 181.47] [fill={rgb, 255:red, 0; green, 0; blue, 0 }  ][line width=0.08]  [draw opacity=0] (8.93,-4.29) -- (0,0) -- (8.93,4.29) -- cycle    ;
%Shape: Circle [id:dp06641612577845035] 
\draw  [fill={rgb, 255:red, 155; green, 155; blue, 155 }  ,fill opacity=0.5 ] (141.33,156) .. controls (141.33,142.19) and (152.53,131) .. (166.33,131) .. controls (180.14,131) and (191.33,142.19) .. (191.33,156) .. controls (191.33,169.81) and (180.14,181) .. (166.33,181) .. controls (152.53,181) and (141.33,169.81) .. (141.33,156) -- cycle ;
%Straight Lines [id:da8787407963649227] 
\draw    (165.83,110.5) -- (166.26,128) ;
\draw [shift={(166.33,131)}, rotate = 268.6] [fill={rgb, 255:red, 0; green, 0; blue, 0 }  ][line width=0.08]  [draw opacity=0] (8.93,-4.29) -- (0,0) -- (8.93,4.29) -- cycle    ;
%Curve Lines [id:da9248383181252493] 
\draw    (125.17,50.83) .. controls (126.72,100.73) and (122.67,105.09) .. (149.33,133.48) ;
\draw [shift={(151,135.25)}, rotate = 226.47] [fill={rgb, 255:red, 0; green, 0; blue, 0 }  ][line width=0.08]  [draw opacity=0] (8.93,-4.29) -- (0,0) -- (8.93,4.29) -- cycle    ;
%Straight Lines [id:da42258789054997026] 
\draw  [dash pattern={on 4.5pt off 4.5pt}]  (190.83,85.5) -- (216.33,85.5) ;
\draw [shift={(219.33,85.5)}, rotate = 180] [fill={rgb, 255:red, 0; green, 0; blue, 0 }  ][line width=0.08]  [draw opacity=0] (8.93,-4.29) -- (0,0) -- (8.93,4.29) -- cycle    ;
%Shape: Circle [id:dp5981610573773304] 
\draw  [color={rgb, 255:red, 0; green, 0; blue, 0 }  ,draw opacity=1 ] (219.33,85.5) .. controls (219.33,71.69) and (230.53,60.5) .. (244.33,60.5) .. controls (258.14,60.5) and (269.33,71.69) .. (269.33,85.5) .. controls (269.33,99.31) and (258.14,110.5) .. (244.33,110.5) .. controls (230.53,110.5) and (219.33,99.31) .. (219.33,85.5) -- cycle ;
%Shape: Circle [id:dp5747327878963249] 
\draw   (178.67,25.83) .. controls (178.67,12.03) and (189.86,0.83) .. (203.67,0.83) .. controls (217.47,0.83) and (228.67,12.03) .. (228.67,25.83) .. controls (228.67,39.64) and (217.47,50.83) .. (203.67,50.83) .. controls (189.86,50.83) and (178.67,39.64) .. (178.67,25.83) -- cycle ;
%Straight Lines [id:da49381714519845277] 
\draw    (219,45.75) -- (230.14,59.89) ;
\draw [shift={(232,62.25)}, rotate = 231.77] [fill={rgb, 255:red, 0; green, 0; blue, 0 }  ][line width=0.08]  [draw opacity=0] (8.93,-4.29) -- (0,0) -- (8.93,4.29) -- cycle    ;
%Shape: Circle [id:dp5085378928142708] 
\draw  [fill={rgb, 255:red, 155; green, 155; blue, 155 }  ,fill opacity=0.5 ] (219.83,156) .. controls (219.83,142.19) and (231.03,131) .. (244.83,131) .. controls (258.64,131) and (269.83,142.19) .. (269.83,156) .. controls (269.83,169.81) and (258.64,181) .. (244.83,181) .. controls (231.03,181) and (219.83,169.81) .. (219.83,156) -- cycle ;
%Straight Lines [id:da2078684877643806] 
\draw    (244.33,110.5) -- (244.76,128) ;
\draw [shift={(244.83,131)}, rotate = 268.6] [fill={rgb, 255:red, 0; green, 0; blue, 0 }  ][line width=0.08]  [draw opacity=0] (8.93,-4.29) -- (0,0) -- (8.93,4.29) -- cycle    ;
%Curve Lines [id:da07456954012731631] 
\draw    (203.67,50.83) .. controls (205.22,100.73) and (201.17,105.09) .. (227.83,133.48) ;
\draw [shift={(229.5,135.25)}, rotate = 226.47] [fill={rgb, 255:red, 0; green, 0; blue, 0 }  ][line width=0.08]  [draw opacity=0] (8.93,-4.29) -- (0,0) -- (8.93,4.29) -- cycle    ;

% Text Node
\draw (46.17,13.67) node [anchor=north west][inner sep=0.75pt]   [align=left] {$\displaystyle z_{a}^{\{t\}}$};
% Text Node
\draw (79.5,75.6) node [anchor=north west][inner sep=0.75pt]  [color={rgb, 255:red, 0; green, 0; blue, 0 }  ,opacity=1 ] [align=left] {$\displaystyle z_{s}^{\{t+1\}}$};
% Text Node
\draw (79,144.5) node [anchor=north west][inner sep=0.75pt]   [align=left] {$\displaystyle x_{O}^{\{t+1\}}$};
% Text Node
\draw (18,74.1) node [anchor=north west][inner sep=0.75pt]  [color={rgb, 255:red, 0; green, 0; blue, 0 }  ,opacity=1 ] [align=left] {$\displaystyle z_{s}^{\{t\}}$};
% Text Node
\draw (109.17,14.17) node [anchor=north west][inner sep=0.75pt]   [align=left] {$\displaystyle z_{a}^{\{t+1\}}$};
% Text Node
\draw (148.5,76.1) node [anchor=north west][inner sep=0.75pt]  [color={rgb, 255:red, 0; green, 0; blue, 0 }  ,opacity=1 ] [align=left] {$\displaystyle z_{s}^{\{t+2\}}$};
% Text Node
\draw (148,146) node [anchor=north west][inner sep=0.75pt]   [align=left] {$\displaystyle x_{O}^{\{t+2\}}$};
% Text Node
\draw (185.67,15.17) node [anchor=north west][inner sep=0.75pt]   [align=left] {$\displaystyle z_{a}^{\{T-1\}}$};
% Text Node
\draw (232,76.1) node [anchor=north west][inner sep=0.75pt]  [color={rgb, 255:red, 0; green, 0; blue, 0 }  ,opacity=1 ] [align=left] {$\displaystyle z_{s}^{\{T\}}$};
% Text Node
\draw (230.5,145) node [anchor=north west][inner sep=0.75pt]   [align=left] {$\displaystyle x_{O}^{\{T\}}$};

\end{tikzpicture}
         }
     }
    \caption{Examples of two directed graphical models developed in different research areas.}
    \label{fig:example_of_graph_idioms}
\end{figure}

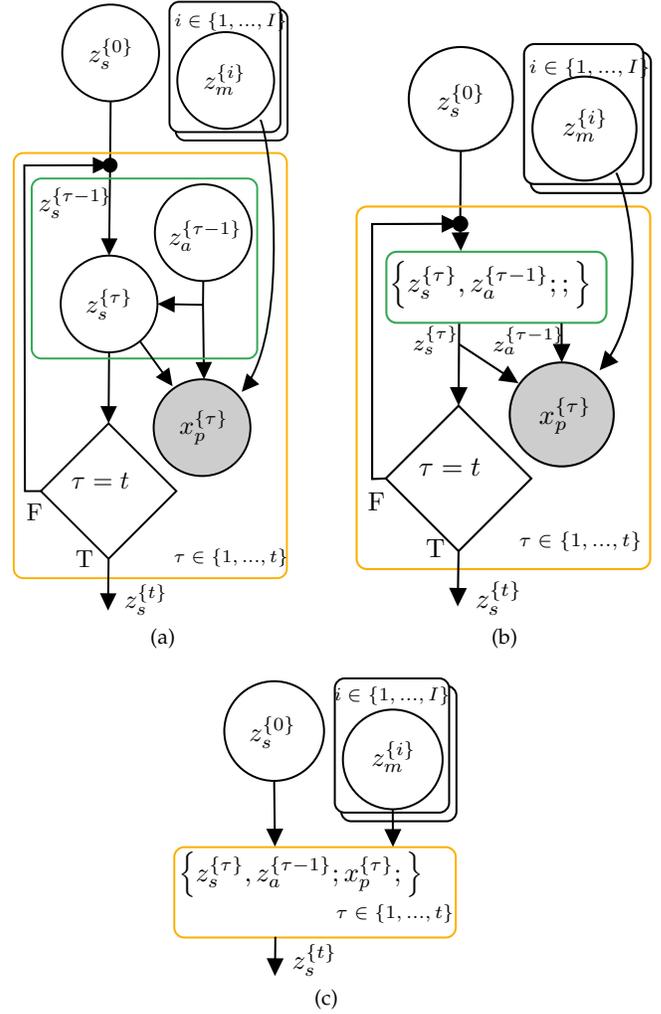
\begin{figure}[!h]
    \centering
    \subfloat[\label{fig:slam_graph_idiom_new_1}]{
        \resizebox{0.49\linewidth}{!}{
            \tikzset{every picture/.style={line width=0.75pt}} %set default line width to 0.75pt        

\begin{tikzpicture}[x=0.75pt,y=0.75pt,yscale=-1,xscale=1]
%uncomment if require: \path (0,331); %set diagram left start at 0, and has height of 331

%Shape: Circle [id:dp005333255655075808] 
\draw  [color={rgb, 255:red, 0; green, 0; blue, 0 }  ,draw opacity=1 ] (57.47,141.07) .. controls (71.27,141.07) and (82.47,152.26) .. (82.47,166.07) .. controls (82.47,179.87) and (71.27,191.07) .. (57.47,191.07) .. controls (43.66,191.07) and (32.47,179.87) .. (32.47,166.07) .. controls (32.47,152.26) and (43.66,141.07) .. (57.47,141.07) -- cycle ;
%Shape: Circle [id:dp3455099767383525] 
\draw  [color={rgb, 255:red, 0; green, 0; blue, 0 }  ,draw opacity=1 ] (106.13,104.2) .. controls (119.94,104.2) and (131.13,115.39) .. (131.13,129.2) .. controls (131.13,143.01) and (119.94,154.2) .. (106.13,154.2) .. controls (92.33,154.2) and (81.13,143.01) .. (81.13,129.2) .. controls (81.13,115.39) and (92.33,104.2) .. (106.13,104.2) -- cycle ;
%Straight Lines [id:da26447655009225524] 
\draw [color={rgb, 255:red, 0; green, 0; blue, 0 }  ,draw opacity=1 ]   (106.13,154.2) -- (106,166.53) -- (106.46,202.03) ;
\draw [shift={(106.5,205.03)}, rotate = 269.26] [fill={rgb, 255:red, 0; green, 0; blue, 0 }  ,fill opacity=1 ][line width=0.08]  [draw opacity=0] (8.93,-4.29) -- (0,0) -- (8.93,4.29) -- cycle    ;
%Straight Lines [id:da6452458793814004] 
\draw [color={rgb, 255:red, 0; green, 0; blue, 0 }  ,draw opacity=1 ]   (58.27,61.27) -- (58,94.3) ;
\draw [shift={(58,94.3)}, rotate = 90.46] [color={rgb, 255:red, 0; green, 0; blue, 0 }  ,draw opacity=1 ][fill={rgb, 255:red, 0; green, 0; blue, 0 }  ,fill opacity=1 ][line width=0.75]      (0, 0) circle [x radius= 3.35, y radius= 3.35]   ;
%Straight Lines [id:da6387898051718865] 
\draw [color={rgb, 255:red, 0; green, 0; blue, 0 }  ,draw opacity=1 ]   (58,94.3) -- (57.5,138.07) ;
\draw [shift={(57.47,141.07)}, rotate = 270.65] [fill={rgb, 255:red, 0; green, 0; blue, 0 }  ,fill opacity=1 ][line width=0.08]  [draw opacity=0] (8.93,-4.29) -- (0,0) -- (8.93,4.29) -- cycle    ;
%Straight Lines [id:da32791010539317345] 
\draw    (57.47,191.07) -- (57.22,224.83) ;
\draw [shift={(57.2,227.83)}, rotate = 270.42] [fill={rgb, 255:red, 0; green, 0; blue, 0 }  ][line width=0.08]  [draw opacity=0] (8.93,-4.29) -- (0,0) -- (8.93,4.29) -- cycle    ;
%Shape: Circle [id:dp9487371456750113] 
\draw  [fill={rgb, 255:red, 155; green, 155; blue, 155 }  ,fill opacity=0.5 ] (105.5,205.03) .. controls (119.31,205.03) and (130.5,216.23) .. (130.5,230.03) .. controls (130.5,243.84) and (119.31,255.03) .. (105.5,255.03) .. controls (91.69,255.03) and (80.5,243.84) .. (80.5,230.03) .. controls (80.5,216.23) and (91.69,205.03) .. (105.5,205.03) -- cycle ;
%Shape: Diamond [id:dp7542474717039827] 
\draw   (92.2,262.83) -- (57.2,297.83) -- (22.2,262.83) -- (57.2,227.83) -- cycle ;
%Shape: Circle [id:dp873439064452262] 
\draw  [color={rgb, 255:red, 0; green, 0; blue, 0 }  ,draw opacity=1 ] (58.27,11.27) .. controls (72.07,11.27) and (83.27,22.46) .. (83.27,36.27) .. controls (83.27,50.07) and (72.07,61.27) .. (58.27,61.27) .. controls (44.46,61.27) and (33.27,50.07) .. (33.27,36.27) .. controls (33.27,22.46) and (44.46,11.27) .. (58.27,11.27) -- cycle ;
%Straight Lines [id:da03552414085422306] 
\draw    (22.2,262.83) -- (14,262.86) -- (13.67,94.44) -- (55,94.31) ;
\draw [shift={(58,94.3)}, rotate = 539.8199999999999] [fill={rgb, 255:red, 0; green, 0; blue, 0 }  ][line width=0.08]  [draw opacity=0] (8.93,-4.29) -- (0,0) -- (8.93,4.29) -- cycle    ;
%Shape: Rectangle [id:dp6754351232240499] 
\draw  [color={rgb, 255:red, 251; green, 176; blue, 5 }  ,draw opacity=1 ] (8.2,93.02) .. controls (8.2,90.26) and (10.44,88.02) .. (13.2,88.02) -- (144.33,88.02) .. controls (147.09,88.02) and (149.33,90.26) .. (149.33,93.02) -- (149.33,302.86) .. controls (149.33,305.63) and (147.09,307.86) .. (144.33,307.86) -- (13.2,307.86) .. controls (10.44,307.86) and (8.2,305.63) .. (8.2,302.86) -- cycle ;
%Straight Lines [id:da6440409427427769] 
\draw [color={rgb, 255:red, 0; green, 0; blue, 0 }  ,draw opacity=1 ]   (106,166.53) -- (85.47,166.13) ;
\draw [shift={(82.47,166.07)}, rotate = 361.13] [fill={rgb, 255:red, 0; green, 0; blue, 0 }  ,fill opacity=1 ][line width=0.08]  [draw opacity=0] (8.93,-4.29) -- (0,0) -- (8.93,4.29) -- cycle    ;
%Shape: Rectangle [id:dp18899657965197525] 
\draw  [color={rgb, 255:red, 52; green, 168; blue, 83 }  ,draw opacity=1 ] (17.5,106.2) .. controls (17.5,103.44) and (19.74,101.2) .. (22.5,101.2) -- (129,101.2) .. controls (131.76,101.2) and (134,103.44) .. (134,106.2) -- (134,189.27) .. controls (134,192.03) and (131.76,194.27) .. (129,194.27) -- (22.5,194.27) .. controls (19.74,194.27) and (17.5,192.03) .. (17.5,189.27) -- cycle ;
%Straight Lines [id:da5977261130339302] 
\draw    (73.43,185.36) -- (88.53,206.21) ;
\draw [shift={(90.29,208.64)}, rotate = 234.1] [fill={rgb, 255:red, 0; green, 0; blue, 0 }  ][line width=0.08]  [draw opacity=0] (8.93,-4.29) -- (0,0) -- (8.93,4.29) -- cycle    ;
%Shape: Rectangle [id:dp3140792287476728] 
\draw   (91.87,18.64) .. controls (91.87,15.88) and (94.11,13.64) .. (96.87,13.64) -- (144.67,13.64) .. controls (147.43,13.64) and (149.67,15.88) .. (149.67,18.64) -- (149.67,76.52) .. controls (149.67,79.28) and (147.43,81.52) .. (144.67,81.52) -- (96.87,81.52) .. controls (94.11,81.52) and (91.87,79.28) .. (91.87,76.52) -- cycle ;
%Shape: Rectangle [id:dp4703906656786503] 
\draw  [fill={rgb, 255:red, 255; green, 255; blue, 255 }  ,fill opacity=1 ] (88.67,14.84) .. controls (88.67,12.08) and (90.91,9.84) .. (93.67,9.84) -- (141,9.84) .. controls (143.76,9.84) and (146,12.08) .. (146,14.84) -- (146,72.19) .. controls (146,74.95) and (143.76,77.19) .. (141,77.19) -- (93.67,77.19) .. controls (90.91,77.19) and (88.67,74.95) .. (88.67,72.19) -- cycle ;
%Shape: Circle [id:dp8097054844981943] 
\draw  [color={rgb, 255:red, 0; green, 0; blue, 0 }  ,draw opacity=1 ] (117.73,25.5) .. controls (131.54,25.5) and (142.73,36.69) .. (142.73,50.5) .. controls (142.73,64.31) and (131.54,75.5) .. (117.73,75.5) .. controls (103.93,75.5) and (92.73,64.31) .. (92.73,50.5) .. controls (92.73,36.69) and (103.93,25.5) .. (117.73,25.5) -- cycle ;
%Curve Lines [id:da06621201820216305] 
\draw    (135.67,70.86) .. controls (140.89,80.99) and (151.56,177.55) .. (127.52,209.64) ;
\draw [shift={(126,211.53)}, rotate = 310.90999999999997] [fill={rgb, 255:red, 0; green, 0; blue, 0 }  ][line width=0.08]  [draw opacity=0] (8.93,-4.29) -- (0,0) -- (8.93,4.29) -- cycle    ;
%Straight Lines [id:da369365207859077] 
\draw    (57.2,297.83) -- (56.96,321.6) ;
\draw [shift={(56.93,324.6)}, rotate = 270.57] [fill={rgb, 255:red, 0; green, 0; blue, 0 }  ][line width=0.08]  [draw opacity=0] (8.93,-4.29) -- (0,0) -- (8.93,4.29) -- cycle    ;

% Text Node
\draw (86.7,121.3) node [anchor=north west][inner sep=0.75pt]  [color={rgb, 255:red, 0; green, 0; blue, 0 }  ,opacity=1 ] [align=left] {$\displaystyle z_{a}^{\{\tau -1\}}$};
% Text Node
\draw (43.87,156.73) node [anchor=north west][inner sep=0.75pt]  [color={rgb, 255:red, 0; green, 0; blue, 0 }  ,opacity=1 ] [align=left] {$\displaystyle z_{s}^{\{\tau \}}$};
% Text Node
\draw (45.17,27.1) node [anchor=north west][inner sep=0.75pt]  [color={rgb, 255:red, 0; green, 0; blue, 0 }  ,opacity=1 ] [align=left] {$\displaystyle z_{s}^{\{0\}}$};
% Text Node
\draw (36.5,251.7) node [anchor=north west][inner sep=0.75pt]   [align=left] {$\displaystyle \tau =t$};
% Text Node
\draw (13.33,267.87) node [anchor=north west][inner sep=0.75pt]   [align=left] {F};
% Text Node
\draw (124.77,297.09) node   [align=left] {\begin{minipage}[lt]{50.64pt}\setlength\topsep{0pt}
{\scriptsize $\displaystyle \tau \in \{1,...,t\}$}
\end{minipage}};
% Text Node
\draw (104.63,40.67) node [anchor=north west][inner sep=0.75pt]  [color={rgb, 255:red, 0; green, 0; blue, 0 }  ,opacity=1 ] [align=left] {$\displaystyle z_{m}^{\{i\}}$};
% Text Node
\draw (92.4,219.4) node [anchor=north west][inner sep=0.75pt]   [align=left] {$\displaystyle x_{p}^{\{\tau \}}$};
% Text Node
\draw (116.6,20.19) node   [align=left] {\begin{minipage}[lt]{37.99pt}\setlength\topsep{0pt}
{\scriptsize $\displaystyle {\textstyle i\in \{1,...,I\}}$}
\end{minipage}};
% Text Node
\draw (38.67,292.2) node [anchor=north west][inner sep=0.75pt]   [align=left] {T};
% Text Node
\draw (63.87,309.4) node [anchor=north west][inner sep=0.75pt]  [color={rgb, 255:red, 0; green, 0; blue, 0 }  ,opacity=1 ] [align=left] {$\displaystyle z_{s}^{\{t\}}$};
% Text Node
\draw (19.5,104.2) node [anchor=north west][inner sep=0.75pt]  [color={rgb, 255:red, 0; green, 0; blue, 0 }  ,opacity=1 ] [align=left] {$\displaystyle z_{s}^{\{\tau -1\}}$};

\end{tikzpicture}
         }
    } 
    \subfloat[\label{fig:slam_graph_idiom_new_2}]{
        \resizebox{0.49\linewidth}{!}{
            \tikzset{every picture/.style={line width=0.75pt}} %set default line width to 0.75pt        

\begin{tikzpicture}[x=0.75pt,y=0.75pt,yscale=-1,xscale=1]
%uncomment if require: \path (0,294); %set diagram left start at 0, and has height of 294

%Straight Lines [id:da2943353731327192] 
\draw [color={rgb, 255:red, 0; green, 0; blue, 0 }  ,draw opacity=1 ]   (58.27,61.27) -- (58,96.3) ;
\draw [shift={(58,96.3)}, rotate = 90.44] [color={rgb, 255:red, 0; green, 0; blue, 0 }  ,draw opacity=1 ][fill={rgb, 255:red, 0; green, 0; blue, 0 }  ,fill opacity=1 ][line width=0.75]      (0, 0) circle [x radius= 3.35, y radius= 3.35]   ;
%Shape: Diamond [id:dp42121061768059076] 
\draw   (92.2,218.83) -- (57.2,253.83) -- (22.2,218.83) -- (57.2,183.83) -- cycle ;
%Shape: Circle [id:dp38071016788532996] 
\draw  [color={rgb, 255:red, 0; green, 0; blue, 0 }  ,draw opacity=1 ] (58.27,11.27) .. controls (72.07,11.27) and (83.27,22.46) .. (83.27,36.27) .. controls (83.27,50.07) and (72.07,61.27) .. (58.27,61.27) .. controls (44.46,61.27) and (33.27,50.07) .. (33.27,36.27) .. controls (33.27,22.46) and (44.46,11.27) .. (58.27,11.27) -- cycle ;
%Straight Lines [id:da9472302165229785] 
\draw    (22.2,218.83) -- (15.67,219.11) -- (15.67,96.44) -- (55,96.31) ;
\draw [shift={(58,96.3)}, rotate = 539.81] [fill={rgb, 255:red, 0; green, 0; blue, 0 }  ][line width=0.08]  [draw opacity=0] (8.93,-4.29) -- (0,0) -- (8.93,4.29) -- cycle    ;
%Shape: Rectangle [id:dp6679275305770069] 
\draw  [color={rgb, 255:red, 251; green, 176; blue, 5 }  ,draw opacity=1 ] (8.2,93.02) .. controls (8.2,90.26) and (10.44,88.02) .. (13.2,88.02) -- (144.33,88.02) .. controls (147.09,88.02) and (149.33,90.26) .. (149.33,93.02) -- (149.33,257.43) .. controls (149.33,260.19) and (147.09,262.43) .. (144.33,262.43) -- (13.2,262.43) .. controls (10.44,262.43) and (8.2,260.19) .. (8.2,257.43) -- cycle ;
%Shape: Rectangle [id:dp8308700348917879] 
\draw   (91.87,18.64) .. controls (91.87,15.88) and (94.11,13.64) .. (96.87,13.64) -- (144.67,13.64) .. controls (147.43,13.64) and (149.67,15.88) .. (149.67,18.64) -- (149.67,76.52) .. controls (149.67,79.28) and (147.43,81.52) .. (144.67,81.52) -- (96.87,81.52) .. controls (94.11,81.52) and (91.87,79.28) .. (91.87,76.52) -- cycle ;
%Shape: Rectangle [id:dp09612176652207016] 
\draw  [fill={rgb, 255:red, 255; green, 255; blue, 255 }  ,fill opacity=1 ] (88.67,14.84) .. controls (88.67,12.08) and (90.91,9.84) .. (93.67,9.84) -- (141,9.84) .. controls (143.76,9.84) and (146,12.08) .. (146,14.84) -- (146,72.19) .. controls (146,74.95) and (143.76,77.19) .. (141,77.19) -- (93.67,77.19) .. controls (90.91,77.19) and (88.67,74.95) .. (88.67,72.19) -- cycle ;
%Shape: Circle [id:dp9963460207080164] 
\draw  [color={rgb, 255:red, 0; green, 0; blue, 0 }  ,draw opacity=1 ] (117.73,25.5) .. controls (131.54,25.5) and (142.73,36.69) .. (142.73,50.5) .. controls (142.73,64.31) and (131.54,75.5) .. (117.73,75.5) .. controls (103.93,75.5) and (92.73,64.31) .. (92.73,50.5) .. controls (92.73,36.69) and (103.93,25.5) .. (117.73,25.5) -- cycle ;
%Curve Lines [id:da8596093931788025] 
\draw    (133,71.78) .. controls (138.2,81.86) and (150.37,134.08) .. (126.88,164.49) ;
\draw [shift={(125,166.78)}, rotate = 310.90999999999997] [fill={rgb, 255:red, 0; green, 0; blue, 0 }  ][line width=0.08]  [draw opacity=0] (8.93,-4.29) -- (0,0) -- (8.93,4.29) -- cycle    ;
%Straight Lines [id:da02121405958508671] 
\draw    (57.5,143.76) -- (57.22,180.83) ;
\draw [shift={(57.2,183.83)}, rotate = 270.43] [fill={rgb, 255:red, 0; green, 0; blue, 0 }  ][line width=0.08]  [draw opacity=0] (8.93,-4.29) -- (0,0) -- (8.93,4.29) -- cycle    ;
%Shape: Circle [id:dp0885378496445064] 
\draw  [fill={rgb, 255:red, 155; green, 155; blue, 155 }  ,fill opacity=0.5 ] (106.8,163.03) .. controls (120.61,163.03) and (131.8,174.23) .. (131.8,188.03) .. controls (131.8,201.84) and (120.61,213.03) .. (106.8,213.03) .. controls (92.99,213.03) and (81.8,201.84) .. (81.8,188.03) .. controls (81.8,174.23) and (92.99,163.03) .. (106.8,163.03) -- cycle ;
%Shape: Rectangle [id:dp3814909991002613] 
\draw  [color={rgb, 255:red, 52; green, 168; blue, 83 }  ,draw opacity=1 ] (22.5,114.77) .. controls (22.5,112) and (24.74,109.77) .. (27.5,109.77) -- (123.3,109.77) .. controls (126.06,109.77) and (128.3,112) .. (128.3,114.77) -- (128.3,138.96) .. controls (128.3,141.72) and (126.06,143.96) .. (123.3,143.96) -- (27.5,143.96) .. controls (24.74,143.96) and (22.5,141.72) .. (22.5,138.96) -- cycle ;
%Straight Lines [id:da0942013781143558] 
\draw    (58,96.3) -- (58.23,106.38) ;
\draw [shift={(58.3,109.38)}, rotate = 268.69] [fill={rgb, 255:red, 0; green, 0; blue, 0 }  ][line width=0.08]  [draw opacity=0] (8.93,-4.29) -- (0,0) -- (8.93,4.29) -- cycle    ;
%Straight Lines [id:da6636568085844292] 
\draw    (106.7,144.16) -- (106.78,160.03) ;
\draw [shift={(106.8,163.03)}, rotate = 269.7] [fill={rgb, 255:red, 0; green, 0; blue, 0 }  ][line width=0.08]  [draw opacity=0] (8.93,-4.29) -- (0,0) -- (8.93,4.29) -- cycle    ;
%Straight Lines [id:da7962478968225024] 
\draw    (57.5,154.28) -- (83.51,171.38) ;
\draw [shift={(86.01,173.03)}, rotate = 213.32999999999998] [fill={rgb, 255:red, 0; green, 0; blue, 0 }  ][line width=0.08]  [draw opacity=0] (8.93,-4.29) -- (0,0) -- (8.93,4.29) -- cycle    ;
%Straight Lines [id:da9300855555043863] 
\draw    (57.2,253.83) -- (56.85,277.43) ;
\draw [shift={(56.8,280.43)}, rotate = 270.86] [fill={rgb, 255:red, 0; green, 0; blue, 0 }  ][line width=0.08]  [draw opacity=0] (8.93,-4.29) -- (0,0) -- (8.93,4.29) -- cycle    ;

% Text Node
\draw (45.17,27.43) node [anchor=north west][inner sep=0.75pt]  [color={rgb, 255:red, 0; green, 0; blue, 0 }  ,opacity=1 ] [align=left] {$\displaystyle z_{s}^{\{0\}}$};
% Text Node
\draw (36.5,207.7) node [anchor=north west][inner sep=0.75pt]   [align=left] {$\displaystyle \tau =t$};
% Text Node
\draw (12.2,224.07) node [anchor=north west][inner sep=0.75pt]   [align=left] {F};
% Text Node
\draw (116.53,249.32) node   [align=left] {\begin{minipage}[lt]{45.33pt}\setlength\topsep{0pt}
{\scriptsize $\displaystyle \tau \in \{1,...,t\}$}
\end{minipage}};
% Text Node
\draw (105.63,39.67) node [anchor=north west][inner sep=0.75pt]  [color={rgb, 255:red, 0; green, 0; blue, 0 }  ,opacity=1 ] [align=left] {$\displaystyle z_{m}^{\{i\}}$};
% Text Node
\draw (94.4,178.4) node [anchor=north west][inner sep=0.75pt]   [align=left] {$\displaystyle x_{p}^{\{\tau \}}$};
% Text Node
\draw (116.8,21.51) node   [align=left] {\begin{minipage}[lt]{38.54pt}\setlength\topsep{0pt}
{\scriptsize $\displaystyle {\textstyle i\in \{1,...,I\}}$}
\end{minipage}};
% Text Node
\draw (33.87,144.19) node [anchor=north west][inner sep=0.75pt]  [color={rgb, 255:red, 0; green, 0; blue, 0 }  ,opacity=1 ] [align=left] {{\footnotesize $\displaystyle z_{s}^{\{\tau \}}$}};
% Text Node
\draw (72.3,144.41) node [anchor=north west][inner sep=0.75pt]  [color={rgb, 255:red, 0; green, 0; blue, 0 }  ,opacity=1 ] [align=left] {{\footnotesize $\displaystyle z_{a}^{\{\tau -1\}}$}};
% Text Node
\draw (40.03,246.67) node [anchor=north west][inner sep=0.75pt]   [align=left] {T};
% Text Node
\draw (63.87,266.97) node [anchor=north west][inner sep=0.75pt]  [color={rgb, 255:red, 0; green, 0; blue, 0 }  ,opacity=1 ] [align=left] {$\displaystyle z_{s}^{\{t\}}$};
% Text Node
\draw (22.5,112.77) node [anchor=north west][inner sep=0.75pt]  [color={rgb, 255:red, 0; green, 0; blue, 0 }  ,opacity=1 ] [align=left] {$\displaystyle \left\{z_{s}^{\{\tau \}} ,z_{a}^{\{\tau -1\}} ;;\right\}$};

\end{tikzpicture}
         }
    }\qquad
    \subfloat[\label{fig:slam_graph_idiom_new_3}]{
        \resizebox{0.5\linewidth}{!}{
            \tikzset{every picture/.style={line width=0.75pt}} %set default line width to 0.75pt        

\begin{tikzpicture}[x=0.75pt,y=0.75pt,yscale=-1,xscale=1]
%uncomment if require: \path (0,167); %set diagram left start at 0, and has height of 167

%Shape: Circle [id:dp2697538530512036] 
\draw  [color={rgb, 255:red, 0; green, 0; blue, 0 }  ,draw opacity=1 ] (58.27,11.27) .. controls (72.07,11.27) and (83.27,22.46) .. (83.27,36.27) .. controls (83.27,50.07) and (72.07,61.27) .. (58.27,61.27) .. controls (44.46,61.27) and (33.27,50.07) .. (33.27,36.27) .. controls (33.27,22.46) and (44.46,11.27) .. (58.27,11.27) -- cycle ;
%Shape: Rectangle [id:dp214655616888632] 
\draw  [color={rgb, 255:red, 251; green, 176; blue, 5 }  ,draw opacity=1 ] (8.2,99.53) .. controls (8.2,96.77) and (10.44,94.53) .. (13.2,94.53) -- (144.33,94.53) .. controls (147.09,94.53) and (149.33,96.77) .. (149.33,99.53) -- (149.33,134.86) .. controls (149.33,137.63) and (147.09,139.86) .. (144.33,139.86) -- (13.2,139.86) .. controls (10.44,139.86) and (8.2,137.63) .. (8.2,134.86) -- cycle ;
%Shape: Rectangle [id:dp6084085767910845] 
\draw   (91.87,18.64) .. controls (91.87,15.88) and (94.11,13.64) .. (96.87,13.64) -- (144.67,13.64) .. controls (147.43,13.64) and (149.67,15.88) .. (149.67,18.64) -- (149.67,76.52) .. controls (149.67,79.28) and (147.43,81.52) .. (144.67,81.52) -- (96.87,81.52) .. controls (94.11,81.52) and (91.87,79.28) .. (91.87,76.52) -- cycle ;
%Shape: Rectangle [id:dp3789372202905994] 
\draw  [fill={rgb, 255:red, 255; green, 255; blue, 255 }  ,fill opacity=1 ] (88.67,14.84) .. controls (88.67,12.08) and (90.91,9.84) .. (93.67,9.84) -- (141,9.84) .. controls (143.76,9.84) and (146,12.08) .. (146,14.84) -- (146,72.19) .. controls (146,74.95) and (143.76,77.19) .. (141,77.19) -- (93.67,77.19) .. controls (90.91,77.19) and (88.67,74.95) .. (88.67,72.19) -- cycle ;
%Shape: Circle [id:dp6235470051798742] 
\draw  [color={rgb, 255:red, 0; green, 0; blue, 0 }  ,draw opacity=1 ] (117.73,25.5) .. controls (131.54,25.5) and (142.73,36.69) .. (142.73,50.5) .. controls (142.73,64.31) and (131.54,75.5) .. (117.73,75.5) .. controls (103.93,75.5) and (92.73,64.31) .. (92.73,50.5) .. controls (92.73,36.69) and (103.93,25.5) .. (117.73,25.5) -- cycle ;
%Straight Lines [id:da5179056839339837] 
\draw    (58.27,61.27) -- (58.63,91.2) ;
\draw [shift={(58.67,94.2)}, rotate = 269.3] [fill={rgb, 255:red, 0; green, 0; blue, 0 }  ][line width=0.08]  [draw opacity=0] (8.93,-4.29) -- (0,0) -- (8.93,4.29) -- cycle    ;
%Straight Lines [id:da9937550522796414] 
\draw    (117.73,75.5) -- (117.82,91.38) ;
\draw [shift={(117.83,94.38)}, rotate = 269.7] [fill={rgb, 255:red, 0; green, 0; blue, 0 }  ][line width=0.08]  [draw opacity=0] (8.93,-4.29) -- (0,0) -- (8.93,4.29) -- cycle    ;
%Straight Lines [id:da09914175899038224] 
\draw    (58.87,139.5) -- (58.7,156.2) ;
\draw [shift={(58.67,159.2)}, rotate = 270.58] [fill={rgb, 255:red, 0; green, 0; blue, 0 }  ][line width=0.08]  [draw opacity=0] (8.93,-4.29) -- (0,0) -- (8.93,4.29) -- cycle    ;

% Text Node
\draw (44.17,27.43) node [anchor=north west][inner sep=0.75pt]  [color={rgb, 255:red, 0; green, 0; blue, 0 }  ,opacity=1 ] [align=left] {$\displaystyle z_{s}^{\{0\}}$};
% Text Node
\draw (123.1,127.98) node   [align=left] {\begin{minipage}[lt]{50.64pt}\setlength\topsep{0pt}
{\scriptsize $\displaystyle \tau \in \{1,...,t\}$}
\end{minipage}};
% Text Node
\draw (105.63,39.67) node [anchor=north west][inner sep=0.75pt]  [color={rgb, 255:red, 0; green, 0; blue, 0 }  ,opacity=1 ] [align=left] {$\displaystyle z_{m}^{\{i\}}$};
% Text Node
\draw (115.02,18.33) node   [align=left] {\begin{minipage}[lt]{40.19pt}\setlength\topsep{0pt}
{\scriptsize $\displaystyle {\textstyle i\in \{1,...,I\}}$}
\end{minipage}};
% Text Node
\draw (65.87,141.64) node [anchor=north west][inner sep=0.75pt]  [color={rgb, 255:red, 0; green, 0; blue, 0 }  ,opacity=1 ] [align=left] {$\displaystyle z_{s}^{\{t\}}$};
% Text Node
\draw (9,95.16) node [anchor=north west][inner sep=0.75pt]  [color={rgb, 255:red, 0; green, 0; blue, 0 }  ,opacity=1 ] [align=left] {$\displaystyle \left\{z_{s}^{\{\tau \}} ,z_{a}^{\{\tau -1\}} ;x_{p}^{\{\tau \}} ;\right\}$};

\end{tikzpicture}
         }
    }
    \caption{Three semantically equivalent generative flow graphs with different levels of abstractions corresponding to the directed graphical model in \cref{fig:slam_graph_idiom}.}
    \label{fig:slam_graph_idiom_new}
\end{figure}

To exemplify the utility of the \textit{Generative Flow Graph} representation consider the graphical model for a classical Markov Decision Process and the simultaneous localization and mapping (SLAM) problem depicted in \cref{fig:example_of_graph_idioms}. With the classical semantics of directed graphical models, it is often the case that size limitations of figures coerce authors to remove some variables from the figure and represent them indirectly by, e.g., dashed arrows as the case in both \cref{fig:slam_graph_idiom} and \cref{fig:markov_decision_process_graph_idiom}. Similarly, the classical semantics of directed graphical models does not represent the influence from other parameters or variables that are not given a probabilistic treatment, even though such variables and parameters might have equal importance for a model. This is especially true if they are not fixed and have to be learned, e.g., if one wants to incorporate artificial neural networks into a model. The classical semantics of directed graphical models also cannot represent dependency structures depending on conditionals giving the illusion that a variable always depends on all of its possible parents, and that all variables in the graph are relevant in all situations. Furthermore, while the semantics of directed graphical models allows us to represent the structure of the joint distribution, $p(Z,X)$, its ability to explicitly express the structure of the posterior distribution, $p(Z|X)$, is limited. Finally, there is no standardized ways of representing a fragment of a graphical model, which hinders discussions at different levels of abstraction. Probabilistic programs easily allow us to incorporate the above in our models and thus a more appropriate representation is needed. The semantics of generative flow graphs shown in \cref{table:generative_flow_graph_semantics} in \cref{sec:appendix} alleviate these problems. Utilizing these semantics we can redraw the directed graphical model in \cref{fig:slam_graph_idiom} in multiple ways with different levels of information as in \cref{fig:slam_graph_idiom_new}. Notice, that the choice of node collections is not unique.

%\begin{figure}
%     \centering
%     \begin{subfigure}[tb]{0.5\linewidth}
%         \centering
%         \resizebox{1.0\linewidth}{!}{
%            \input{tikz/slam_graph_idiom_new_1.tikz}
%         }
%         \caption{}
%         \label{fig:slam_graph_idiom_new_1}
%     \end{subfigure}%
%     \begin{subfigure}[tb]{0.5\linewidth}
%         \centering
%         \resizebox{1.0\linewidth}{!}{
%            \input{tikz/slam_graph_idiom_new_2.tikz}
%         }
%         \caption{}
%         \label{fig:slam_graph_idiom_new_2}
%     \end{subfigure}
%     \hfill
%     \begin{subfigure}[tb]{0.5\linewidth}
%         \centering
%         \resizebox{1.0\linewidth}{!}{
%            \input{tikz/slam_graph_idiom_new_3.tikz}
%         }
%         \caption{}
%         \label{fig:slam_graph_idiom_new_3}
%     \end{subfigure}
%     \caption{Three semantically equivalent generative flow graphs with different levels of abstractions corresponding to the directed graphical model in \cref{fig:slam_graph_idiom}. \textcolor{red}{consider adding params to examplify information often neglected in graphical models.} \textcolor{red}{Consider moving obs outside collection, to relate to the global obs in S-msg-passing}}
%     \label{fig:slam_graph_idiom_new}
%\end{figure}

One advantage of the semantics of directed graphical models is that for graphs with no cycles \cite{10.5555/1795555} such models represents a specific factorization of the joint probability of all the random variables in the model of the form:
\begin{align}
    &p( x^{\left\{1;\overline{n}\right\}} ,z^{\left\{1;\overline{m}\right\}}) \label{eq:DAG_fact}\\
    &\quad=\prod _{n=1}^{\overline{n}} p( x^{\left\{n\right\}} |\text{Pa}Z( x^{\left\{n\right\}}))\prod _{m=1}^{\overline{m}} p( z^{\left\{m\right\}} |PaZ(z^{\left\{m\right\}})) \nonumber
\end{align}
where $x^{\left\{n\right\}}$ and $z^{\left\{m\right\}}$ are the n'th observed and the m'th latent random variable in the model, respectively. This is in principle also true for the generative flow graph representation if it neither contains any cycles, just with the additional explicit representation of dependency on parameters. For generative flow graphs, we can similarly to \cref{eq:DAG_fact} write up a factorization by including a factor of the form 
\begin{align*}
    p_{\text{Pa}\theta( z^{\left\{m\right\}})}( z^{\left\{m\right\}} |\text{Pa}Z(z^{\left\{m\right\}}))
\end{align*}
for each latent random variable node, $z^{\left\{m\right\}}$, in the graph, and a factor of the form
\begin{align*}
    p_{\text{Pa}\theta( x^{\left\{n\right\}})}( x^{\left\{n\right\}} |\text{Pa}Z( x^{\left\{n\right\}}))
\end{align*}
for each observed random variable node, $x^{\left\{n\right\}}$, in the graph, and finally a factor of the form
\begin{align}
 & p_{\Theta ,\text{Pa}\Theta(C^{\left\{k\right\}})} (Z,X|\text{Pa}Z(C^{\left\{k\right\}})) \label{eq:node_collection_factor}%\\
 %& =p_{\Theta ,\text{Pa}\Theta(C^{\left\{k\right\}})} (X|Z,\text{Pa}Z(C^{\left\{k\right\}}))p_{\Theta ,\text{Pa}\Theta(C^{\left\{k\right\}})} (z|\text{Pa}Z(C^{\left\{k\right\}})) \nonumber
\end{align}

for each node collection, $C^{\left\{k\right\}} = \{Z;X;\Theta\}$. If a parent node of $y$ is a node collection $\{Z;X;\Theta\}$ then $\text{Pa}Z(y)=Z$ and $\text{Pa}\theta(y)=\Theta$ unless a subset of the variables or parameters in the node collection is explicitly specified next to the parent link. If the internal structure of a node collection is known from somewhere else, the factor in \cref{eq:node_collection_factor} can of course be replaced by the corresponding factorization. The catch, however, is that a probabilistic program, and thus also generative flow graphs, potentially can denote models with an unbounded number of random variables and parameters making it impossible to write up the full factorization explicitly. On the other hand, this just emphasizes the need for alternative ways of representing probabilistic programs other than pseudo-code.

%\begin{figure}
%     \centering
%     \begin{subfigure}[tb]{0.45\linewidth}
%         \centering
%         \resizebox{1.0\linewidth}{!}{
%            \input{tikz/detached_links_1.tikz}
%         }
%         \caption{}
%         \label{fig:detached_links_1}
%     \end{subfigure}%
%     \hfill
%     \begin{subfigure}[tb]{0.45\linewidth}
%         \centering
%         \resizebox{1.0\linewidth}{!}{
%            \input{tikz/detached_links_2.tikz}
%         }
%         \caption{}
%         \label{fig:detached_links_2}
%     \end{subfigure}
%     \caption{Two generative flow graphs representations of a simple model with two parameters, two latent variables, and two observed variables. In both graphs $\theta_1$ and $z_1$ are needed to generate $z_2$, however, in (b) we have explicitly constrained the inference of $z_2$ to not influence learning of $\theta_1$ and inference of $z_1$. Thus, the evidence provided by $x_2$ neither are allowed to have an influence on $\theta_1$ and $z_1$. Thereby, the model represented by the nodes on the left hand side of the dashed line in (b) can be seen as an independent problem.}
%     \label{fig:detached_links}
%\end{figure}

\begin{figure}[!t]
    \centering
    \subfloat[\label{fig:detached_links_1}]{
        \resizebox{0.4\linewidth}{!}{
            \tikzset{every picture/.style={line width=0.75pt}} %set default line width to 0.75pt        

\begin{tikzpicture}[x=0.75pt,y=0.75pt,yscale=-1,xscale=1]
%uncomment if require: \path (0,194); %set diagram left start at 0, and has height of 194

%Shape: Circle [id:dp7353550616790101] 
\draw  [color={rgb, 255:red, 0; green, 0; blue, 0 }  ,draw opacity=1 ] (26,70) .. controls (39.81,70) and (51,81.19) .. (51,95) .. controls (51,108.81) and (39.81,120) .. (26,120) .. controls (12.19,120) and (1,108.81) .. (1,95) .. controls (1,81.19) and (12.19,70) .. (26,70) -- cycle ;
%Shape: Circle [id:dp9429735891337547] 
\draw  [color={rgb, 255:red, 0; green, 0; blue, 0 }  ,draw opacity=1 ] (106,70) .. controls (119.81,70) and (131,81.19) .. (131,95) .. controls (131,108.81) and (119.81,120) .. (106,120) .. controls (92.19,120) and (81,108.81) .. (81,95) .. controls (81,81.19) and (92.19,70) .. (106,70) -- cycle ;
%Straight Lines [id:da05525604030828668] 
\draw    (51,95) -- (66,95) ;
%Straight Lines [id:da6174771046816008] 
\draw    (60,95) -- (78,95) ;
\draw [shift={(81,95)}, rotate = 180] [fill={rgb, 255:red, 0; green, 0; blue, 0 }  ][line width=0.08]  [draw opacity=0] (8.93,-4.29) -- (0,0) -- (8.93,4.29) -- cycle    ;
%Shape: Circle [id:dp6517396229020851] 
\draw  [color={rgb, 255:red, 0; green, 0; blue, 0 }  ,draw opacity=1 ][fill={rgb, 255:red, 155; green, 155; blue, 155 }  ,fill opacity=1 ] (26,138.83) .. controls (39.81,138.83) and (51,150.03) .. (51,163.83) .. controls (51,177.64) and (39.81,188.83) .. (26,188.83) .. controls (12.19,188.83) and (1,177.64) .. (1,163.83) .. controls (1,150.03) and (12.19,138.83) .. (26,138.83) -- cycle ;
%Shape: Circle [id:dp5492031221840359] 
\draw  [color={rgb, 255:red, 0; green, 0; blue, 0 }  ,draw opacity=1 ][fill={rgb, 255:red, 155; green, 155; blue, 155 }  ,fill opacity=1 ] (106,138.83) .. controls (119.81,138.83) and (131,150.03) .. (131,163.83) .. controls (131,177.64) and (119.81,188.83) .. (106,188.83) .. controls (92.19,188.83) and (81,177.64) .. (81,163.83) .. controls (81,150.03) and (92.19,138.83) .. (106,138.83) -- cycle ;
%Straight Lines [id:da08418610682899774] 
\draw    (26,120) -- (26,135.83) ;
\draw [shift={(26,138.83)}, rotate = 270] [fill={rgb, 255:red, 0; green, 0; blue, 0 }  ][line width=0.08]  [draw opacity=0] (8.93,-4.29) -- (0,0) -- (8.93,4.29) -- cycle    ;
%Straight Lines [id:da2425220790319913] 
\draw    (106,120) -- (106,135.83) ;
\draw [shift={(106,138.83)}, rotate = 270] [fill={rgb, 255:red, 0; green, 0; blue, 0 }  ][line width=0.08]  [draw opacity=0] (8.93,-4.29) -- (0,0) -- (8.93,4.29) -- cycle    ;
%Shape: Square [id:dp6176699053164441] 
\draw   (1,1) -- (51,1) -- (51,51) -- (1,51) -- cycle ;
%Straight Lines [id:da13159972509792084] 
\draw    (26,51.17) -- (26,67) ;
\draw [shift={(26,70)}, rotate = 270] [fill={rgb, 255:red, 0; green, 0; blue, 0 }  ][line width=0.08]  [draw opacity=0] (8.93,-4.29) -- (0,0) -- (8.93,4.29) -- cycle    ;
%Shape: Square [id:dp5470773222078109] 
\draw   (81,0.8) -- (131,0.8) -- (131,50.8) -- (81,50.8) -- cycle ;
%Straight Lines [id:da07179638736976468] 
\draw    (105.4,50.6) -- (105.91,67) ;
\draw [shift={(106,70)}, rotate = 268.23] [fill={rgb, 255:red, 0; green, 0; blue, 0 }  ][line width=0.08]  [draw opacity=0] (8.93,-4.29) -- (0,0) -- (8.93,4.29) -- cycle    ;
%Straight Lines [id:da6579888118972566] 
\draw    (51,51) -- (68.4,64.8) ;
%Straight Lines [id:da5568435437123351] 
\draw    (68.4,64.8) -- (83.45,76.74) ;
\draw [shift={(85.8,78.6)}, rotate = 218.42000000000002] [fill={rgb, 255:red, 0; green, 0; blue, 0 }  ][line width=0.08]  [draw opacity=0] (8.93,-4.29) -- (0,0) -- (8.93,4.29) -- cycle    ;

% Text Node
\draw (17.4,85.67) node [anchor=north west][inner sep=0.75pt]  [color={rgb, 255:red, 0; green, 0; blue, 0 }  ,opacity=1 ] [align=left] {$\displaystyle z_{a}$};
% Text Node
\draw (97.4,85.67) node [anchor=north west][inner sep=0.75pt]  [color={rgb, 255:red, 0; green, 0; blue, 0 }  ,opacity=1 ] [align=left] {$\displaystyle z_{b}$};
% Text Node
\draw (17.4,154.5) node [anchor=north west][inner sep=0.75pt]  [color={rgb, 255:red, 0; green, 0; blue, 0 }  ,opacity=1 ] [align=left] {$\displaystyle x_{a}$};
% Text Node
\draw (97.4,154.5) node [anchor=north west][inner sep=0.75pt]  [color={rgb, 255:red, 0; green, 0; blue, 0 }  ,opacity=1 ] [align=left] {$\displaystyle x_{b}$};
% Text Node
\draw (18,16.2) node [anchor=north west][inner sep=0.75pt]  [color={rgb, 255:red, 0; green, 0; blue, 0 }  ,opacity=1 ] [align=left] {$\displaystyle \theta _{a}$};
% Text Node
\draw (97,16) node [anchor=north west][inner sep=0.75pt]  [color={rgb, 255:red, 0; green, 0; blue, 0 }  ,opacity=1 ] [align=left] {$\displaystyle \theta _{b}$};

\end{tikzpicture}
         }
    }
    \subfloat[\label{fig:detached_links_2}]{
        \resizebox{0.4\linewidth}{!}{
            \tikzset{every picture/.style={line width=0.75pt}} %set default line width to 0.75pt        

\begin{tikzpicture}[x=0.75pt,y=0.75pt,yscale=-1,xscale=1]
%uncomment if require: \path (0,194); %set diagram left start at 0, and has height of 194

%Shape: Circle [id:dp5551949184020153] 
\draw  [color={rgb, 255:red, 0; green, 0; blue, 0 }  ,draw opacity=1 ] (26,70) .. controls (39.81,70) and (51,81.19) .. (51,95) .. controls (51,108.81) and (39.81,120) .. (26,120) .. controls (12.19,120) and (1,108.81) .. (1,95) .. controls (1,81.19) and (12.19,70) .. (26,70) -- cycle ;
%Shape: Circle [id:dp3037910567541733] 
\draw  [color={rgb, 255:red, 0; green, 0; blue, 0 }  ,draw opacity=1 ] (106,70) .. controls (119.81,70) and (131,81.19) .. (131,95) .. controls (131,108.81) and (119.81,120) .. (106,120) .. controls (92.19,120) and (81,108.81) .. (81,95) .. controls (81,81.19) and (92.19,70) .. (106,70) -- cycle ;
%Straight Lines [id:da5152761660032619] 
\draw    (51,95) -- (66,95) ;
%Straight Lines [id:da817460181034247] 
\draw    (60,95) -- (78,95) ;
\draw [shift={(81,95)}, rotate = 180] [fill={rgb, 255:red, 0; green, 0; blue, 0 }  ][line width=0.08]  [draw opacity=0] (8.93,-4.29) -- (0,0) -- (8.93,4.29) -- cycle    ;
\draw [shift={(60,95)}, rotate = 0] [color={rgb, 255:red, 0; green, 0; blue, 0 }  ][line width=0.75]      (5.59,-5.59) .. controls (2.5,-5.59) and (0,-3.09) .. (0,0) .. controls (0,3.09) and (2.5,5.59) .. (5.59,5.59) ;
%Shape: Circle [id:dp052690014452144496] 
\draw  [color={rgb, 255:red, 0; green, 0; blue, 0 }  ,draw opacity=1 ][fill={rgb, 255:red, 155; green, 155; blue, 155 }  ,fill opacity=1 ] (26,138.83) .. controls (39.81,138.83) and (51,150.03) .. (51,163.83) .. controls (51,177.64) and (39.81,188.83) .. (26,188.83) .. controls (12.19,188.83) and (1,177.64) .. (1,163.83) .. controls (1,150.03) and (12.19,138.83) .. (26,138.83) -- cycle ;
%Shape: Circle [id:dp42963316786701844] 
\draw  [color={rgb, 255:red, 0; green, 0; blue, 0 }  ,draw opacity=1 ][fill={rgb, 255:red, 155; green, 155; blue, 155 }  ,fill opacity=1 ] (106,138.83) .. controls (119.81,138.83) and (131,150.03) .. (131,163.83) .. controls (131,177.64) and (119.81,188.83) .. (106,188.83) .. controls (92.19,188.83) and (81,177.64) .. (81,163.83) .. controls (81,150.03) and (92.19,138.83) .. (106,138.83) -- cycle ;
%Straight Lines [id:da6517301529358859] 
\draw    (26,120) -- (26,135.83) ;
\draw [shift={(26,138.83)}, rotate = 270] [fill={rgb, 255:red, 0; green, 0; blue, 0 }  ][line width=0.08]  [draw opacity=0] (8.93,-4.29) -- (0,0) -- (8.93,4.29) -- cycle    ;
%Straight Lines [id:da4307809646351546] 
\draw    (106,120) -- (106,135.83) ;
\draw [shift={(106,138.83)}, rotate = 270] [fill={rgb, 255:red, 0; green, 0; blue, 0 }  ][line width=0.08]  [draw opacity=0] (8.93,-4.29) -- (0,0) -- (8.93,4.29) -- cycle    ;
%Shape: Square [id:dp90023918838256] 
\draw   (1,1) -- (51,1) -- (51,51) -- (1,51) -- cycle ;
%Straight Lines [id:da024455643151067408] 
\draw    (26,51.17) -- (26,67) ;
\draw [shift={(26,70)}, rotate = 270] [fill={rgb, 255:red, 0; green, 0; blue, 0 }  ][line width=0.08]  [draw opacity=0] (8.93,-4.29) -- (0,0) -- (8.93,4.29) -- cycle    ;
%Shape: Square [id:dp17326933726057114] 
\draw   (81,0.8) -- (131,0.8) -- (131,50.8) -- (81,50.8) -- cycle ;
%Straight Lines [id:da3706383085196767] 
\draw    (105.4,50.6) -- (105.91,67) ;
\draw [shift={(106,70)}, rotate = 268.23] [fill={rgb, 255:red, 0; green, 0; blue, 0 }  ][line width=0.08]  [draw opacity=0] (8.93,-4.29) -- (0,0) -- (8.93,4.29) -- cycle    ;
%Straight Lines [id:da6219881401099487] 
\draw    (51,51) -- (68.4,64.8) ;
\draw [shift={(68.4,64.8)}, rotate = 38.42] [color={rgb, 255:red, 0; green, 0; blue, 0 }  ][line width=0.75]      (5.59,-5.59) .. controls (2.5,-5.59) and (0,-3.09) .. (0,0) .. controls (0,3.09) and (2.5,5.59) .. (5.59,5.59) ;
%Straight Lines [id:da3022589798657769] 
\draw    (68.4,64.8) -- (83.45,76.74) ;
\draw [shift={(85.8,78.6)}, rotate = 218.42000000000002] [fill={rgb, 255:red, 0; green, 0; blue, 0 }  ][line width=0.08]  [draw opacity=0] (8.93,-4.29) -- (0,0) -- (8.93,4.29) -- cycle    ;
%Straight Lines [id:da06964238912638954] 
\draw [color={rgb, 255:red, 0; green, 0; blue, 0 }  ,draw opacity=0.5 ] [dash pattern={on 4.5pt off 4.5pt}]  (64.8,1.01) -- (65,188.61) ;

% Text Node
\draw (17.4,85.67) node [anchor=north west][inner sep=0.75pt]  [color={rgb, 255:red, 0; green, 0; blue, 0 }  ,opacity=1 ] [align=left] {$\displaystyle z_{a}$};
% Text Node
\draw (97.4,85.67) node [anchor=north west][inner sep=0.75pt]  [color={rgb, 255:red, 0; green, 0; blue, 0 }  ,opacity=1 ] [align=left] {$\displaystyle z_{b}$};
% Text Node
\draw (17.4,154.5) node [anchor=north west][inner sep=0.75pt]  [color={rgb, 255:red, 0; green, 0; blue, 0 }  ,opacity=1 ] [align=left] {$\displaystyle x_{a}$};
% Text Node
\draw (97.4,154.5) node [anchor=north west][inner sep=0.75pt]  [color={rgb, 255:red, 0; green, 0; blue, 0 }  ,opacity=1 ] [align=left] {$\displaystyle x_{b}$};
% Text Node
\draw (17,16.2) node [anchor=north west][inner sep=0.75pt]  [color={rgb, 255:red, 0; green, 0; blue, 0 }  ,opacity=1 ] [align=left] {$\displaystyle \theta _{a}$};
% Text Node
\draw (97,16) node [anchor=north west][inner sep=0.75pt]  [color={rgb, 255:red, 0; green, 0; blue, 0 }  ,opacity=1 ] [align=left] {$\displaystyle \theta _{b}$};

\end{tikzpicture}
         }
    }
    \caption{Two generative flow graphs representations of a simple model with two parameters, two latent variables, and two observed variables. In both graphs $\theta_a$ and $z_a$ are needed to generate $z_b$, however, in (b) we have explicitly constrained the inference of $z_b$ to not influence the learning of $\theta_a$ and inference of $z_a$. Thus, the evidence provided by $x_b$ neither is allowed to have an influence on $\theta_a$ and $z_a$. Thereby, the model represented by the nodes on the left-hand side of the dashed line in (b) can be seen as an independent problem.}
    \label{fig:detached_links}
\end{figure}
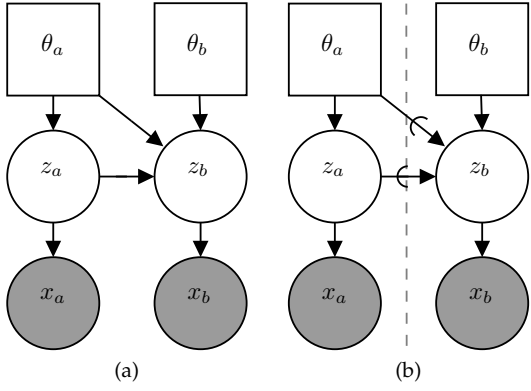

Besides the possibility of expressing a factorization of the joint prior distribution, the detached link allows us to express additional structure for the posterior distribution, $p(z|x)$. Consider the two generative flow graphs in \cref{fig:detached_links}. By applying standard manipulations we can obtain the factorization in \cref{eq:detached_1} for the graph in \cref{fig:detached_links_1}.
\begin{align}
 & p_{\theta _{a} ,\theta _{b}}( z_{a} ,z_{b} |x_{a} ,x_{b}) \nonumber\\
 & \quad =p_{\theta _{a} ,\theta _{b}}( z_{b} |z_{a} ,x_{a} ,x_{b}) p_{\theta _{a} ,\theta _{b}}( z_{a} |x_{a} ,x_{b}) \nonumber\\
 & \quad =p_{\theta _{a} ,\theta _{b}}( z_{b} |z_{a} ,x_{b}) p_{\theta _{a} ,\theta _{b}}( z_{a} |x_{a} ,x_{b}) \label{eq:detached_1}
\end{align}
wheres from the definition of the detached link we can write the factorization in \cref{eq:detached_2} for the graph in \cref{fig:detached_links_2}.
\begin{align}
 & p_{\theta _{a} ,\theta _{b}}( z_{a} ,z_{b} |x_{a} ,x_{b}) \nonumber\\
 & \quad =p_{\breve{\theta }_{a} ,\theta _{b}}( z_{b} |\breve{z}_{a} ,\breve{x}_{a} ,x_{b}) p_{\theta _{a}}( z_{a} |x_{a}) \nonumber\\
 & \quad =p_{\breve{\theta }_{a} ,\theta _{b}}( z_{b} |\breve{z}_{a} ,x_{b}) p_{\theta _{a}}( z_{a} |x_{a}) \label{eq:detached_2}
\end{align}
%
%Where we added a breve over $\theta_1$ to emphasize that $\theta_1$ can be considered a constant in that factor.
The main difference between these two factorizations is the distribution over the latent variable $z_{a}$. In \cref{eq:detached_1} the distribution over the latent variable $z_{a}$ depends on the evidence provided by both observations $x_{a}$ and $x_{b}$, and is influenced by both parameters $\theta_{a}$ and $\theta_{b}$. In \cref{eq:detached_2} the distribution over $z_{a}$ depends only on the evidence provided by the observations $x_{a}$, and is only influenced by the parameter $\theta_{a}$. As such the inference problem of obtaining the posterior distribution over $z_a$ is independent of the inference problem of obtaining the posterior distribution over $z_b$, but not conversely. In general, for model consisting of $\overline{a}$ node collections, $C^{( a)} =\left\{Z^{( a)} ;X^{( a)} ;\Theta ^{( a)}\right\}$, connected only by detached links we can write the factorization of the posterior as
\begin{align}
    p_{\Theta } (Z|X)=\prod _{a=1}^{\overline{a}} p_{\Theta ^{\{a\}} ,Pa\breve{\Theta }\left( C^{\{a\}}\right)}\left( Z^{\{a\}} |Pa\breve{Z}\left( C^{\{a\}}\right) ,X^{\{a\}}\right) \label{eq:collections_factorization}
\end{align}
where %$AnV\left( C^{\{a\}}\right) =\left\{AnZ\left( C^{\{a\}}\right) ,AnX\left( C^{\{a\}}\right)\right\}$ denotes the set of latent and observed variables ancestral to the node collection, $C^{\{a\}}$, and 
the breves are used to emphasize that the variables and parameters are related through a detached link. The possible benefit of being able to express such structure will become clear in \cref{sec:inference}.

Another added benefit of the generative flow graph representation is to express models by different levels of abstraction. As an example consider the three different factorization of the simultaneous localisation and mapping problem given in \cref{eq:slam_fact_1}, \cref{eq:slam_fact_2}, and \cref{eq:slam_fact_3}.

\begin{align}
& p\left( z_{s}^{\{0;t\}} ,z_{a}^{\{0;t-1\}} ,x_{p}^{\{1;t\}} ,z_{m}^{\{0;I\}}\right)\nonumber \\
& \quad =p\left( z_{s}^{\{0;t\}} ,z_{a}^{\{0;t-1\}}, x_{p}^{\{1;t\}} |z_{s}^{\{0\}}, z_{m}^{\{0;I\}}\right) \label{eq:slam_fact_1}\\
& \quad\quad\quad \cdot p\left( z_{s}^{\{0\}}\right)\prod _{i=1}^{I} p\left( z_{m}^{\{i\}}\right)\nonumber\\
%
%& \quad =p\left( x_{p}^{\{1;t\}} |z_{s}^{\{0;t\}} ,z_{a}^{\{0;t-1\}} ,z_{m}^{\{0;I\}}\right)\label{eq:slam_fact_1}\\
%
%& \quad\quad\quad \cdot p\left( z_{s}^{\{0;t\}} ,z_{a}^{\{0;t-1\}} |z_{s}^{\{0\}}\right) p\left( z_{s}^{\{0\}}\right)\prod _{i=1}^{I} p\left( z_{m}^{\{i\}}\right)\nonumber \\
%
& \quad =p\left( z_{s}^{\{0\}}\right)\prod _{i=1}^{I} p\left( z_{m}^{\{i\}}\right)\label{eq:slam_fact_2}\\
& \quad\quad \quad \cdot \prod _{\tau =1}^{t}\left(\begin{matrix*}[l]
p\left( x_{p}^{\{\tau \}} |z_{s}^{\{\tau \}} ,z_{a}^{\{\tau -1\}} ,z_{m}^{\{0;I\}}\right)\\
\quad \cdot p\left( z_{s}^{\{\tau \}} ,z_{a}^{\{\tau -1\}} |z_{s}^{\{\tau -1\}}\right)
\end{matrix*}\right)\nonumber  \\
& \quad =p\left( z_{s}^{\{0\}}\right)\prod _{i=1}^{I} p\left( z_{m}^{\{i\}}\right)\label{eq:slam_fact_3}\\
& \quad\quad \quad \cdot \prod _{\tau =1}^{t}\left(\begin{matrix*}[l]
p\left( x_{p}^{\{\tau \}} |z_{s}^{\{\tau \}} ,z_{a}^{\{\tau -1\}} ,z_{m}^{\{0;I\}}\right)\\
\quad \cdot p\left( z_{s}^{\{\tau \}} |z_{a}^{\{\tau -1\}} ,z_{s}^{\{\tau -1\}}\right) p\left( z_{a}^{\{\tau -1\}}\right)
\end{matrix*}\right)\nonumber 
\end{align}

The generative flow graphs in \cref{fig:slam_graph_idiom_new_1}, \cref{fig:slam_graph_idiom_new_2}, and \cref{fig:slam_graph_idiom_new_3}
corresponds directly to the factorization in \cref{eq:slam_fact_3}, \cref{eq:slam_fact_2}, and \cref{eq:slam_fact_1}, respectively. Thereby, they represents different levels of abstractions for the same model. As such generative flow graphs simply yield better expressibility over their directed graphical model counterparts. % In the next section we will discuss how generative flow graphs can be used as a tool for discovering probabilistic programming idioms.

\subsection{Probabilistic Programming Idioms} \label{sec:ppl_idioms}

%\begin{figure}
%     \centering
%     \begin{subfigure}[tb]{0.5\linewidth}
%         \centering
%         \resizebox{1.0\linewidth}{!}{
%            \input{tikz/markov_decision_process_graph_idiom_new_1.tikz}
%         }
%         \caption{}
%         \label{fig:MDP_graph_idiom_new_1}
%     \end{subfigure}%
%     \begin{subfigure}[tb]{0.5\linewidth}
%         \centering
%         \resizebox{1.0\linewidth}{!}{
%            \input{tikz/markov_decision_process_graph_idiom_new_2.tikz}
%         }
%         \caption{}
%         \label{fig:MDP_graph_idiom_new_2}
%     \end{subfigure}
%     \caption{Two semantically equivalent generative flow graphs corresponding to the directed graphical model in \cref{fig:markov_decision_process_graph_idiom}.}
%     \label{fig:MDP_graph_idiom_new}
%\end{figure}

\begin{figure}
    \centering
    \subfloat[\label{fig:MDP_graph_idiom_new_1}]{
        \resizebox{0.45\linewidth}{!}{
            \tikzset{every picture/.style={line width=0.75pt}} %set default line width to 0.75pt        

\begin{tikzpicture}[x=0.75pt,y=0.75pt,yscale=-1,xscale=1]
%uncomment if require: \path (0,283); %set diagram left start at 0, and has height of 283

%Shape: Circle [id:dp22720068943405614] 
\draw  [color={rgb, 255:red, 0; green, 0; blue, 0 }  ,draw opacity=1 ] (57.47,115.07) .. controls (71.27,115.07) and (82.47,126.26) .. (82.47,140.07) .. controls (82.47,153.87) and (71.27,165.07) .. (57.47,165.07) .. controls (43.66,165.07) and (32.47,153.87) .. (32.47,140.07) .. controls (32.47,126.26) and (43.66,115.07) .. (57.47,115.07) -- cycle ;
%Shape: Circle [id:dp037495091996808894] 
\draw  [color={rgb, 255:red, 0; green, 0; blue, 0 }  ,draw opacity=1 ] (106.13,79.2) .. controls (119.94,79.2) and (131.13,90.39) .. (131.13,104.2) .. controls (131.13,118.01) and (119.94,129.2) .. (106.13,129.2) .. controls (92.33,129.2) and (81.13,118.01) .. (81.13,104.2) .. controls (81.13,90.39) and (92.33,79.2) .. (106.13,79.2) -- cycle ;
%Straight Lines [id:da14623158646847134] 
\draw [color={rgb, 255:red, 0; green, 0; blue, 0 }  ,draw opacity=1 ]   (106.13,129.2) -- (106,139.96) -- (105.54,176.03) ;
\draw [shift={(105.5,179.03)}, rotate = 270.73] [fill={rgb, 255:red, 0; green, 0; blue, 0 }  ,fill opacity=1 ][line width=0.08]  [draw opacity=0] (8.93,-4.29) -- (0,0) -- (8.93,4.29) -- cycle    ;
%Straight Lines [id:da8215244897391389] 
\draw [color={rgb, 255:red, 0; green, 0; blue, 0 }  ,draw opacity=1 ]   (58.27,52.27) -- (58,70.3) ;
\draw [shift={(58,70.3)}, rotate = 90.85] [color={rgb, 255:red, 0; green, 0; blue, 0 }  ,draw opacity=1 ][fill={rgb, 255:red, 0; green, 0; blue, 0 }  ,fill opacity=1 ][line width=0.75]      (0, 0) circle [x radius= 3.35, y radius= 3.35]   ;
\draw [shift={(58.27,52.27)}, rotate = 90.85] [color={rgb, 255:red, 0; green, 0; blue, 0 }  ,draw opacity=1 ][line width=0.75]      (5.59,-5.59) .. controls (2.5,-5.59) and (0,-3.09) .. (0,0) .. controls (0,3.09) and (2.5,5.59) .. (5.59,5.59) ;
%Straight Lines [id:da7559562186681266] 
\draw [color={rgb, 255:red, 0; green, 0; blue, 0 }  ,draw opacity=1 ]   (58,70.3) -- (57.5,112.07) ;
\draw [shift={(57.47,115.07)}, rotate = 270.68] [fill={rgb, 255:red, 0; green, 0; blue, 0 }  ,fill opacity=1 ][line width=0.08]  [draw opacity=0] (8.93,-4.29) -- (0,0) -- (8.93,4.29) -- cycle    ;
%Straight Lines [id:da6276634129410894] 
\draw    (57.47,165.07) -- (57.22,198.83) ;
\draw [shift={(57.2,201.83)}, rotate = 270.42] [fill={rgb, 255:red, 0; green, 0; blue, 0 }  ][line width=0.08]  [draw opacity=0] (8.93,-4.29) -- (0,0) -- (8.93,4.29) -- cycle    ;
%Shape: Circle [id:dp786906927650658] 
\draw  [fill={rgb, 255:red, 155; green, 155; blue, 155 }  ,fill opacity=0.5 ] (105.5,179.03) .. controls (119.31,179.03) and (130.5,190.23) .. (130.5,204.03) .. controls (130.5,217.84) and (119.31,229.03) .. (105.5,229.03) .. controls (91.69,229.03) and (80.5,217.84) .. (80.5,204.03) .. controls (80.5,190.23) and (91.69,179.03) .. (105.5,179.03) -- cycle ;
%Shape: Diamond [id:dp021697903569183774] 
\draw   (92.2,236.83) -- (57.2,271.83) -- (22.2,236.83) -- (57.2,201.83) -- cycle ;
%Shape: Circle [id:dp4854781845132683] 
\draw  [color={rgb, 255:red, 0; green, 0; blue, 0 }  ,draw opacity=1 ] (58.27,2.27) .. controls (72.07,2.27) and (83.27,13.46) .. (83.27,27.27) .. controls (83.27,41.07) and (72.07,52.27) .. (58.27,52.27) .. controls (44.46,52.27) and (33.27,41.07) .. (33.27,27.27) .. controls (33.27,13.46) and (44.46,2.27) .. (58.27,2.27) -- cycle ;
%Straight Lines [id:da9918150324994723] 
\draw    (22.2,236.83) -- (14.67,237.11) -- (15.67,70.44) -- (55,70.31) ;
\draw [shift={(58,70.3)}, rotate = 539.81] [fill={rgb, 255:red, 0; green, 0; blue, 0 }  ][line width=0.08]  [draw opacity=0] (8.93,-4.29) -- (0,0) -- (8.93,4.29) -- cycle    ;
%Shape: Rectangle [id:dp5499666234490392] 
\draw  [color={rgb, 255:red, 66; green, 133; blue, 244 }  ,draw opacity=1 ] (8.2,67.02) .. controls (8.2,64.26) and (10.44,62.02) .. (13.2,62.02) -- (139,62.02) .. controls (141.76,62.02) and (144,64.26) .. (144,67.02) -- (144,271.77) .. controls (144,274.53) and (141.76,276.77) .. (139,276.77) -- (13.2,276.77) .. controls (10.44,276.77) and (8.2,274.53) .. (8.2,271.77) -- cycle ;
%Straight Lines [id:da7216659273496056] 
\draw [color={rgb, 255:red, 0; green, 0; blue, 0 }  ,draw opacity=1 ]   (106,139.96) -- (85.47,140.05) ;
\draw [shift={(82.47,140.07)}, rotate = 359.73] [fill={rgb, 255:red, 0; green, 0; blue, 0 }  ,fill opacity=1 ][line width=0.08]  [draw opacity=0] (8.93,-4.29) -- (0,0) -- (8.93,4.29) -- cycle    ;
%Shape: Rectangle [id:dp17192635152798363] 
\draw  [color={rgb, 255:red, 52; green, 168; blue, 83 }  ,draw opacity=1 ] (28.2,80.96) .. controls (28.2,78.19) and (30.44,75.96) .. (33.2,75.96) -- (129,75.96) .. controls (131.76,75.96) and (134,78.19) .. (134,80.96) -- (134,163.27) .. controls (134,166.03) and (131.76,168.27) .. (129,168.27) -- (33.2,168.27) .. controls (30.44,168.27) and (28.2,166.03) .. (28.2,163.27) -- cycle ;
%Straight Lines [id:da4253597102768396] 
\draw    (73.43,159.36) -- (88.53,180.21) ;
\draw [shift={(90.29,182.64)}, rotate = 234.1] [fill={rgb, 255:red, 0; green, 0; blue, 0 }  ][line width=0.08]  [draw opacity=0] (8.93,-4.29) -- (0,0) -- (8.93,4.29) -- cycle    ;

% Text Node
\draw (86.7,95.3) node [anchor=north west][inner sep=0.75pt]  [color={rgb, 255:red, 0; green, 0; blue, 0 }  ,opacity=1 ] [align=left] {$\displaystyle z_{a}^{\{\tau -1\}}$};
% Text Node
\draw (44.53,131.73) node [anchor=north west][inner sep=0.75pt]  [color={rgb, 255:red, 0; green, 0; blue, 0 }  ,opacity=1 ] [align=left] {$\displaystyle z_{s}^{\{\tau \}}$};
% Text Node
\draw (91.5,193.7) node [anchor=north west][inner sep=0.75pt]   [align=left] {$\displaystyle x_{O}^{\{\tau \}}$};
% Text Node
\draw (49.17,18.43) node [anchor=north west][inner sep=0.75pt]  [color={rgb, 255:red, 0; green, 0; blue, 0 }  ,opacity=1 ] [align=left] {$\displaystyle z_{s}^{\{t\}}$};
% Text Node
\draw (36.5,228.7) node [anchor=north west][inner sep=0.75pt]   [align=left] {$\displaystyle \tau =T$};
% Text Node
\draw (11.8,240.07) node [anchor=north west][inner sep=0.75pt]   [align=left] {F};
% Text Node
\draw (106.52,265.76) node   [align=left] {\begin{minipage}[lt]{58.91pt}\setlength\topsep{0pt}
{\scriptsize $\displaystyle \tau \in \{t+1,...,T\}$}
\end{minipage}};

\end{tikzpicture}
         }
    }
    \subfloat[\label{fig:MDP_graph_idiom_new_2}]{
        \resizebox{0.45\linewidth}{!}{
            \tikzset{every picture/.style={line width=0.75pt}} %set default line width to 0.75pt        

\begin{tikzpicture}[x=0.75pt,y=0.75pt,yscale=-1,xscale=1]
%uncomment if require: \path (0,238); %set diagram left start at 0, and has height of 238

%Straight Lines [id:da9435434393703401] 
\draw [color={rgb, 255:red, 0; green, 0; blue, 0 }  ,draw opacity=1 ]   (58.27,51.27) -- (58,66.3) ;
\draw [shift={(58,66.3)}, rotate = 91.02] [color={rgb, 255:red, 0; green, 0; blue, 0 }  ,draw opacity=1 ][fill={rgb, 255:red, 0; green, 0; blue, 0 }  ,fill opacity=1 ][line width=0.75]      (0, 0) circle [x radius= 3.35, y radius= 3.35]   ;
\draw [shift={(58.27,51.27)}, rotate = 91.02] [color={rgb, 255:red, 0; green, 0; blue, 0 }  ,draw opacity=1 ][line width=0.75]      (5.59,-5.59) .. controls (2.5,-5.59) and (0,-3.09) .. (0,0) .. controls (0,3.09) and (2.5,5.59) .. (5.59,5.59) ;
%Straight Lines [id:da5987510591414473] 
\draw    (57.2,113.76) -- (57.2,156.83) ;
\draw [shift={(57.2,159.83)}, rotate = 270] [fill={rgb, 255:red, 0; green, 0; blue, 0 }  ][line width=0.08]  [draw opacity=0] (8.93,-4.29) -- (0,0) -- (8.93,4.29) -- cycle    ;
%Shape: Circle [id:dp8302075610240829] 
\draw  [fill={rgb, 255:red, 155; green, 155; blue, 155 }  ,fill opacity=0.5 ] (106.5,139.03) .. controls (120.31,139.03) and (131.5,150.23) .. (131.5,164.03) .. controls (131.5,177.84) and (120.31,189.03) .. (106.5,189.03) .. controls (92.69,189.03) and (81.5,177.84) .. (81.5,164.03) .. controls (81.5,150.23) and (92.69,139.03) .. (106.5,139.03) -- cycle ;
%Shape: Diamond [id:dp31321836909221634] 
\draw   (92.2,194.83) -- (57.2,229.83) -- (22.2,194.83) -- (57.2,159.83) -- cycle ;
%Shape: Circle [id:dp5872730104905186] 
\draw  [color={rgb, 255:red, 0; green, 0; blue, 0 }  ,draw opacity=1 ] (58.27,1.27) .. controls (72.07,1.27) and (83.27,12.46) .. (83.27,26.27) .. controls (83.27,40.07) and (72.07,51.27) .. (58.27,51.27) .. controls (44.46,51.27) and (33.27,40.07) .. (33.27,26.27) .. controls (33.27,12.46) and (44.46,1.27) .. (58.27,1.27) -- cycle ;
%Straight Lines [id:da41635306781776893] 
\draw    (22.2,194.83) -- (13.67,195.11) -- (14.2,66.63) -- (55,66.32) ;
\draw [shift={(58,66.3)}, rotate = 539.5699999999999] [fill={rgb, 255:red, 0; green, 0; blue, 0 }  ][line width=0.08]  [draw opacity=0] (8.93,-4.29) -- (0,0) -- (8.93,4.29) -- cycle    ;
%Shape: Rectangle [id:dp6774998091591842] 
\draw  [color={rgb, 255:red, 66; green, 133; blue, 244 }  ,draw opacity=1 ] (8.2,64.83) .. controls (8.2,62.07) and (10.44,59.83) .. (13.2,59.83) -- (133.4,59.83) .. controls (136.16,59.83) and (138.4,62.07) .. (138.4,64.83) -- (138.4,231.86) .. controls (138.4,234.63) and (136.16,236.86) .. (133.4,236.86) -- (13.2,236.86) .. controls (10.44,236.86) and (8.2,234.63) .. (8.2,231.86) -- cycle ;
%Shape: Rectangle [id:dp7669520497749791] 
\draw  [color={rgb, 255:red, 52; green, 168; blue, 83 }  ,draw opacity=1 ] (17.2,84.77) .. controls (17.2,82) and (19.44,79.77) .. (22.2,79.77) -- (118,79.77) .. controls (120.76,79.77) and (123,82) .. (123,84.77) -- (123,108.96) .. controls (123,111.72) and (120.76,113.96) .. (118,113.96) -- (22.2,113.96) .. controls (19.44,113.96) and (17.2,111.72) .. (17.2,108.96) -- cycle ;
%Straight Lines [id:da043455889843920126] 
\draw    (58,66.3) -- (58,76.38) ;
\draw [shift={(58,79.38)}, rotate = 270] [fill={rgb, 255:red, 0; green, 0; blue, 0 }  ][line width=0.08]  [draw opacity=0] (8.93,-4.29) -- (0,0) -- (8.93,4.29) -- cycle    ;
%Straight Lines [id:da20105932970535267] 
\draw    (106.4,114.16) -- (106.49,136.03) ;
\draw [shift={(106.5,139.03)}, rotate = 269.77] [fill={rgb, 255:red, 0; green, 0; blue, 0 }  ][line width=0.08]  [draw opacity=0] (8.93,-4.29) -- (0,0) -- (8.93,4.29) -- cycle    ;
%Straight Lines [id:da2098056700812021] 
\draw    (57.43,130.17) -- (83.22,147.37) ;
\draw [shift={(85.71,149.03)}, rotate = 213.69] [fill={rgb, 255:red, 0; green, 0; blue, 0 }  ][line width=0.08]  [draw opacity=0] (8.93,-4.29) -- (0,0) -- (8.93,4.29) -- cycle    ;

% Text Node
\draw (19.2,82.77) node [anchor=north west][inner sep=0.75pt]  [color={rgb, 255:red, 0; green, 0; blue, 0 }  ,opacity=1 ] [align=left] {$\displaystyle \left\{z_{s}^{\{\tau \}} ,z_{a}^{\{\tau -1\}} ;;\right\}$};
% Text Node
\draw (93.17,153.7) node [anchor=north west][inner sep=0.75pt]   [align=left] {$\displaystyle x_{O}^{\{\tau \}}$};
% Text Node
\draw (46.17,17.43) node [anchor=north west][inner sep=0.75pt]  [color={rgb, 255:red, 0; green, 0; blue, 0 }  ,opacity=1 ] [align=left] {$\displaystyle z_{s}^{\{t\}}$};
% Text Node
\draw (35.5,186.7) node [anchor=north west][inner sep=0.75pt]   [align=left] {$\displaystyle \tau =T$};
% Text Node
\draw (11.2,198.07) node [anchor=north west][inner sep=0.75pt]   [align=left] {F};
% Text Node
\draw (99.95,227.16) node   [align=left] {\begin{minipage}[lt]{56.51pt}\setlength\topsep{0pt}
{\scriptsize $\displaystyle \tau \in \{t+1,...,T\}$}
\end{minipage}};
% Text Node
\draw (33.86,114.19) node [anchor=north west][inner sep=0.75pt]  [color={rgb, 255:red, 0; green, 0; blue, 0 }  ,opacity=1 ] [align=left] {{\footnotesize $\displaystyle z_{s}^{\{\tau \}}$}};
% Text Node
\draw (73,114.91) node [anchor=north west][inner sep=0.75pt]  [color={rgb, 255:red, 0; green, 0; blue, 0 }  ,opacity=1 ] [align=left] {{\footnotesize $\displaystyle z_{a}^{\{\tau -1\}}$}};

\end{tikzpicture}
         }
    }
    \caption{Two semantically equivalent generative flow graphs corresponding to the directed graphical model in \cref{fig:markov_decision_process_graph_idiom}.}
    \label{fig:MDP_graph_idiom_new}
\end{figure}
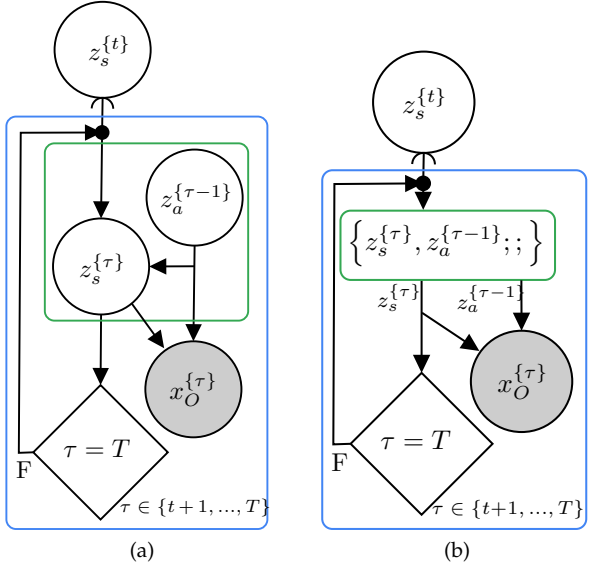

We have already discussed how probabilistic programming idioms can be seen as a means to achieve functional elegance. In this section, we describe how such idioms can be discovered by inspecting generative flow graphs. We define probabilistic programming idioms as follows:\\

\textit{“Probabilistic programming idioms are reusable code fragments of probabilistic programs sharing an equivalent semantic role in their enclosing probabilistic programs.”}\\

To identify such probabilistic programming idioms, we can look for node collections containing the same nodes and with the same internal structure in at least two different probabilistic programs. Consider for example the node collection $\left\{z_{s}^{\{\tau \}} ,z_{a}^{\{\tau -1\}} ;;\right\}$ highlighted with a green border in the generative flow graph for both the simultaneous localization and mapping problem and Markov Decision Process depicted in \cref{fig:slam_graph_idiom_new} and \cref{fig:MDP_graph_idiom_new}, respectively. From \cref{fig:slam_graph_idiom_new_1} and \cref{fig:MDP_graph_idiom_new_1} it is clear that the internal structure of this node collection is identical in both graphs, and that it represents the factorization
\begin{equation*}
p\left( z_{s}^{\{\tau \}} |z_{s}^{\{\tau -1\}} ,z_{a}^{\{\tau -1\}}\right) p\left( z_{a}^{\{\tau-1 \}}\right).
\end{equation*}
Assuming that the distributions $p\left( z_{s}^{\{\tau \}} |z_{s}^{\{\tau-1 \}} ,z_{a}^{\{\tau -1\}}\right)$ and $p\left( z_{a}^{\{\tau-1 \}}\right)$ are the same in both models, we could possible create a probabilistic program for this node collection once, and then reuse it in both models. This probabilistic program should then take a sample $z_{s}^{\{\tau-1 \}}$ as input. From this input the program could sample both $z_{s}^{\{\tau \}}$ and $z_{a}^{\{\tau-1 \}}$ from "hard-coded" distributions $p\left( z_{s}^{\{\tau \}} |z_{s}^{\{\tau-1 \}} ,z_{a}^{\{\tau-1 \}}\right)$ and $p\left( z_{a}^{\{\tau-1 \}}\right)$ using the sample function or keyword of the probabilistic programming language. Finally, the program should return both of these samples. While this approach might work perfectly for some applications, the two distributions $p\left( z_{s}^{\{\tau-1 \}}|z_{s}^{\{\tau-1 \}} ,z_{a}^{\{\tau -1\}}\right)$ and $p\left( z_{a}^{\{\tau -1\}}\right)$ are usually application specific, limiting the usability for an idiom in which they are "hard-coded". A far more general approach would be to allow the probabilistic program to instead take the two distributions as input or having these distributions as free variables, allowing us to re-use the code fragment even for problems where these distributions are not necessarily the same. Rather than fully defining a model of a part of cognition, such a probabilistic program would constitute a template method for the generative flow of that part of cognition. Specific utilization of the model could then be done via a function closure specifying the free distributions. While the benefits of the above example arguably might be limited since the internal structure of the node collection is relatively simple, it is not hard to imagine more complex structures. Consider for instance the node collection highlighted with a blue border in \cref{fig:MDP_graph_idiom_new}. By constructing an appropriate probabilistic program for this node collection we have defined a probabilistic programming idiom constituting the foundation for optimal control and reinforcement learning. % which is widely applicable.

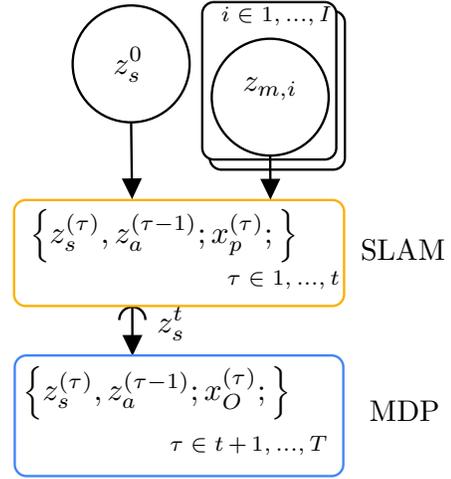
\begin{figure}[!t]
\centering
\resizebox{0.7\linewidth}{!}{
    \tikzset{every picture/.style={line width=0.75pt}} %set default line width to 0.75pt        

\begin{tikzpicture}[x=0.75pt,y=0.75pt,yscale=-1,xscale=1]
%uncomment if require: \path (0,212); %set diagram left start at 0, and has height of 212

%Shape: Rectangle [id:dp6107759749613908] 
\draw   (87.37,11.03) .. controls (87.37,8.27) and (89.61,6.03) .. (92.37,6.03) -- (140.17,6.03) .. controls (142.93,6.03) and (145.17,8.27) .. (145.17,11.03) -- (145.17,68.91) .. controls (145.17,71.67) and (142.93,73.91) .. (140.17,73.91) -- (92.37,73.91) .. controls (89.61,73.91) and (87.37,71.67) .. (87.37,68.91) -- cycle ;
%Shape: Rectangle [id:dp312922353619834] 
\draw  [fill={rgb, 255:red, 255; green, 255; blue, 255 }  ,fill opacity=1 ] (84.17,7.23) .. controls (84.17,4.47) and (86.41,2.23) .. (89.17,2.23) -- (136.5,2.23) .. controls (139.26,2.23) and (141.5,4.47) .. (141.5,7.23) -- (141.5,64.58) .. controls (141.5,67.34) and (139.26,69.58) .. (136.5,69.58) -- (89.17,69.58) .. controls (86.41,69.58) and (84.17,67.34) .. (84.17,64.58) -- cycle ;
%Straight Lines [id:da5415418044970761] 
\draw [color={rgb, 255:red, 0; green, 0; blue, 0 }  ,draw opacity=1 ]   (54.27,132.67) -- (54.32,150.61) ;
\draw [shift={(54.33,153.61)}, rotate = 269.82] [fill={rgb, 255:red, 0; green, 0; blue, 0 }  ,fill opacity=1 ][line width=0.08]  [draw opacity=0] (8.93,-4.29) -- (0,0) -- (8.93,4.29) -- cycle    ;
\draw [shift={(54.27,132.67)}, rotate = 89.82] [color={rgb, 255:red, 0; green, 0; blue, 0 }  ,draw opacity=1 ][line width=0.75]      (5.59,-5.59) .. controls (2.5,-5.59) and (0,-3.09) .. (0,0) .. controls (0,3.09) and (2.5,5.59) .. (5.59,5.59) ;
%Shape: Circle [id:dp6238944418971979] 
\draw  [color={rgb, 255:red, 0; green, 0; blue, 0 }  ,draw opacity=1 ] (53.77,3.66) .. controls (67.57,3.66) and (78.77,14.85) .. (78.77,28.66) .. controls (78.77,42.46) and (67.57,53.66) .. (53.77,53.66) .. controls (39.96,53.66) and (28.77,42.46) .. (28.77,28.66) .. controls (28.77,14.85) and (39.96,3.66) .. (53.77,3.66) -- cycle ;
%Shape: Rectangle [id:dp2236048112247897] 
\draw  [color={rgb, 255:red, 251; green, 176; blue, 5 }  ,draw opacity=1 ] (3.7,91.92) .. controls (3.7,89.16) and (5.94,86.92) .. (8.7,86.92) -- (139.83,86.92) .. controls (142.59,86.92) and (144.83,89.16) .. (144.83,91.92) -- (144.83,127.25) .. controls (144.83,130.02) and (142.59,132.25) .. (139.83,132.25) -- (8.7,132.25) .. controls (5.94,132.25) and (3.7,130.02) .. (3.7,127.25) -- cycle ;
%Shape: Circle [id:dp37452907284245307] 
\draw  [color={rgb, 255:red, 0; green, 0; blue, 0 }  ,draw opacity=1 ] (113.23,17.89) .. controls (127.04,17.89) and (138.23,29.08) .. (138.23,42.89) .. controls (138.23,56.7) and (127.04,67.89) .. (113.23,67.89) .. controls (99.43,67.89) and (88.23,56.7) .. (88.23,42.89) .. controls (88.23,29.08) and (99.43,17.89) .. (113.23,17.89) -- cycle ;
%Straight Lines [id:da5884719609103972] 
\draw    (53.77,53.66) -- (54.13,83.59) ;
\draw [shift={(54.17,86.59)}, rotate = 269.3] [fill={rgb, 255:red, 0; green, 0; blue, 0 }  ][line width=0.08]  [draw opacity=0] (8.93,-4.29) -- (0,0) -- (8.93,4.29) -- cycle    ;
%Straight Lines [id:da1287318364262493] 
\draw    (113.23,67.89) -- (113.32,83.77) ;
\draw [shift={(113.33,86.77)}, rotate = 269.7] [fill={rgb, 255:red, 0; green, 0; blue, 0 }  ][line width=0.08]  [draw opacity=0] (8.93,-4.29) -- (0,0) -- (8.93,4.29) -- cycle    ;
%Shape: Rectangle [id:dp38037247999190704] 
\draw  [color={rgb, 255:red, 66; green, 133; blue, 244 }  ,draw opacity=1 ] (4,158.43) .. controls (4,155.67) and (6.24,153.43) .. (9,153.43) -- (139.5,153.43) .. controls (142.26,153.43) and (144.5,155.67) .. (144.5,158.43) -- (144.5,200.03) .. controls (144.5,202.79) and (142.26,205.03) .. (139.5,205.03) -- (9,205.03) .. controls (6.24,205.03) and (4,202.79) .. (4,200.03) -- cycle ;

% Text Node
\draw (44.67,19.82) node [anchor=north west][inner sep=0.75pt]  [color={rgb, 255:red, 0; green, 0; blue, 0 }  ,opacity=1 ] [align=left] {$\displaystyle z_{s}^{0}$};
% Text Node
\draw (128.6,121.08) node   [align=left] {\begin{minipage}[lt]{50.64pt}\setlength\topsep{0pt}
{\scriptsize $\displaystyle \tau \in 1,...,t$}
\end{minipage}};
% Text Node
\draw (100.13,33.06) node [anchor=north west][inner sep=0.75pt]  [color={rgb, 255:red, 0; green, 0; blue, 0 }  ,opacity=1 ] [align=left] {$\displaystyle z_{m,i}$};
% Text Node
\draw (9,89.16) node [anchor=north west][inner sep=0.75pt]  [color={rgb, 255:red, 0; green, 0; blue, 0 }  ,opacity=1 ] [align=left] {$\displaystyle \left\{z_{s}^{( \tau )} ,z_{a}^{( \tau -1)} ;x_{p}^{( \tau )} ;\right\}$};
% Text Node
\draw (63.37,131.03) node [anchor=north west][inner sep=0.75pt]  [color={rgb, 255:red, 0; green, 0; blue, 0 }  ,opacity=1 ] [align=left] {$\displaystyle z_{s}^{t}$};
% Text Node
\draw (120.07,8.31) node   [align=left] {\begin{minipage}[lt]{41.62pt}\setlength\topsep{0pt}
{\scriptsize $\displaystyle {\textstyle i\in 1,...,I}$}
\end{minipage}};
% Text Node
\draw (6,156.43) node [anchor=north west][inner sep=0.75pt]  [color={rgb, 255:red, 0; green, 0; blue, 0 }  ,opacity=1 ] [align=left] {$\displaystyle \left\{z_{s}^{( \tau )} ,z_{a}^{( \tau -1)} ;x_{O}^{( \tau )} ;\right\}$};
% Text Node
\draw (103.97,191.77) node   [align=left] {\begin{minipage}[lt]{50.64pt}\setlength\topsep{0pt}
{\scriptsize $\displaystyle \tau \in t+1,...,T$}
\end{minipage}};
% Text Node
\draw (150.4,101.8) node [anchor=north west][inner sep=0.75pt]   [align=left] {SLAM};
% Text Node
\draw (154.2,171.4) node [anchor=north west][inner sep=0.75pt]   [align=left] {MDP};

\end{tikzpicture}
}
\caption{A combination of the generative flow graphs for the simultaneous localization and mapping problem and the Markov Decision Process shown in \cref{fig:slam_graph_idiom_new} and \cref{fig:MDP_graph_idiom_new}, respectively, could potentially constitute an end-to-end navigation behavior for a mobile robot.}
\label{fig:SLAM_plus_MDP}
\end{figure}

When we have developed such probabilistic programming idioms it empowers us to mix and match them to construct higher-level intelligence without worrying about all details of the underlying models. % as exemplified in \cref{fig:SLAM_plus_MDP}. 
E.g. \cref{fig:SLAM_plus_MDP} implies that the output of a specific model for the simultaneous localization and mapping problem is used as the input to a Markov decision process, but leaves out details about their internal structures.

\section{Inference Algorithms}\label{sec:inference}
% Performing efficient inference on general probabilistic programs is hard!\\
% Give an overview on current Inference algorithms and their limitations\\
% comment on how the generative nature of probabilistic programs might hinder parallelization of computations, and how we might be able to alleviate this by msg-passing.\\
% comment on how msg-passing algorithms might also make it possible to use different inference algorithms for different parts of cognition. Accurate inference might be prefered in some situations, wheres we might need fast but less accurate inference in other situations\\

As stated in \cref{sec:probabilistic_programs} a probabilistic program is a simultaneous representation of a joint distribution, $p_{\Theta}(Z,X)$, and a conditional distribution, $p_{\Theta}(X|Z)$. Having defined a model as such distributions we are usually interested in answering queries about the unconditioned/latent random variables, $Z$, given information about the conditioned/observed random variables, $X=\overline{X}$. In the combined navigation problem illustrated in \cref{fig:SLAM_plus_MDP} we are interested in determining which action to take, $z_a^{\{\tau \}}$, given prior perceived information, $x_p^{\{\tau \}}$ for $\tau \in 1,...,t$, and future optimality variables, $x_O^{\{\tau \}}$ for $\tau \in t+1,...,T$.
Often queries of interest are statistics such as the posterior mean and variance of specific random variables, or the posterior probability of a random variable being within a given set. Still, it could also simply be to sample from the posterior, $p_{\Theta}\left(Z|X=\overline{X}\right)$. All of these queries are somehow related to the posterior distribution given by
\begin{align}
    p_{\Theta}\left( { Z }|{ X=\overline { X }  } \right) &=\frac { p_{\Theta} \left( X=\overline { X } ,Z \right)  }{ { p_{\Theta} }\left( X=\overline { X }  \right)  } \nonumber \\
    &=\frac { { p_{\Theta} }\left( X=\overline { X } ,Z \right)  }{ \int { { p_{\Theta} }\left( X=\overline { X } ,Z \right) dZ }  }. \label{eq:intractable}
\end{align}

The marginalization by the integral in the denominator of \cref{eq:intractable} in general does not have an analytical solution or is intractable to compute in most realistic problems and approximate inference is therefore necessary \cite{8588399}. Through time, an abundance of algorithms has been developed to find an approximation to the posterior in specific problems. Unfortunately, many of these algorithms cannot be applied to general probabilistic programs mainly due to the possible unbounded number of random variables \cite{vandemeent2018introduction}. Possible applicable inference algorithms can roughly be divided into Monte Carlo based algorithms such as Sequential Monte Carlo, Metropolis-Hastings, and Hamiltonian Monte Carlo, and the optimization based Variational Inference algorithms such as Stochastic Variational inference. As the size and complexity of models of cognition increase, the computational efficiency of inference algorithms becomes a paramount necessity to achieve sufficient efficiency of the framework presented in \cref{sec:our_model}. While Monte Carlo methods often converge to the true posterior in the limit, convergence can be slow. Conversely, Variational Inference algorithms are often faster even though they can suffer from simplified posterior approximations \cite{8588399}. Also, as Variational Inference methods are based on optimization they provide a natural synergy with data-driven discriminative techniques such as deep learning. By accepting that robots are not necessarily supposed to behave optimally, but rather should behave as agents with bounded rationality, the above characteristic makes Variational Inference algorithms an especially interesting choice for cognitive robotics. Therefore, \cref{sec:VI} is devoted to giving the reader an introduction to the overall concept of Variational Inference. \Cref{sec:msg_passing} and \ref{sec:SVI} present two specific solution approaches commonly used in variational inference, namely Message-passing algorithms and stochastic variational inference, respectively. Both approaches have their weaknesses. Therefore, in \cref{sec:SMP} we outline a way of combining these two approaches to overcome their weaknesses. The idea of combining Message-passing with stochastic variational inference, we have presented before\cite{damgaard2022SVIFDPR}, however, here we generalizes the idea to generative flow graphs.

% Thus besides efficiently answering queries about the unconditioned/latent random variables, an objective of equally importance might be to learn appropriate values for some of the parameters $\theta$ in a model.

\subsection{Variational inference}\label{sec:VI}
%\textcolor{red}{Write about the problems with VI mentioned in \cite{vandemeent2018introduction} - i.e. no guarantee that all sample sites will be sampled.}
% Variational inference uses optimization to approximate one distribution, ${ p }\left(  Z   \right)$, by another simpler distribution, ${ q }\left( Z \right)$, called the variational distribution. Notice that, in general, ${ p }\left(  Z   \right)$ does not need to be a conditional distribution, ${ p }\left( { Z }|{ X=\overline { X }  } \right)$, as in \cref{eq:intractable}. However, for the sake of the topic in this paper we will focus on the conditional distribution case.
Variational inference is an optimization based approach to approximate one distribution, ${ p }\left(  Z   \right)$, by another simpler distribution, ${ q }\left( Z \right)$. ${ q }\left( Z \right)$ is usually called the variational distribution. In general, variational inference is not only used to approximate conditional distribution, ${ p }\left( { Z }|{ X=\overline { X }  } \right)$, as in \cref{eq:intractable}. However, with the presented framework in mind we will limit our presentation to this case, and focus on variational inference problems on the form
\begin{equation}
    q^{ * }\left( Z \right) =\arg\min _{ q\left( Z \right) \in Q }{ { D }\left( { p_{\Theta} \left( { Z }|{ X=\overline { X }  } \right)  }||{ q\left( Z \right)  } \right)  } \label{eq:opt_problem}
\end{equation}
%
%where $D$ is a so-called divergence measure, measuring the similarity between $p$ and $q$, and $Q$ is the family of variational distributions that we want to find our approximation from.
where $D$ is a measure of the similarity between $p$ and $q$ often called a divergence measure, and $Q$ is the family of variational distributions from which the approximation should be found. The notation $D(p||q)$ denotes a divergence measure and that the order of the arguments, $p$ and $q$, matters.
%Notice here that even though it is commonly used, the origin and meaning of the "$||$" notation is a bit unclear. However, in this paper we simply use it 
% https://stats.stackexchange.com/questions/269183/what-is-the-meaning-of-double-vertical-bar-in-this-kl-divergence-equation
% https://math.stackexchange.com/questions/1597380/origin-of-the-notation-for-statistical-divergence
%
% The family of variational distributions, $Q$, is usually selected as a compromise between how good an approximation one wants and computational efficiency, and should be chosen such that the given queries can be easily answered. It is important to stress that any variational inference method is more or less biased via the choice of the family of variational distributions, $Q$. Furthermore, the divergence measure, $D$, can have a rather large impact on the approximation.
The choice the family of variational distributions, $Q$, is a compromise between computational efficiency and precise an approximation one wants. Furthermore, $Q$ should be chosen such that we can easily answer given queries. It is important to stress that any variational inference method is more or less biased via the choice of the family of variational distributions, $Q$. As a consequence we cannot view the original model in isolation, and have to consider the variational distribution, $q\left( Z \right) $, as an implicit part of the cognitive model. Besides the family of variational distributions, the choice of the divergence measure, $D$, can substantially impact the properties of the approximation. \textcolor{black}{ However, empirical results suggest that for the family of $\alpha$-divergences, subsuming the commonly used Kullback–Leibler divergence, all choices will give similar results as long as the approximating family, $Q$, is a good fit to the true posterior distribution \cite{minka2005divergence}. }

\subsection{Message-Passing}\label{sec:msg_passing}
Message-passing algorithms solves a possible complicated variational inference problem as defined by \cref{eq:opt_problem} by it down into a series of more tractable sub-problems \cite{minka2005divergence}. The methods are known as message-passing algorithms due to the way that the solution to one sub-problem is distributed to the other sub-problems. Message-passing algorithms assumes that the model of a problem, $p(Z|X)$, can be factorized into a product of probability distributions
\begin{align}
p(Z|X) & =\prod_{ a\in A }^{  } p^{\{ a\}}( Z|X).
\end{align}
% Notice that the factorization need not be unique and that each probability distribution, $p^{\{ a\}}( Z|X)$, can depend on any number of the variables of $p(Z|X)$. Similarly, we can choose a variational distribution, $q\left( Z \right)$, that factorizes into a similar form 
This factorization need not be unique and each factor, $p^{\{ a\}}( Z|X)$, can depend on any number of the variables of $p(Z|X)$. The variational distribution, $q\left( Z \right)$, should furthermore be chosen such that it factorizes into a similar form 
\begin{align}
    q\left( Z \right) =\prod_{ a\in A }^{  }{ q^{ \left\{ a \right\}  }{ \left( Z \right)  } }. \label{eq:msg_passing_q_factor}
\end{align}
%
%Now by defining the product of all other than the $a$'th factor of $q\left( Z \right)$ and $p(Z|X)$, respectively as
With these assumptions, define the product of all other than the $a$'th factor of $q\left( Z \right)$ and $p(Z|X)$, respectively as
\begin{align}
   q^{ \backslash a }\left( Z \right) &=\prod _{ b\in A\backslash a }^{  }{ q^{ \left\{ b \right\} }{ \left( Z \right)  } } ,\\
   p^{ \backslash a }\left( Z|X \right) &=\prod _{ b\in A\backslash a }^{  }{ p^{ \left\{ b \right\}  }{ \left( Z|X \right)  } }. 
\end{align}
%
%and by further assuming that $q^{ \backslash a }\left( Z \right) \approx p^{ \backslash a }\left( Z|X \right) $ is in fact a good approximation, it is possible to rewrite our full problem in \cref{eq:opt_problem} into a series of approximate sub-problems on the form
With these definitions it is possible to rewrite the problem in \cref{eq:opt_problem} into a series of approximate sub-problems on the form
\begin{figure}
    \centering
        %\fcolorbox{black}{gray!30}{%https://tex.stackexchange.com/questions/20640/how-to-add-border-for-an-image
            \algsetup{indent=2em}
            \begin{algorithmic}[1]
               \STATE Initialize $q^{ \left\{ a \right\} * }{ \left( Z \right)  }$ for all $a\in A$
               \REPEAT 
                    \STATE Pick a factor $a\in A$
                    \STATE Solve \cref{eq:msg_optimization} to find $q^{ \left\{ a \right\} * }{ \left( Z \right)  }$ 
               \UNTIL{$q^{ \left\{ a \right\} * }{ \left( Z \right)  }$ converges for all $a\in A$}
            \end{algorithmic}
        %}
    \caption{The generic message-passing algorithm. The loop in line 2 can potentially be run in parallel and in a distributed fashion.}
    \label{Alg:generic_msg_passing}
\end{figure}
\begin{align}
    & q^{ \left\{ a \right\} * }{ \left( Z \right)  }\approx  \label{eq:msg_optimization}\\
    & \quad \arg\min _{ q^{ \left\{ a \right\}  \in Q^{ \left\{ a \right\} } } }{ D\left[ p^{ \left\{ a \right\}  }{ \left( Z|X  \right)  }{ q^{ \setminus a }\left( Z \right)  }  || q^{ \left\{ a \right\}  }{ \left( Z \right)  }{ q^{ \setminus a }\left( Z \right)  } \right] }  \nonumber
\end{align}
where $q^{ \backslash a }\left( Z \right)$ is assumed to be a good approximation and thus is kept fixed.
%Assuming a sensible choice of factor families, $Q^{ \left\{ a \right\} }$, from which $q^{ \left\{ a \right\}  }$ can be chosen, the problem in \cref{eq:msg_optimization} can be more tractable than the original problem, and by iterating over these coupled sub-problems as shown in \cref{Alg:generic_msg_passing}, we can obtain an approximate solution to our original problem. 
If the factor families, $Q^{ \left\{ a \right\} }$, from which $q^{ \left\{ a \right\}  }$ can be chosen, have been chosen sensible, the problem in \cref{eq:msg_optimization} can be more tractable than the original problem, and an approximate solution to the original problem can then be obtain by iterating over these coupled sub-problems as shown in \cref{Alg:generic_msg_passing}. In principle, we can even use different divergence measures for each sub-problem to do mismatched message-passing, which could make some of the sub-problems easier to solve as described in \cite{minka2005divergence}
%This way of solving a variational inference problem is known as message-passing because the solution of each sub-problem, $q^{ \left( a \right) * }{ \left( z \right)  }$, can be interpreted as a message send to the other sub-problems. 

%The approach is not guaranteed to converge for general problems. Furthermore, \cref{eq:msg_optimization} might still be a hard problem to solve, and thus previously in practice, the approach has been limited to problems for which \cref{eq:msg_optimization} can be solved analytically such as fully discrete or Gaussian problems \cite{minka2005divergence}. However, besides breaking the original problem into a series of more tractable sub-problems, this solution approach gives a principle way of solving the original problem in a distributed and parallel fashion, which could be a huge benefit for large models. Furthermore, depending on the dependency structure of the problem, not all sub-problems depend on the solution of the other sub-problems, which can significantly reduce the amount of communication needed due to sparsely connected networks.
In general, the approach is not guaranteed to converge, and \cref{eq:msg_optimization} might still be a hard problem to solve. In the past, this has limited the approach to relatively simple problems such as fully discrete or Gaussian problems for which \cref{eq:msg_optimization} can be solved analytically \cite{minka2005divergence}. Therefore, the true power of the method is the principle way in which it allows for solving problems in a distributed and parallel fashion, which can be a huge benefit for large models. Furthermore, if the sub-problems are sparsely connected, meaning that sub-problems does not depend on the solution to all of the other sub-problems, the amount of communication needed can be significantly reduced.

\subsection{Stochastic Variational inference}\label{sec:SVI}
The approach taken by Stochastic Variational inference (SVI) is to reformulate a variational inference problem, e.g., \cref{eq:opt_problem} or \cref{eq:msg_optimization}, to a dual maximization problem with an objective, $L$, that that can be solved with stochastic optimization \cite{JMLR:v14:hoffman13a}. Stochastic Variational inference assumes that the variational distribution, $q$, is parameterized by some parameters, $\Phi$. To obtain the dual problem and the objective function, $L$, of the resulting maximization problem of course the steps and assumptions needed depends on whether we have chosen the use the Kullback–Leibler divergence {\cite{JMLR:v14:hoffman13a}${}^{,}$\cite{pmlr-v33-ranganath14}${}^{,}$\cite{JMLR:v18:16-107}} or the $\alpha$-divergences \cite{NIPS2016_7750ca35}. Nevertheless, the resulting problem ends up being on the form
\begin{equation}
     \Phi^{*} = \arg \max _{ \Phi }{ { \quad { \underbrace { L\left( { p_{ \Theta }\left( Z,X=\overline { X }  \right)  },{ { q }_{ \Phi  }\left( Z \right)  } \right)  }_{ { E }_{ Z\sim q_{ \Phi  }\left( Z \right)  }\left[ l\left( Z, \Theta, \Phi   \right)  \right]  }  } } }. \label{eq:opt_problem_param}
\end{equation}
\textcolor{black}{This dual objective function, $L$, does not depend on the posterior, $p_{ \Theta }\left( { Z }|{ X=\overline { X }  } \right)$, but only the variational distribution, $q_{ \Phi  }\left( Z \right)$ and the unconditional distribution $p_{ \Theta }\left( Z,X=\overline { X }  \right)$ making the problem much easier to work with.} Besides being dual to \cref{eq:opt_problem} it turns out that for the family of alpha-divergences with $\alpha>0$, $L$ is also an lower bound on the log evidence, $log \left(p_\Theta\left(Z\right)\right)$ \cite{NIPS2016_7750ca35}. Since the log evidence is a measure of how well a model fits the data, we can instead consider the optimization problem \cite{DBLP:journals/corr/KingmaW13}
\begin{equation}
     \Theta^{*},\Phi^{*} = \arg \max _{ \Theta, \Phi }{ { \quad { \underbrace { L\left( { p_{ \Theta }\left( Z,X=\overline { X }  \right)  },{ { q }_{ \Phi  }\left( Z \right)  } \right)  }_{ { E }_{ Z\sim q_{ \Phi  }\left( Z \right)  }\left[ l\left( Z, \Theta, \Phi  \right)  \right]  }  } } }. \label{eq:opt_problem_param_2}
\end{equation}
that allows us to simultaneously fit the posterior approximation, $q_\Phi$, and model parameters, $\Theta$, to the data, $\overline{X}$. % Furthermore, by utilizing the reparameterization trick {\cite{DBLP:journals/corr/KingmaW13}${}^{,}$\cite{Salimans_2013}${}^{,}$\cite{pmlr-v32-rezende14}} or the REINFORCE-gradient \cite{Reinforce} it is possible to obtain an unbiased estimate of the gradient, $\overline{{ \nabla  }_{ W }L}$, of the dual objective $L$ where $W=\left\{\Theta, \Phi\right\}$. Stochastic gradient ascent can then be used to iteratively optimize the objective through the update equation
An unbiased estimate of the gradient, $\overline{{ \nabla  }_{ W }L}$, of this dual objective $L$ where $W=\left\{\Theta, \Phi\right\}$, can be obtained by utilizing the REINFORCE-gradient \cite{Reinforce} or the reparameterization trick {\cite{DBLP:journals/corr/KingmaW13}${}^{,}$\cite{Salimans_2013}${}^{,}$\cite{pmlr-v32-rezende14}}. The objective can then iteratively be optimized by stochastic gradient ascent via the the update equation
\begin{align}
    { W }^{ \{l\}  }={ W }^{ \{l-1\}  }+\rho ^{ \{l-1\}  }\overline{{ \nabla  }_{ W }L}^{ \{l\}  }\left( W^{ \{l-1\}  } \right)
\end{align}
where superscript $\{l\}$ is used to denote the $l$'th iteration. %\textcolor{red}{If the sequence of learning rates, $\rho ^{ \{l-1\} }$, follows the Robbins-Monro conditions,}
Stochastic gradient ascent converges to a maximum of the objective function $L$ if the sequence of learning rates, $\rho ^{ \{l-1\} }$, follows the Robbins-Monro conditions given by
\begin{align}
    \sum _{ l=1 }^{ \infty  }{ \rho ^{ \{l\}  } } =\infty, \qquad \qquad
    \sum _{ l=1 }^{ \infty  }{ { \left( \rho ^{ \{l\}  } \right)  }^{ 2 } } <\infty.
\end{align}
%
%then stochastic gradient ascent converges to a maximum of the objective function $L$, and since \cref{eq:opt_problem_param_2} is dual to the original minimization problem it provides a solution to the original problem. An unbiased gradient estimator with low variance is pivotal for this method and variance reduction methods are often necessary. However, a discussion of this subject is outside the scope of this paper and can often be done automatically by probabilistic programming libraries/languages such as Pyro \cite{bingham2019pyro}.
Since \cref{eq:opt_problem_param_2} is dual to the original minimization problem, this maximum also provides a solution to the original problem. Although Robbins-Monro conditions are satisfied it is often necessary to apply variance reduction methods to obtain unbiased gradient estimators with sufficiently low variance. Fortunately, reduction methods can often be applied automatically by probabilistic programming libraries/languages such as Pyro \cite{bingham2019pyro}. % Besides providing the basic algorithms for stochastic variational inference, such modern probabilistic programming languages also provide ways of defining a wide variety of probability distributions and extensions to stochastic variational inference that permits incorporating and learning of parameterized functions, such as neural networks, into the unconditional distribution $p\left( z,x=\overline { x }  \right)$. Thereby, making the approach very versatile.
One benefit \textcolor{black}{of solving variational inference problems with stochastic optimization is that noisy gradient estimates are often relatively cheap to compute due to, e.g., subsampling of data.} Another benefit is that \textcolor{black}{the use of noisy gradient estimates can cause algorithms to escape shallow local optima of complex objective functions} \cite{JMLR:v14:hoffman13a}. The downside of Stochastic variational inference is that it is inherently serial and that it requires the parameters to fit in the memory of a single processor \cite{pmlr-v89-zhang19c}. This could potentially be a problem for cognitive robotics where large models with lots of variables and parameters presumably are necessary to obtain a high level of intelligence, and where queries have to be answered within different time scales. I.e. signals to motors have to be updated frequently while high-level decisions can be allowed to take a longer time.

\subsection{Stochastic Message-Passing}\label{sec:SMP}
To summarize the previous sections Message-passing algorithms exploits the dependency structure of a given variational inference problem to decompose the overall problem into a series of simpler variational inference sub-problems, that can be solved in a distributed fashion \cite{minka2005divergence}. Message-passing algorithms do not give specific directions on how to solve these sub-problems, and thus classically required tedious analytical derivations, that effectively limited the usability of the method. On the other hand, modern Stochastic variational inference methods directly solve such variational inference problems utilizing stochastic optimization while simultaneously learning model parameters. By fusion of these two approaches, we could potentially overcome the serial nature of Stochastic variational inference to solve large-scale complex problems in a parallel and distributed fashion. However, to do so we need to find an appropriate factorization of a given problem. Again we can make use of the semantics of generative flow graphs. Assuming that we can divide all the nodes of a given generative flow graph into a set, $C^{\{A\}}$, of node collections $C^{\{a\}} =\left\{Z^{\{a\}} ;X^{\{a\}} ;\Theta ^{\{a\}}\right\}$ and a set of "global" observed variable nodes, $X_{G}$, having more than one node collection as parent, we can write the posterior factorization
\begin{align}
    & p_{\Theta } (Z|X) \nonumber\\
    & \quad=p_{\Theta } (Z^{\{A\}} |X^{\{A\}} ,X_{G} )\nonumber\\
    & \quad=\frac{1}{p(X_{G} |X^{\{A\}} )} p_{\Theta } (Z^{\{A\}} ,X_{G} |X^{\{A\}} )\nonumber\\
    & \quad=\frac{p(X_{G} |Z^{\{A\}} ,X^{\{A\}} )}{p_{\Theta } (X_{G} |X^{\{A\}} )} p_{\Theta } (Z^{\{A\}} |X^{\{A\}} )\nonumber\\
    & \quad=\frac{p(X_{G} |Z^{\{A\}} )}{p_{\Theta } (X_{G} |X^{\{A\}} )} p_{\Theta } (Z^{\{A\}} |X^{\{A\}} )\label{eq:msg_factorization_2}\\
    & \quad=\frac{p(X_{G} |Z^{\{A\}} )}{p_{\Theta } (X_{G} |X^{\{A\}} )}\label{eq:msg_factorization_1}\\
    & \quad\quad\quad \cdot \prod _{a\in A} p_{\Theta ^{\{a\}} ,\text{Pa}\breve{\Theta }\left( C^{\{a\}}\right)}\left( Z^{\{a\}} |\text{Pa}\breve{Z}\left( C^{\{a\}}\right) ,X^{\{a\}}\right)\nonumber
\end{align}
where \cref{eq:msg_factorization_2} follows from conditional independence between $X_{G}$ and $X^{\{A\}}$ given $Z^{\{A\}}$, and \cref{eq:msg_factorization_1} follows from \cref{eq:collections_factorization}. %, and we do not explicitly write the parameters for brevity. 
Following the procedure of message-passing we choose a variational distribution that factorizes as
\begin{align}
q_{\Phi}( Z) & =\prod _{a\in A} q_{\Phi ^{\{a\}}}\left( Z^{\{a\}}\right)\label{eq:S-msg_passing_q_factor}
\end{align}
Notice, that in \cref{eq:S-msg_passing_q_factor} we have exactly one factor for each node collection and that this factor only contains the latent variables of that node collection. This is unlike \cref{eq:msg_passing_q_factor} where a latent variable could be present in multiple factors. By combining \cref{eq:msg_factorization_1} and \cref{eq:S-msg_passing_q_factor} we can write an approximate posterior distribution related to the a'th node collection $ p( Z|X) \approx \tilde{p}^{\{a\}}( Z|X)$ where 
\begin{align*}
 & \tilde{p}_{\Theta ^{\{a\}}}^{\{a\}} (Z|X)=\\
 & \quad p_{\Theta ^{\{a\}} ,\text{Pa}\breve{\Theta }\left( C^{\{a\}}\right)}\left( Z^{\{a\}} |\text{Pa}\breve{Z}\left( C^{\{a\}}\right) ,X^{\{a\}}\right)\\
 & \quad \quad \cdot \frac{p(X_{G} |Z^{\{A\}} )}{\tilde{p}^{\{a\}} (X_{G} |X^{\{a\}} )}\prod _{b\in A\setminus a} q_{\breve{\Phi }^{\{b\}}}\left( Z^{\{b\}}\right)
\end{align*}
Where $\tilde{p}^{\{a\}} (X_{G} |X^{\{a\}} )$ is defined in \cref{eq:marginal_likelyhood_approx} in \cref{app:S-msg}. Based on \cref{eq:msg_optimization} we can then define approximate sub-problems as
\begin{equation}
\underset{\Phi ^{\{a\}}}{\min} \ D\left[ q_{\Phi ^{\{a\}}}( Z) ||\tilde{p}_{\Theta ^{\{a\}}}^{\{a\}}( Z|X)\right] \label{eq:S-msg_passing_sub_problems}
\end{equation}
Each of these sub-problems can then be solved successively or in parallel potentially on distributed compute instances as outlined in \cref{Alg:generic_msg_passing} and utilizing Stochastic variational inference as described in \cref{sec:SVI}. To see how this choice of factorization affects the posterior approximations and the learning of model parameters, $\Theta$, consider the KL-divergence as divergence measure. Considering the KL-divergence we can rewrite the objective in \cref{eq:S-msg_passing_sub_problems} as shown in  \cref{eq:msg_svi_kl_rewritten1} through  \cref{eq:msg_svi_kl_rewritten2} in \cref{app:S-msg} to obtain the following local dual objective for stochastic variational inference
\begin{align}
 & L_{KL}^{\{a\}}\left( \Theta ^{\{a\}} ,\Phi ^{\{a\}}\right) = \label{eq:s_kl_rewritten_2}\\
 & \quad \quad E_{Z\sim \tilde{q}_{\text{Pa}\breve{Z}}^{\{a\}}}\left[ LogEvd_{X_{G} ,\ X^{\{a\}}}^{\{a\}}\left( \Theta ^{\{a\}}\right)\right] -C \nonumber\\
 & \quad \quad \quad \quad \quad \quad \quad -D_{KL}\left[ q_{\Phi ^{\{a\}}} (Z)||\tilde{p}_{\Theta ^{\{a\}}}^{\{a\}} (Z|X)\right] \nonumber
\end{align}
Where $C$ is a constant with respect to $\Theta ^{\{a\}}$ and $\Phi ^{\{a\}}$, and $LogEvd^{\{a\}}\left( X_{G} ,\ X^{\{a\}}\right)$ is the joint log-evidence over global observed variables, $X_{G}$, and observed variables, $\ X^{\{a\}}$, local to the a'th node collection. Since the first term on the right-hand side is constant with respect to $\Phi ^{\{a\}}$, maximizing this local dual objective with respect to $\Phi ^{\{a\}}$ will minimize the KL-divergence. Furthermore, since $D_{KL}\left[ q_{\Phi ^{\{a\}}} (Z)||\tilde{p}_{\Theta ^{\{a\}}}^{\{a\}} (Z|X)\right] \geq 0$ by definition it follows from \cref{eq:s_kl_rewritten_2} that
\begin{align*}
 & E_{Z\sim \tilde{q}_{\text{Pa}\breve{Z}}^{\{a\}}}\left[ LogEvd^{\{a\}}\left( X_{G} ,\ X^{\{a\}}\right)\right] -C\\
 & \quad \quad \quad \quad \quad \quad \quad \quad \quad \geq L_{KL}^{\{a\}}\left( \Theta ^{\{a\}} ,\Phi ^{\{a\}}\right)
\end{align*}
%
%That is, $L_{KL}^{\{a\}}$ is a lower bound on the expected log-evidence of the global, $X_{G}$, and the local, $x^{\{a\}}$, observed variables, where the expectation is taken with respect to the variational distribution over latent variables parent to the a'th node collection. 

Therefore, by maximizing the local dual objective, $L_{KL}^{\{a\}}\left( \Theta ^{\{a\}} ,\Phi ^{\{a\}}\right)$, with respect to the local model parameters, $\Theta ^{\{a\}}$, we will push the expected joint log-evidence over the global, $X_{G}$, and the local, $X^{\{a\}}$, observed variables higher, where the expectation is taken with respect to the joint variational distribution over latent variables parent to the a'th node collection. In summary, this means that we can simultaneously fit our local model parameters, $\Theta ^{\{a\}}$, to the evidence and obtain an approximate local posterior distribution, $q_{\Phi ^{\{a\}}}( z^{\{a\}})$. While the above derivations were made for the KL-divergence, similar derivations can be done for the more general family of $\alpha$-divergences. 

To evaluate this local dual objective, we only need information related to the local node collection, its parents, and other node collections having the same global observed variables as children. Thereby providing substantially computational speedups for generative flow graphs with sparsely connected node collections and global observed variables. To use this procedure with a standard probabilistic programming language, we would have to create a probabilistic program fragment for each node collection, their corresponding variational distribution, and the global observed variables. These fragments would then have to be composed together to form the local objectives potentially in an automated fashion.

So far within this section, we have assumed that all sub-problems are solved through a variational problem as in \cref{eq:S-msg_passing_sub_problems}. However, there are in principle no reasons why we could not use estimates of sub-posteriors, $q( z^{\{b\}})$, obtained through other means in \cref{eq:S-msg_passing_sub_problems}, as long as we can sample from these sub-posteriors. Thus, making the outlined method very flexible to combine with other methods, albeit analysis of the results obtained through the combined inference becomes more difficult. It is also important to stress that the factorization used above is not unique. It would be interesting to investigate if other factorizations could be employed, and for which problems these factorizations could be useful.

In summary, if we can divide a generative flow graph representing an overall model of cognition into node collections and global observed variables, then we can utilize the combination of Message-Passing and Stochastic Variational inference presented within this section, to distribute the computational burden of performing inference within this model. At the same time, we can learn local model parameters. Thus, yielding a very flexible tool allowing us to fully specify the part of a model that we are certain about, and potentially learn the rest.

\section{Probabilistic Programming Languages}\label{sec:ppl}
So far, our focus has been on the representation of models defined by probabilistic programs, and on how to answer queries related to these models via modern probabilistic inference. However, we have not considered how this is made possible by probabilistic programming languages and their relation to deterministic programming languages. Here we will not give a detailed introduction to probabilistic programming and refer interested readers to other sources{\cite{vandemeent2018introduction}${}^{,}$\cite{10.5555/2851115}${}^{,}$\cite{10.5555/3033232}}. Instead, we will give a short overview of languages relevant to modeling cognition.

As already mentioned in \cref{sec:probabilistic_programs} the main characteristics of a probabilistic program is a construct for sampling randomly from distributions and another construct for condition values of variables in the program. The purpose of probabilistic programming languages is to provide these two constructs and to handle the underlying machinery for implementing inference algorithms and performing inference from these constructs. As with any other programming language, design decisions are not universally applicable or desirable, and different trade-offs are purposefully made to achieve different goals. This fact combined with theoretical advancements has resulted in several different probabilistic programming languages. For an extensive list see \cite{vandemeent2018introduction}. Some of these are domain-specific aimed at performing inference in a restricted class of probabilistic programs, such as STAN \cite{JSSv076i01}. These restrictions are usually employed to obtain more efficient inference. More interesting for the framework presented in \cref{sec:our_model}, however, are languages self-identifying as universal or general-purpose, such as Pyro \cite{bingham2019pyro} and Venture {\cite{mansinghka2014venture}${}^{,}$\cite{10.1145/3192366.3192409}}. These languages aim at performing inference in arbitrary probabilistic programs. Thus maximizing the flexibility for modeling cognition. 

A recent trend has been to build probabilistic programming languages on top of deep-learning libraries such as PyTorch \cite{NEURIPS2019_9015} and TensorFlow \cite{tensorflow2015-whitepaper}. This is done both to use the efficient tensor math, automatic differentiation, and hardware acceleration that these libraries provide and to get tighter integration of deep-learning models within probabilistic models. Examples of such languages are Pyro \cite{bingham2019pyro} and ProbTorch \cite{siddharth2017learning} build upon PyTorch, and Edward \cite{tran2017deep} build upon TensorFlow. Again, when considering the use within the framework presented in \cref{sec:our_model}, the languages based on PyTorch or TensorFlow 2.0 could potentially have the edge over others due to the dynamic approach to constructing computation graphs. This is because the dynamic computation graphs, more easily allow us to define dynamic models which include recursion and unbounded numbers of random choices \cite{vandemeent2018introduction}. Constructs potentially being indispensable for models of higher-level cognition supposed to evolve.

Python as the high-level general-purpose programming language it is makes modeling effortless in these languages. However, being based on python the computational efficiency of these languages is potentially limited by the need for interpretation. For this reason the relatively new project called NumPyro \cite{DBLP:journals/corr/abs-1912-11554} is  under active development. NumPyro provides a backend to Pyro based on NumPy \cite{harris2020array} and JAX \cite{47008} which enables just-in-time compilation, and thus potentially could provide much better computational efficiency, which again is essential for any practical robotic system.

To summarize, the choice of which probabilistic programming language to use depends on the end flexibility needed to model cognition. However, universal or general-purpose languages based on deep-learning libraries possibly with just-in-time compilation and hardware acceleration seem promising for general modeling of cognition, and especially for cognitive robotics.

%\section{Probabilistic Programming Languages}\label{sec:ppls}
% Almost all languages have pseudo-random value generators or packages; what they lack in comparison to probabilistic programming languages is syntactic constructs for conditioning and evaluators that implement conditioning.
%While most programming languages have pseudo-random value generators what really defines Probabilistic Programming Languages is the syntactic constructs for conditioning. \\
%explain the inference controller, and how this allows us to implement inference algorithms and probabilistic programs in different languages.

%\section{Deterministic Programming Languages}
%consider explaining why pytorch might be preferable due to dynamic computation graphs.

\section{Application Examples}\label{sec:our_examples}
%Indsæt resultaterne fra de 2 andre papers og der skal bare refereres til de dertilhørende papers og git repos. En kort beskrivelse af eksperimentet (måske et billede af setup). Figurer der viser resultagerne.

To demonstrate the concepts presented within this paper and the utility of the framework we have begun an initiative to implement some general applicable probabilistic programming idioms with basis in the "Standard Model of the Mind" \cite{Laird_Lebiere_Rosenbloom_2017} which is available as a GitHub repository\cite{ProbMind}. The repository currently contain one such idiom called "Planning". The purpose of this idiom is to provide basic functionality to plan future actions of a robot based on concepts of desirability, progress, information gain, and constraints. To keep the idiom generally applicable it is implemented as an abstract python class with methods containing the main functionality, and some abstract methods that needs to be specified on a per application basis. Besides this idiom, the repository also contains two different applications examples utilizing this idiom. 
\begin{figure}[!h]
    \centering
    \includegraphics[width=0.8\linewidth]{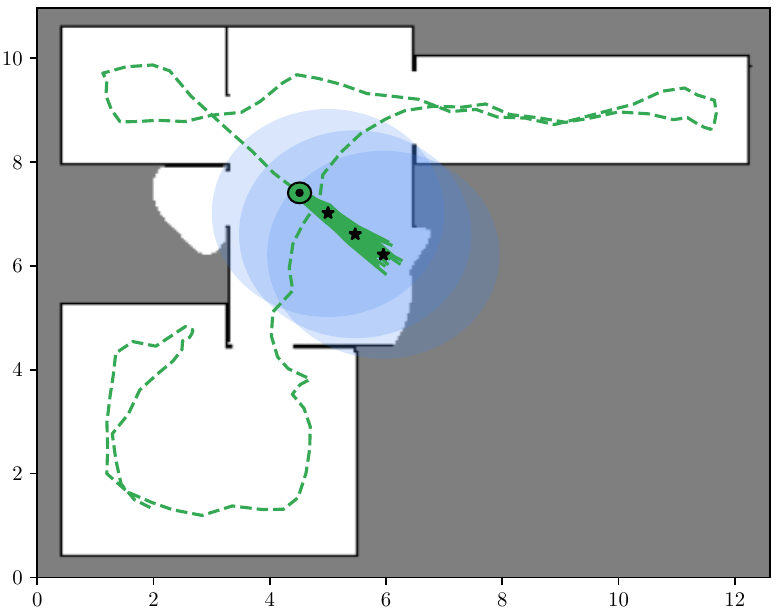}
    \caption{Results of a simulation of high-level robot motion planning with the goal of exploring an unknown environment with a lidar as the perceptual input. Gray indicate unexplored parts of the environment, white indicate unoccupied areas, black indicates obstacles, the green circle with a black boarder shows the current location of the robot, the green dashed line shows the robots past path, the solid green lines shows samples from the future optimal path distribution, the black stars shows the mean of these samples, and the transparent blue circles illustrates the lidars range at these positions.}\label{fig:exploration_example}
\end{figure}

The first example demonstrates high-level robot motion planning with the goal of exploring an environment represented by a grid map in the long-term memory as illustrated in \cref{fig:exploration_example}\cite{damgaard2022AKS}.

\begin{figure}[!h]
    \centering
    \includegraphics[width=0.8\linewidth]{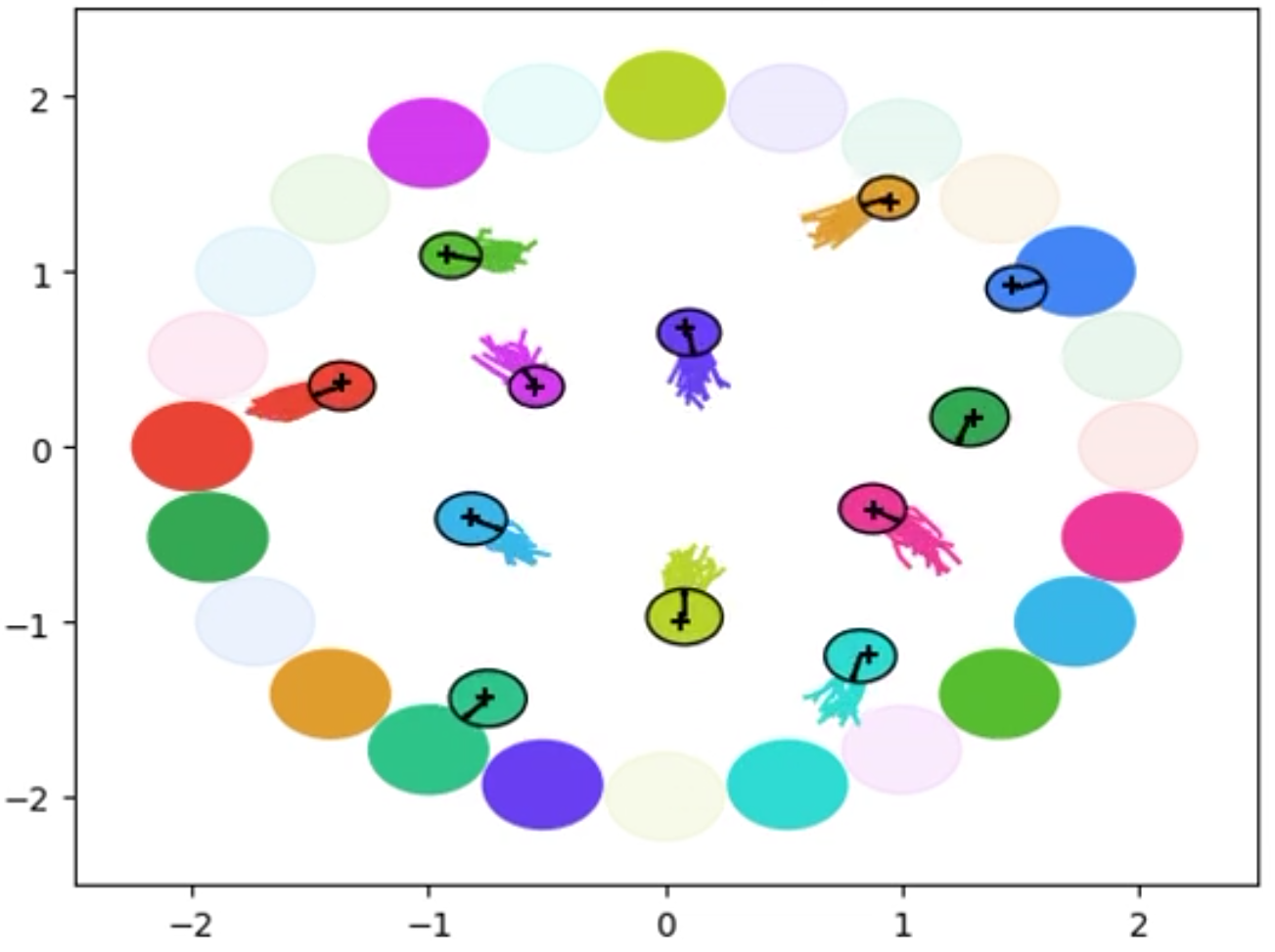}
    \caption{Snapshot of a simulation with 12 robots utilizing the "Planning" idiom to plan actions towards their goal while avoiding collision with each other. Colored circles with a black boarder indicates the current location of the robots, solid colored lines indicates samples of their future planned path distribution, colored circles indicates their current goals, and transparent colored circles indicates their last goal.}\label{fig:multiRobot}
\end{figure}

The second example relies heavily on the Stochastic Message-Passing approach described in \cref{sec:SMP} to implement a simplistic form of theory-of-mind\cite{damgaard2022SVIFDPR}. In this application N robots each uses the "Planning" idiom to plan low-level actions towards their goals while avoiding collisions with the other robots given knowledge about the other robots expected future path as illustrated in \cref{fig:multiRobot}. This approach have both been verified through simulations and a real-world experiment.

Although the repository currently do not contain a broad range of cognitive capabilities, the two examples demonstrates most of the concepts presented in \cref{sec:probabilistic_programs} through \cref{sec:ppl}. Furthermore, the fact that the two vastly different applications are implemented from the same probabilistic programming idiom demonstrates how models developed in the proposed framework can encourage cooperation and re-use of existing results.

% \section{Discussion}\label{sec:Discussion}
% 1) Vi skal huske at kommentere på hvad der ligner en øgelse i kompleksitet i forhold til den originale model.
% 2) "To what extent, the method can reduce the development cost of cognitive systems?"
% 3) "In what way can the framework encourage and mitigate cooperation and re-use of existing results?"
% 4) 

\section{Conclusion}\label{sec:conclusion}
Inspired by Sigma's Cognitive Hourglass Model \cite{RosenbloomDemskiUstun+2017+1+103}, we have outlined a new framework for developing cognitive architectures for cognitive robotics. With probabilistic programs at the center, this framework is sufficiently general to span the full spectrum of emergent, symbolic, and hybrid architectures. By dividing cognitive architectures into a series of layers this framework provides levels of abstractions between models of cognition and the algorithms that implement them on computational devices. %Furthermore, this new framework divides the task of developing a cognitive architecture into a series of layers building on top of each other. Through these layers this framework provides imperative levels of abstractions between models of cognition and the algorithms that implements them on computational devices. 
Some of these layers also directly relate to other fields of research, thereby encouraging better cooperation. 

%By the importance of probabilistic programs within this framework, 
We also presented a novel graphical representation of probabilistic programs which we call generative flow graphs. We showed how such generative flow graphs can help identify important universal fragments of probabilistic programs and models. Fragments that could potentially be re-used in the development of other cognitive architectures. Thereby, again encouraging cooperation and easier re-use of existing results.

Furthermore, we introduced the problem of inference within probabilistic programs. We briefly reviewed possible approaches and argued that variational inference approaches seems interesting for cognitive robotics. We introduced two commonly used approaches called Message-Passing and Stochastic Variational Inference. We also outline the weaknesses of each approach and proposed a combined approach that we call Stochastic Message-Passing. The proposed approach provides a principle way of distributing the computational burden of inference and parameter learning.

To support implementations within the framework we reviewed existing probabilistic programming languages providing the necessary machinery to implement inference algorithms for and perform inference in probabilistic programs.

Finally, we provided a brief introduction to a new initiative that both provide evidence for the applicability of the frameworks and concepts presented within this paper, but also functions as a starting point and a tool for researchers who wants to work within the framework.

The main topics within this paper have been the framework itself, the representation of cognitive models, and the computational burden.  These topics are indeed essential ingredients of the framework and pose interesting research directions by themselves. % However, it will be interesting to see how the proposed framework hopefully can aid the development of cognitive architectures in the future.

%
%\section{Discussion}\label{sec:discussion}

\section{Author Contributions}
Conceptualization, M.R.D., R.P., and T.B.; Investigation, M.R.D.; Methodology, M.R.D.; Formal Analysis, M.R.D.; Visualization, M.R.D.; Writing – Original Draft, M.R.D.; Writing – Review \& Editing, M.R.D., R.P., and T.B.; Supervision, R.P., and T.B.

\section{Declaration of interests}
The authors declare no competing interests.

\addtolength{\textheight}{-0cm}   % This command serves to balance the column lengths

%%%%%%%%%%%%%%%%%%%%%%%%%%%%%%%%%%%%%%%%%%%%%%%%%%%%%%%%%%%%%%%%%%%%%%%%%%%%%%%%
% \section*{ACKNOWLEDGMENT}
%%% --- The following two lines are what needs to be added --- %%%

\bibliographystyle{IEEEtran}
\bibliography{IEEEabrv,bibliography}

\begin{thebibliography}{10}
\providecommand{\url}[1]{#1}
\csname url@rmstyle\endcsname
\providecommand{\newblock}{\relax}
\providecommand{\bibinfo}[2]{#2}
\providecommand\BIBentrySTDinterwordspacing{\spaceskip=0pt\relax}
\providecommand\BIBentryALTinterwordstretchfactor{4}
\providecommand\BIBentryALTinterwordspacing{\spaceskip=\fontdimen2\font plus
\BIBentryALTinterwordstretchfactor\fontdimen3\font minus
  \fontdimen4\font\relax}
\providecommand\BIBforeignlanguage[2]{{%
\expandafter\ifx\csname l@#1\endcsname\relax
\typeout{** WARNING: IEEEtran.bst: No hyphenation pattern has been}%
\typeout{** loaded for the language `#1'. Using the pattern for}%
\typeout{** the default language instead.}%
\else
\language=\csname l@#1\endcsname
\fi
#2}}

\bibitem{RosenbloomDemskiUstun+2017+1+103}
\BIBentryALTinterwordspacing
P.~S. Rosenbloom, A.~Demski, and V.~Ustun, ``The sigma cognitive architecture
  and system: Towards functionally elegant grand unification,'' \emph{Journal
  of Artificial General Intelligence}, vol.~7, no.~1, pp. 1--103, 2017.
  [Online]. Available: \url{https://doi.org/10.1515/jagi-2016-0001}
\BIBentrySTDinterwordspacing

\bibitem{Haazebroek}
\BIBentryALTinterwordspacing
P.~Haazebroek, S.~van Dantzig, and B.~Hommel, ``A computational model of
  perception and action for cognitive robotics,'' \emph{Cognitive Processing},
  vol.~12, no.~4, p. 355, 2011. [Online]. Available:
  \url{https://doi.org/10.1007/s10339-011-0408-x}
\BIBentrySTDinterwordspacing

\bibitem{electronics10070793}
\BIBentryALTinterwordspacing
J.~Zhong, C.~Ling, A.~Cangelosi, A.~Lotfi, and X.~Liu, ``On the gap between
  domestic robotic applications and computational intelligence,''
  \emph{Electronics}, vol.~10, no.~7, 2021. [Online]. Available:
  \url{https://www.mdpi.com/2079-9292/10/7/793}
\BIBentrySTDinterwordspacing

\bibitem{40yearsOfCog}
\BIBentryALTinterwordspacing
I.~Kotseruba and J.~K. Tsotsos, ``40 years of cognitive architectures: core
  cognitive abilities and practical applications,'' \emph{Artificial
  Intelligence Review}, vol.~53, no.~1, pp. 17--94, 2020. [Online]. Available:
  \url{https://doi.org/10.1007/s10462-018-9646-y}
\BIBentrySTDinterwordspacing

\bibitem{LAIRD1991113}
\BIBentryALTinterwordspacing
J.~E. Laird, E.~S. Yager, M.~Hucka, and C.~M. Tuck, ``Robo-soar: An integration
  of external interaction, planning, and learning using soar,'' \emph{Robotics
  and Autonomous Systems}, vol.~8, no.~1, pp. 113--129, 1991, special Issue
  Toward Learning Robots. [Online]. Available:
  \url{https://www.sciencedirect.com/science/article/pii/092188909190017F}
\BIBentrySTDinterwordspacing

\bibitem{CARACaS}
T.~Huntsberger, ``Envisioning cognitive robots for future space exploration,''
  \emph{Proceedings of SPIE - The International Society for Optical
  Engineering}, vol. 7710, 04 2010.

\bibitem{RoboCog}
P.~Bustos, J.~Martínez-Gómez, I.~García-Varea, L.~Rodríguez-Ruiz,
  P.~Bachiller, L.~Calderita, L.~Manso, A.~Sánchez, A.~Bandera, and
  J.~Bandera, ``Multimodal interaction with loki,'' in \emph{Proceedings of
  Workshop of Physical Agents}, 09 2013.

\bibitem{6812892}
P.~Domingos and D.~Lowd, \emph{Markov Logic: An Interface Layer for Artificial
  Intelligence}, 1st~ed.\hskip 1em plus 0.5em minus 0.4em\relax Morgan and
  Claypool Publishers, 2009.

\bibitem{Laird_Lebiere_Rosenbloom_2017}
\BIBentryALTinterwordspacing
J.~E. Laird, C.~Lebiere, and P.~S. Rosenbloom, ``A standard model of the mind:
  Toward a common computational framework across artificial intelligence,
  cognitive science, neuroscience, and robotics,'' \emph{AI Magazine}, vol.~38,
  no.~4, pp. 13--26, Dec. 2017. [Online]. Available:
  \url{https://ojs.aaai.org/index.php/aimagazine/article/view/2744}
\BIBentrySTDinterwordspacing

\bibitem{ipHourglass}
{Steve Deering}, ``Watching the waist of the protocol hourglass,'' Keynote at
  ICNP '98
  \textsc{url:}~\url{https://ant.isi.edu/csci551/images/3/32/Deering98a.pdf},
  10 1998.

\bibitem{910572}
F.~Kschischang, B.~Frey, and H.-A. Loeliger, ``Factor graphs and the
  sum-product algorithm,'' \emph{IEEE Transactions on Information Theory},
  vol.~47, no.~2, pp. 498--519, 2001.

\bibitem{vandemeent2018introduction}
J.-W. van~de Meent, B.~Paige, H.~Yang, and F.~Wood, ``An introduction to
  probabilistic programming,'' 2018.

\bibitem{SimonH.A1956Rcat}
H.~A. Simon, ``\BIBforeignlanguage{eng}{Rational choice and the structure of
  the environment},'' \emph{\BIBforeignlanguage{eng}{Psychological review}},
  vol.~63, no.~2, pp. 129--138, 1956.

\bibitem{6696672}
M.~Fadlil, K.~Ikeda, K.~Abe, T.~Nakamura, and T.~Nagai, ``Integrated concept of
  objects and human motions based on multi-layered multimodal lda,'' in
  \emph{2013 IEEE/RSJ International Conference on Intelligent Robots and
  Systems}, 2013, pp. 2256--2263.

\bibitem{Taniguchi2020}
\BIBentryALTinterwordspacing
A.~Taniguchi, Y.~Hagiwara, T.~Taniguchi, and T.~Inamura, ``Improved and
  scalable online learning of spatial concepts and language models with
  mapping,'' \emph{Autonomous Robots}, July 2020. [Online]. Available:
  \url{https://doi.org/10.1007/s10514-020-09905-0}
\BIBentrySTDinterwordspacing

\bibitem{doi:10.1080/01691864.2020.1817777}
\BIBentryALTinterwordspacing
------, ``Spatial concept-based navigation with human speech instructions via
  probabilistic inference on bayesian generative model,'' \emph{Advanced
  Robotics}, vol.~34, no.~19, pp. 1213--1228, 2020. [Online]. Available:
  \url{https://doi.org/10.1080/01691864.2020.1817777}
\BIBentrySTDinterwordspacing

\bibitem{10.3389/frobt.2019.00131}
\BIBentryALTinterwordspacing
K.~Miyazawa, T.~Horii, T.~Aoki, and T.~Nagai, ``Integrated cognitive
  architecture for robot learning of action and language,'' \emph{Frontiers in
  Robotics and AI}, vol.~6, 2019. [Online]. Available:
  \url{https://www.frontiersin.org/article/10.3389/frobt.2019.00131}
\BIBentrySTDinterwordspacing

\bibitem{10.3389/fnbot.2018.00025}
\BIBentryALTinterwordspacing
T.~Nakamura, T.~Nagai, and T.~Taniguchi, ``Serket: An architecture for
  connecting stochastic models to realize a large-scale cognitive model,''
  \emph{Frontiers in Neurorobotics}, vol.~12, 2018. [Online]. Available:
  \url{https://www.frontiersin.org/article/10.3389/fnbot.2018.00025}
\BIBentrySTDinterwordspacing

\bibitem{10.1007/s00354-019-00084-w}
\BIBentryALTinterwordspacing
T.~Taniguchi, T.~Nakamura, M.~Suzuki, R.~Kuniyasu, K.~Hayashi, A.~Taniguchi,
  T.~Horii, and T.~Nagai, ``Neuro-serket: Development of integrative cognitive
  system through the composition of deep probabilistic generative models,''
  \emph{New Gen. Comput.}, vol.~38, no.~1, p. 23–48, mar 2020. [Online].
  Available: \url{https://doi.org/10.1007/s00354-019-00084-w}
\BIBentrySTDinterwordspacing

\bibitem{10.1145/2593882.2593900}
\BIBentryALTinterwordspacing
A.~D. Gordon, T.~A. Henzinger, A.~V. Nori, and S.~K. Rajamani, ``Probabilistic
  programming,'' in \emph{Future of Software Engineering Proceedings}, ser.
  FOSE 2014.\hskip 1em plus 0.5em minus 0.4em\relax New York, NY, USA:
  Association for Computing Machinery, 2014, p. 167–181. [Online]. Available:
  \url{https://doi.org/10.1145/2593882.2593900}
\BIBentrySTDinterwordspacing

\bibitem{10.5555/1795555}
D.~Koller and N.~Friedman, \emph{Probabilistic Graphical Models: Principles and
  Techniques - Adaptive Computation and Machine Learning}.\hskip 1em plus 0.5em
  minus 0.4em\relax The MIT Press, 2009.

\bibitem{1638022}
H.~Durrant-Whyte and T.~Bailey, ``Simultaneous localization and mapping: part
  i,'' \emph{IEEE Robotics Automation Magazine}, vol.~13, no.~2, pp. 99--110,
  2006.

\bibitem{DBLP:journals/corr/abs-1805-00909}
\BIBentryALTinterwordspacing
S.~Levine, ``Reinforcement learning and control as probabilistic inference:
  Tutorial and review,'' \emph{CoRR}, vol. abs/1805.00909, 2018. [Online].
  Available: \url{http://arxiv.org/abs/1805.00909}
\BIBentrySTDinterwordspacing

\bibitem{8588399}
C.~{Zhang}, J.~{Bütepage}, H.~{Kjellström}, and S.~{Mandt}, ``Advances in
  variational inference,'' \emph{IEEE Transactions on Pattern Analysis and
  Machine Intelligence}, vol.~41, no.~8, pp. 2008--2026, 2019.

\bibitem{damgaard2022SVIFDPR}
\BIBentryALTinterwordspacing
M.~R. Damgaard, R.~Pedersen, and T.~Bak, ``Study of variational inference for
  flexible distributed probabilistic robotics,'' \textit{Preprints} 2022020053,
  2022. [Online]. Available:
  \url{https://doi.org/10.20944/preprints202202.0053.v1}
\BIBentrySTDinterwordspacing

\bibitem{minka2005divergence}
\BIBentryALTinterwordspacing
T.~Minka, ``Divergence measures and message passing,'' Microsoft, Tech. Rep.
  MSR-TR-2005-173, January 2005. [Online]. Available:
  \url{https://www.microsoft.com/en-us/research/publication/divergence-measures-and-message-passing/}
\BIBentrySTDinterwordspacing

\bibitem{JMLR:v14:hoffman13a}
\BIBentryALTinterwordspacing
M.~D. Hoffman, D.~M. Blei, C.~Wang, and J.~Paisley, ``Stochastic variational
  inference,'' \emph{Journal of Machine Learning Research}, vol.~14, no.~4, pp.
  1303--1347, 2013. [Online]. Available:
  \url{http://jmlr.org/papers/v14/hoffman13a.html}
\BIBentrySTDinterwordspacing

\bibitem{pmlr-v33-ranganath14}
\BIBentryALTinterwordspacing
R.~Ranganath, S.~Gerrish, and D.~Blei, ``{Black Box Variational Inference},''
  in \emph{Proceedings of the Seventeenth International Conference on
  Artificial Intelligence and Statistics}, ser. Proceedings of Machine Learning
  Research, S.~Kaski and J.~Corander, Eds., vol.~33.\hskip 1em plus 0.5em minus
  0.4em\relax Reykjavik, Iceland: PMLR, 22--25 Apr 2014, pp. 814--822.
  [Online]. Available: \url{http://proceedings.mlr.press/v33/ranganath14.html}
\BIBentrySTDinterwordspacing

\bibitem{JMLR:v18:16-107}
\BIBentryALTinterwordspacing
A.~Kucukelbir, D.~Tran, R.~Ranganath, A.~Gelman, and D.~M. Blei, ``Automatic
  differentiation variational inference,'' \emph{Journal of Machine Learning
  Research}, vol.~18, no.~14, pp. 1--45, 2017. [Online]. Available:
  \url{http://jmlr.org/papers/v18/16-107.html}
\BIBentrySTDinterwordspacing

\bibitem{NIPS2016_7750ca35}
\BIBentryALTinterwordspacing
Y.~Li and R.~E. Turner, ``R\'{e}nyi divergence variational inference,'' in
  \emph{Advances in Neural Information Processing Systems}, D.~Lee,
  M.~Sugiyama, U.~Luxburg, I.~Guyon, and R.~Garnett, Eds., vol.~29.\hskip 1em
  plus 0.5em minus 0.4em\relax Curran Associates, Inc., 2016. [Online].
  Available:
  \url{https://proceedings.neurips.cc/paper/2016/file/7750ca3559e5b8e1f44210283368fc16-Paper.pdf}
\BIBentrySTDinterwordspacing

\bibitem{DBLP:journals/corr/KingmaW13}
\BIBentryALTinterwordspacing
D.~P. Kingma and M.~Welling, ``Auto-encoding variational bayes,'' in \emph{2nd
  International Conference on Learning Representations, {ICLR} 2014, Banff, AB,
  Canada, April 14-16, 2014, Conference Track Proceedings}, Y.~Bengio and
  Y.~LeCun, Eds., 2014. [Online]. Available:
  \url{http://arxiv.org/abs/1312.6114}
\BIBentrySTDinterwordspacing

\bibitem{Reinforce}
\BIBentryALTinterwordspacing
R.~J. Williams, ``Simple statistical gradient-following algorithms for
  connectionist reinforcement learning,'' \emph{Machine Learning}, vol.~8,
  no.~3, pp. 229--256, 1992. [Online]. Available:
  \url{https://doi.org/10.1007/BF00992696}
\BIBentrySTDinterwordspacing

\bibitem{Salimans_2013}
\BIBentryALTinterwordspacing
T.~Salimans and D.~A. Knowles, ``Fixed-form variational posterior approximation
  through stochastic linear regression,'' \emph{Bayesian Analysis}, vol.~8,
  no.~4, p. 837–882, Dec 2013. [Online]. Available:
  \url{http://dx.doi.org/10.1214/13-BA858}
\BIBentrySTDinterwordspacing

\bibitem{pmlr-v32-rezende14}
D.~J. Rezende, S.~Mohamed, and D.~Wierstra, ``Stochastic backpropagation and
  approximate inference in deep generative models,'' in \emph{Proceedings of
  the 31st International Conference on International Conference on Machine
  Learning - Volume 32}, ser. ICML'14.\hskip 1em plus 0.5em minus 0.4em\relax
  JMLR.org, 2014, p. II–1278–II–1286.

\bibitem{bingham2019pyro}
\BIBentryALTinterwordspacing
E.~Bingham, J.~P. Chen, M.~Jankowiak, F.~Obermeyer, N.~Pradhan, T.~Karaletsos,
  R.~Singh, P.~A. Szerlip, P.~Horsfall, and N.~D. Goodman, ``Pyro: Deep
  universal probabilistic programming,'' \emph{J. Mach. Learn. Res.}, vol.~20,
  pp. 28:1--28:6, 2019. [Online]. Available:
  \url{http://jmlr.org/papers/v20/18-403.html}
\BIBentrySTDinterwordspacing

\bibitem{pmlr-v89-zhang19c}
\BIBentryALTinterwordspacing
J.~Zhang, P.~Raman, S.~Ji, H.-F. Yu, S.~Vishwanathan, and I.~Dhillon, ``Extreme
  stochastic variational inference: Distributed inference for large scale
  mixture models,'' in \emph{Proceedings of the Twenty-Second International
  Conference on Artificial Intelligence and Statistics}, ser. Proceedings of
  Machine Learning Research, K.~Chaudhuri and M.~Sugiyama, Eds., vol.~89.\hskip
  1em plus 0.5em minus 0.4em\relax PMLR, 16--18 Apr 2019, pp. 935--943.
  [Online]. Available: \url{https://proceedings.mlr.press/v89/zhang19c.html}
\BIBentrySTDinterwordspacing

\bibitem{10.5555/2851115}
C.~Davidson-Pilon, \emph{Bayesian Methods for Hackers: Probabilistic
  Programming and Bayesian Inference}, 1st~ed.\hskip 1em plus 0.5em minus
  0.4em\relax Addison-Wesley Professional, 2015.

\bibitem{10.5555/3033232}
A.~Pfeffer, \emph{Practical Probabilistic Programming}, 1st~ed.\hskip 1em plus
  0.5em minus 0.4em\relax USA: Manning Publications Co., 2016.

\bibitem{JSSv076i01}
\BIBentryALTinterwordspacing
B.~Carpenter, A.~Gelman, M.~D. Hoffman, D.~Lee, B.~Goodrich, M.~Betancourt,
  M.~Brubaker, J.~Guo, P.~Li, and A.~Riddell, ``Stan: A probabilistic
  programming language,'' \emph{Journal of Statistical Software}, vol.~76,
  no.~1, p. 1–32, 2017. [Online]. Available:
  \url{https://www.jstatsoft.org/index.php/jss/article/view/v076i01}
\BIBentrySTDinterwordspacing

\bibitem{mansinghka2014venture}
V.~Mansinghka, D.~Selsam, and Y.~Perov, ``Venture: a higher-order probabilistic
  programming platform with programmable inference,'' 2014.

\bibitem{10.1145/3192366.3192409}
\BIBentryALTinterwordspacing
V.~K. Mansinghka, U.~Schaechtle, S.~Handa, A.~Radul, Y.~Chen, and M.~Rinard,
  ``Probabilistic programming with programmable inference,'' in
  \emph{Proceedings of the 39th ACM SIGPLAN Conference on Programming Language
  Design and Implementation}, ser. PLDI 2018.\hskip 1em plus 0.5em minus
  0.4em\relax New York, NY, USA: Association for Computing Machinery, 2018, p.
  603–616. [Online]. Available: \url{https://doi.org/10.1145/3192366.3192409}
\BIBentrySTDinterwordspacing

\bibitem{NEURIPS2019_9015}
\BIBentryALTinterwordspacing
A.~Paszke, S.~Gross, F.~Massa, A.~Lerer, J.~Bradbury, G.~Chanan, T.~Killeen,
  Z.~Lin, N.~Gimelshein, L.~Antiga, A.~Desmaison, A.~Kopf, E.~Yang, Z.~DeVito,
  M.~Raison, A.~Tejani, S.~Chilamkurthy, B.~Steiner, L.~Fang, J.~Bai, and
  S.~Chintala, ``Pytorch: An imperative style, high-performance deep learning
  library,'' in \emph{Advances in Neural Information Processing Systems 32},
  H.~Wallach, H.~Larochelle, A.~Beygelzimer, F.~d\textquotesingle
  Alch\'{e}-Buc, E.~Fox, and R.~Garnett, Eds.\hskip 1em plus 0.5em minus
  0.4em\relax Curran Associates, Inc., 2019, pp. 8024--8035. [Online].
  Available:
  \url{http://papers.neurips.cc/paper/9015-pytorch-an-imperative-style-high-performance-deep-learning-library.pdf}
\BIBentrySTDinterwordspacing

\bibitem{tensorflow2015-whitepaper}
\BIBentryALTinterwordspacing
M.~Abadi, A.~Agarwal, P.~Barham, E.~Brevdo, Z.~Chen, C.~Citro, G.~S. Corrado,
  A.~Davis, J.~Dean, M.~Devin, S.~Ghemawat, I.~Goodfellow, A.~Harp, G.~Irving,
  M.~Isard, Y.~Jia, R.~Jozefowicz, L.~Kaiser, M.~Kudlur, J.~Levenberg,
  D.~Man\'{e}, R.~Monga, S.~Moore, D.~Murray, C.~Olah, M.~Schuster, J.~Shlens,
  B.~Steiner, I.~Sutskever, K.~Talwar, P.~Tucker, V.~Vanhoucke, V.~Vasudevan,
  F.~Vi\'{e}gas, O.~Vinyals, P.~Warden, M.~Wattenberg, M.~Wicke, Y.~Yu, and
  X.~Zheng, ``{TensorFlow}: Large-scale machine learning on heterogeneous
  systems,'' 2015, software available from tensorflow.org. [Online]. Available:
  \url{https://www.tensorflow.org/}
\BIBentrySTDinterwordspacing

\bibitem{siddharth2017learning}
N.~Siddharth, B.~Paige, J.-W. van~de Meent, A.~Desmaison, N.~D. Goodman,
  P.~Kohli, F.~Wood, and P.~H.~S. Torr, ``Learning disentangled representations
  with semi-supervised deep generative models,'' 2017.

\bibitem{tran2017deep}
D.~Tran, M.~D. Hoffman, R.~A. Saurous, E.~Brevdo, K.~Murphy, and D.~M. Blei,
  ``Deep probabilistic programming,'' 2017.

\bibitem{DBLP:journals/corr/abs-1912-11554}
\BIBentryALTinterwordspacing
D.~Phan, N.~Pradhan, and M.~Jankowiak, ``Composable effects for flexible and
  accelerated probabilistic programming in numpyro,'' \emph{CoRR}, vol.
  abs/1912.11554, 2019. [Online]. Available:
  \url{http://arxiv.org/abs/1912.11554}
\BIBentrySTDinterwordspacing

\bibitem{harris2020array}
\BIBentryALTinterwordspacing
C.~R. Harris, K.~J. Millman, S.~J. van~der Walt, R.~Gommers, P.~Virtanen,
  D.~Cournapeau, E.~Wieser, J.~Taylor, S.~Berg, N.~J. Smith, R.~Kern, M.~Picus,
  S.~Hoyer, M.~H. van Kerkwijk, M.~Brett, A.~Haldane, J.~F. del R{\'{i}}o,
  M.~Wiebe, P.~Peterson, P.~G{\'{e}}rard-Marchant, K.~Sheppard, T.~Reddy,
  W.~Weckesser, H.~Abbasi, C.~Gohlke, and T.~E. Oliphant, ``Array programming
  with {NumPy},'' \emph{Nature}, vol. 585, no. 7825, pp. 357--362, Sept. 2020.
  [Online]. Available: \url{https://doi.org/10.1038/s41586-020-2649-2}
\BIBentrySTDinterwordspacing

\bibitem{47008}
\BIBentryALTinterwordspacing
R.~Frostig, M.~Johnson, and C.~Leary, ``Compiling machine learning programs via
  high-level tracing,'' in \emph{Proceedings of Machine Learning and Systems
  (MLSys) 2018}, 2018. [Online]. Available:
  \url{https://mlsys.org/Conferences/doc/2018/146.pdf}
\BIBentrySTDinterwordspacing

\bibitem{ProbMind}
M.~R. Damgaard, ``{ProbMind},'' \url{https://github.com/damgaardmr/probMind},
  2022.

\bibitem{damgaard2022AKS}
M.~R. Damgaard, R.~Pedersen, and T.~Bak, ``A probabilistic programming idiom
  for active knowledge search,'' 2022.

\end{thebibliography}

\clearpage
\onecolumn
%\appendix\label{sec:appendix}
\section{Supplemental information}\label{sec:appendix}

\begin{table*}
\centering
%\begin{tabular}{m{0.125\textwidth}|m{0.25\textwidth}|m{0.58\textwidth}}
\begin{tabular}{m{0.10\textwidth}|m{0.26\textwidth}|m{0.58\textwidth}}
Symbol & Description & Meaning \\ \hline
\resizebox{0.6\linewidth}{!}{\tikzset{every picture/.style={line width=0.75pt}} %set default line width to 0.75pt        

\begin{tikzpicture}[x=0.75pt,y=0.75pt,yscale=-1,xscale=1]
%uncomment if require: \path (0,59); %set diagram left start at 0, and has height of 59

%Shape: Circle [id:dp9730536550098421] 
\draw   (2,26.67) .. controls (2,12.86) and (13.19,1.67) .. (27,1.67) .. controls (40.81,1.67) and (52,12.86) .. (52,26.67) .. controls (52,40.47) and (40.81,51.67) .. (27,51.67) .. controls (13.19,51.67) and (2,40.47) .. (2,26.67) -- cycle ;

% Text Node
\draw (19,15) node [anchor=north west][inner sep=0.75pt]   [align=left] {$\displaystyle z$};

\end{tikzpicture}} &      Circle without colored background       &  A node symbolizing a probabilistic variable, corresponding to a "\textit{sample}" function or keyword in the probabilistic program. We denote this as a "\textit{Latent Variable Node}".    \\ \hline
\resizebox{0.6\linewidth}{!}{\tikzset{every picture/.style={line width=0.75pt}} %set default line width to 0.75pt        

\begin{tikzpicture}[x=0.75pt,y=0.75pt,yscale=-1,xscale=1]
%uncomment if require: \path (0,59); %set diagram left start at 0, and has height of 59

%Shape: Circle [id:dp9730536550098421] 
\draw  [fill={rgb, 255:red, 155; green, 155; blue, 155 }  ,fill opacity=0.5 ] (2,26.67) .. controls (2,12.86) and (13.19,1.67) .. (27,1.67) .. controls (40.81,1.67) and (52,12.86) .. (52,26.67) .. controls (52,40.47) and (40.81,51.67) .. (27,51.67) .. controls (13.19,51.67) and (2,40.47) .. (2,26.67) -- cycle ;

% Text Node
\draw (19,15) node [anchor=north west][inner sep=0.75pt]   [align=left] {$\displaystyle x$};

\end{tikzpicture}} &   Circle with colored background   &   A node symbolizing an observed probabilistic variable, corresponding to a "\textit{observe}" function or keyword in the probabilistic program. We denote this as an "\textit{Observed Variable Node}".      \\ \hline
\resizebox{0.6\linewidth}{!}{\tikzset{every picture/.style={line width=0.75pt}} %set default line width to 0.75pt        

\begin{tikzpicture}[x=0.75pt,y=0.75pt,yscale=-1,xscale=1]
%uncomment if require: \path (0,59); %set diagram left start at 0, and has height of 59

%Shape: Square [id:dp8159501461296683] 
\draw   (2,1.4) -- (52,1.4) -- (52,51.4) -- (2,51.4) -- cycle ;

% Text Node
\draw (20,16) node [anchor=north west][inner sep=0.75pt]   [align=left] {$\displaystyle \theta $};

\end{tikzpicture}} &      Square without colored background      &     A node symbolizing learn-able parameters in the probabilistic program. That is model parameters that can change at run-time. We denote this as a "\textit{Variable Parameter Node}"     \\ \hline
\resizebox{0.6\linewidth}{!}{\tikzset{every picture/.style={line width=0.75pt}} %set default line width to 0.75pt        

\begin{tikzpicture}[x=0.75pt,y=0.75pt,yscale=-1,xscale=1]
%uncomment if require: \path (0,59); %set diagram left start at 0, and has height of 59

%Shape: Square [id:dp8159501461296683] 
\draw [fill={rgb, 255:red, 155; green, 155; blue, 155 }  ,fill opacity=0.5 ]  (2,1.4) -- (52,1.4) -- (52,51.4) -- (2,51.4) -- cycle ;

% Text Node
\draw (20,16) node [anchor=north west][inner sep=0.75pt]   [align=left] {$\displaystyle \breve{\theta } $};

\end{tikzpicture}} &      Square with colored background      &  A node symbolizing fixed parameters in the probabilistic program. Use-full for representing parameters that cannot change at run-time such as tuning parameters. We denote this as a "\textit{Fixed parameter node}"     \\ \hline
\resizebox{0.8\linewidth}{!}{\tikzset{every picture/.style={line width=0.75pt}} %set default line width to 0.75pt        

\begin{tikzpicture}[x=0.75pt,y=0.75pt,yscale=-1,xscale=1]
%uncomment if require: \path (0,24); %set diagram left start at 0, and has height of 24

%Straight Lines [id:da3170322102246792] 
\draw    (10,10.83) -- (66.33,10.52) ;
\draw [shift={(69.33,10.5)}, rotate = 539.6800000000001] [fill={rgb, 255:red, 0; green, 0; blue, 0 }  ][line width=0.08]  [draw opacity=0] (8.93,-4.29) -- (0,0) -- (8.93,4.29) -- cycle    ;

\end{tikzpicture}} &  simple arrow  &   Link showing the generative path in a Probabilistic Program. The arrow can start in Latent Variable Nodes and Parameter nodes, and only point towards \textit{latent variable nodes} and \textit{observed variable nodes}. In large graphs or in cases where the origin of a link can be uncertain the readability can be improved by adding the name of the node from which the link originates next to the link. We denote this as a "\textit{Generative Link}"      \\ \hline
\resizebox{0.8\linewidth}{!}{\tikzset{every picture/.style={line width=0.75pt}} %set default line width to 0.75pt        

\begin{tikzpicture}[x=0.75pt,y=0.75pt,yscale=-1,xscale=1]
%uncomment if require: \path (0,24); %set diagram left start at 0, and has height of 24

%Straight Lines [id:da38218043773704213] 
\draw    (10,10.83) -- (39.67,10.67) ;
%Straight Lines [id:da8082493005015137] 
\draw    (39.67,10.67) -- (66.33,10.52) ;
\draw [shift={(69.33,10.5)}, rotate = 539.6800000000001] [fill={rgb, 255:red, 0; green, 0; blue, 0 }  ][line width=0.08]  [draw opacity=0] (8.93,-4.29) -- (0,0) -- (8.93,4.29) -- cycle    ;
\draw [shift={(39.67,10.67)}, rotate = 359.68] [color={rgb, 255:red, 0; green, 0; blue, 0 }  ][line width=0.75]      (5.59,-5.59) .. controls (2.5,-5.59) and (0,-3.09) .. (0,0) .. controls (0,3.09) and (2.5,5.59) .. (5.59,5.59) ;

\end{tikzpicture}} &  Half circle on arrow  &    Symbolizing that operations downstream of this link in the probabilistic program should not influence nodes upstream to this link in the generative path. I.e. information such as accumulated gradients in a backward pass of automatic differentiation should not be propagated back through this link, corresponding to a ".detach()" and ".stop\_gradient" call in PyTorch and TensorFlow, respectively. We denote this as a "\textit{Detached Link}". A "variable parameter node" having only a detached link, can be considered a "fixed parameter node" or vice versa.  \\ \hline
\resizebox{0.8\linewidth}{!}{\tikzset{every picture/.style={line width=0.75pt}} %set default line width to 0.75pt        

\begin{tikzpicture}[x=0.75pt,y=0.75pt,yscale=-1,xscale=1]
%uncomment if require: \path (0,35); %set diagram left start at 0, and has height of 35

%Straight Lines [id:da38218043773704213] 
\draw  [dash pattern={on 4.5pt off 4.5pt}]  (10,10.83) -- (39.67,10.67) ;
%Straight Lines [id:da2230120618703899] 
\draw    (10,10.83) -- (36.67,10.67) ;
\draw [shift={(36.67,10.67)}, rotate = 359.64] [color={rgb, 255:red, 0; green, 0; blue, 0 }  ][fill={rgb, 255:red, 0; green, 0; blue, 0 }  ][line width=0.75]      (0, 0) circle [x radius= 3.35, y radius= 3.35]   ;
%Straight Lines [id:da44203736758286816] 
\draw    (39.67,10.67) -- (66.33,10.52) ;
\draw [shift={(69.33,10.5)}, rotate = 539.6800000000001] [fill={rgb, 255:red, 0; green, 0; blue, 0 }  ][line width=0.08]  [draw opacity=0] (8.93,-4.29) -- (0,0) -- (8.93,4.29) -- cycle    ;
%Straight Lines [id:da6188940529579046] 
\draw    (36.5,30.88) -- (36.64,13.67) ;
\draw [shift={(36.67,10.67)}, rotate = 450.47] [fill={rgb, 255:red, 0; green, 0; blue, 0 }  ][line width=0.08]  [draw opacity=0] (8.93,-4.29) -- (0,0) -- (8.93,4.29) -- cycle    ;

\end{tikzpicture}} &  Simple arrow pointing towards another simple arrow  &   Depending on the context, samples from different \textit{latent variable nodes} may be used to generate the next sample. The generative link from the most recent sampled \textit{latent variable node} in the generative path is the new active link. Used, e.g., to represent for loops. \\ \hline
\resizebox{1.0\linewidth}{!}{\tikzset{every picture/.style={line width=0.75pt}} %set default line width to 0.75pt        

\begin{tikzpicture}[x=0.75pt,y=0.75pt,yscale=-1,xscale=1]
%uncomment if require: \path (0,59); %set diagram left start at 0, and has height of 59

%Shape: Rectangle [id:dp6845645781697427] 
\draw   (1.78,6.89) .. controls (1.78,4.13) and (4.02,1.89) .. (6.78,1.89) -- (64.33,1.89) .. controls (67.09,1.89) and (69.33,4.13) .. (69.33,6.89) -- (69.33,46.78) .. controls (69.33,49.54) and (67.09,51.78) .. (64.33,51.78) -- (6.78,51.78) .. controls (4.02,51.78) and (1.78,49.54) .. (1.78,46.78) -- cycle ;

% Text Node
\draw (3.57,16.14) node [anchor=north west][inner sep=0.75pt]   [align=left] {$\displaystyle \{z ; x_1, x_2; \theta \}$};

\end{tikzpicture}} &      Polygon with rounded corners     &   Collection of nodes with internal dependency structure. The full structure of links between nodes can be shown inside the polygon. Alternatively, the names of the nodes can be written between curly brackets, $\{z;x;\theta\}$, with a semi-colon separating variables nodes, $z$, observed nodes, $x$, and parameter nodes, $\theta$, in that order. Finally, a node collection can also simply be defined somewhere else, $C = \{z;x;\theta\}$, and referred to by, e.g., a single letter, in which case we encourage the use of capital letters to emphasize the difference from the other types of nodes. In cases where one or more types of nodes are not present in a collection, both semi-colons should still be there. E.g., $\{;x_1,x_2;\}$ for a collection the two observed nodes $x_1$ and $x_2$. Such a collection of nodes directly corresponds to the factor $p_\theta(x_1,x_2|z)$ in a factorization of the joint distribution over all variables in a model. We denote this as a "\textit{Node Collection}". A node is only allowed to be within one node collection unless it is within a node collection fully nested within another. \\ \hline
\resizebox{1.0\linewidth}{!}{\tikzset{every picture/.style={line width=0.75pt}} %set default line width to 0.75pt        

\begin{tikzpicture}[x=0.75pt,y=0.75pt,yscale=-1,xscale=1]
%uncomment if require: \path (0,59); %set diagram left start at 0, and has height of 59

%Shape: Rectangle [id:dp6845645781697427] 
\draw   (1.78,6.89) .. controls (1.78,4.13) and (4.02,1.89) .. (6.78,1.89) -- (83.89,1.89) .. controls (86.65,1.89) and (88.89,4.13) .. (88.89,6.89) -- (88.89,46.78) .. controls (88.89,49.54) and (86.65,51.78) .. (83.89,51.78) -- (6.78,51.78) .. controls (4.02,51.78) and (1.78,49.54) .. (1.78,46.78) -- cycle ;

% Text Node
\draw (3.57,16.14) node [anchor=north west][inner sep=0.75pt]   [align=left] {$\displaystyle \{z^{(n)} ;x^{(n)} ;\theta^{(n)}\}$};
% Text Node
\draw (50.22,33.89) node [anchor=north west][inner sep=0.75pt]   [align=left] {$n\in N$};

\end{tikzpicture}} &    Index specification in one corner of a polygon with rounded corners       &   Collection of nodes with indexed names. When the index is used to name nodes it is good practice to use the index in a superscript and encapsulate it in round brackets to emphasize that it is an index. We denote such a node collection an "\textit{Indexed Node Collection}". The valid index should be clear from the context. E.g., in the case of a loop around the indexed collection, the index is incremented each time the loop enters the indexed node collection. When no such loops exist around the indexed node collection that could potentially cause ambiguity, it can be used instead of writing all the variables in a node collection with the curly brackets. For an example see \cref{fig:slam_graph_idiom_new_3}\\ \hline
\resizebox{1.0\linewidth}{!}{\tikzset{every picture/.style={line width=0.75pt}} %set default line width to 0.75pt        

\begin{tikzpicture}[x=0.75pt,y=0.75pt,yscale=-1,xscale=1]
%uncomment if require: \path (0,59); %set diagram left start at 0, and has height of 59

%Shape: Rectangle [id:dp006649263477550127] 
\draw   (4.78,10.11) .. controls (4.78,7.35) and (7.02,5.11) .. (9.78,5.11) -- (86.89,5.11) .. controls (89.65,5.11) and (91.89,7.35) .. (91.89,10.11) -- (91.89,50) .. controls (91.89,52.76) and (89.65,55) .. (86.89,55) -- (9.78,55) .. controls (7.02,55) and (4.78,52.76) .. (4.78,50) -- cycle ;
%Shape: Rectangle [id:dp6845645781697427] 
\draw  [fill={rgb, 255:red, 255; green, 255; blue, 255 }  ,fill opacity=1 ] (1.78,6.89) .. controls (1.78,4.13) and (4.02,1.89) .. (6.78,1.89) -- (83.89,1.89) .. controls (86.65,1.89) and (88.89,4.13) .. (88.89,6.89) -- (88.89,46.78) .. controls (88.89,49.54) and (86.65,51.78) .. (83.89,51.78) -- (6.78,51.78) .. controls (4.02,51.78) and (1.78,49.54) .. (1.78,46.78) -- cycle ;

% Text Node
\draw (3.57,16.14) node [anchor=north west][inner sep=0.75pt]   [align=left] {$\displaystyle \{z^{(n)} ;x^{(n)} ;\theta^{(n)} \}$};
% Text Node
\draw (50.0,33.89) node [anchor=north west][inner sep=0.75pt]   [align=left] {$n\in N$};

\end{tikzpicture}} &    Stacked polygons with rounded corners   &  Explicitly representation of multiple identical collections of nodes conditionally independent given their parents or simply independent if there are no parents. Here the one index is used for each of the independent collections.    \\ \hline
\resizebox{0.8\linewidth}{!}{\tikzset{every picture/.style={line width=0.75pt}} %set default line width to 0.75pt        

\begin{tikzpicture}[x=0.75pt,y=0.75pt,yscale=-1,xscale=1]
%uncomment if require: \path (0,24); %set diagram left start at 0, and has height of 24

%Straight Lines [id:da38218043773704213] 
\draw  [dash pattern={on 4.5pt off 4.5pt}]  (10,10.83) -- (39.67,10.67) ;
%Straight Lines [id:da490255891022759] 
\draw  [dash pattern={on 4.5pt off 4.5pt}]  (10,10.83) -- (66.33,10.52) ;
\draw [shift={(69.33,10.5)}, rotate = 539.6800000000001] [fill={rgb, 255:red, 0; green, 0; blue, 0 }  ][line width=0.08]  [draw opacity=0] (8.93,-4.29) -- (0,0) -- (8.93,4.29) -- cycle    ;

\end{tikzpicture}} &  Dashed arrow  &    A link symbolizing an indirect relation between nodes. Such a link is non-generative, meaning that it is not directly used as a parameter in the generation of other samples, but only influences the generative path of other samples. We denote this as an "\textit{Influence Link}"     \\ \hline
\resizebox{0.95\linewidth}{!}{\tikzset{every picture/.style={line width=0.75pt}} %set default line width to 0.75pt        

\begin{tikzpicture}[x=0.75pt,y=0.75pt,yscale=-1,xscale=1]
%uncomment if require: \path (0,92); %set diagram left start at 0, and has height of 92

%Straight Lines [id:da6332824380891224] 
\draw    (1,35.57) -- (16.77,35.57) ;
\draw [shift={(19.77,35.57)}, rotate = 180] [fill={rgb, 255:red, 0; green, 0; blue, 0 }  ][line width=0.08]  [draw opacity=0] (8.93,-4.29) -- (0,0) -- (8.93,4.29) -- cycle    ;
%Shape: Diamond [id:dp7934791889858492] 
\draw   (89.77,35.57) -- (54.77,70.57) -- (19.77,35.57) -- (54.77,0.57) -- cycle ;
%Straight Lines [id:da2792915393577655] 
\draw    (89.77,35.57) -- (105.54,35.57) ;
\draw [shift={(108.54,35.57)}, rotate = 180] [fill={rgb, 255:red, 0; green, 0; blue, 0 }  ][line width=0.08]  [draw opacity=0] (8.93,-4.29) -- (0,0) -- (8.93,4.29) -- cycle    ;
%Straight Lines [id:da17766998382413446] 
\draw    (54.77,70.57) -- (54.84,85.86) ;
\draw [shift={(54.86,88.86)}, rotate = 269.73] [fill={rgb, 255:red, 0; green, 0; blue, 0 }  ][line width=0.08]  [draw opacity=0] (8.93,-4.29) -- (0,0) -- (8.93,4.29) -- cycle    ;
%Straight Lines [id:da5302927154180501] 
\draw  [dash pattern={on 4.5pt off 4.5pt}]  (4.71,85.14) -- (35.53,56.06) ;
\draw [shift={(37.71,54)}, rotate = 496.66] [fill={rgb, 255:red, 0; green, 0; blue, 0 }  ][line width=0.08]  [draw opacity=0] (8.93,-4.29) -- (0,0) -- (8.93,4.29) -- cycle    ;

% Text Node
\draw (23,26.14) node [anchor=north west][inner sep=0.75pt]   [align=left] {condition};
% Text Node
\draw (79.71,11.14) node [anchor=north west][inner sep=0.75pt]   [align=left] {True};
% Text Node
\draw (61.14,67.29) node [anchor=north west][inner sep=0.75pt]   [align=left] {False};

\end{tikzpicture}} &    A Polygon with generative links connected at vertices, influence links towards the polygon connected at edges, and conditional values nearby generative links pointing away from the polygon  &  Node representing a condition changing the "direction" of the generative flow in a probabilistic program. We denote this as "\textit{conditioned generative branching}".    \\ \hline
\resizebox{0.95\linewidth}{!}{\tikzset{every picture/.style={line width=0.75pt}} %set default line width to 0.75pt        

\begin{tikzpicture}[x=0.75pt,y=0.75pt,yscale=-1,xscale=1]
%uncomment if require: \path (0,92); %set diagram left start at 0, and has height of 92

%Straight Lines [id:da7691882454643624] 
\draw    (1,54.57) -- (16.77,54.57) ;
\draw [shift={(19.77,54.57)}, rotate = 180] [fill={rgb, 255:red, 0; green, 0; blue, 0 }  ][line width=0.08]  [draw opacity=0] (8.93,-4.29) -- (0,0) -- (8.93,4.29) -- cycle    ;
%Shape: Diamond [id:dp9495480542104542] 
\draw   (89.77,54.57) -- (54.77,89.57) -- (19.77,54.57) -- (54.77,19.57) -- cycle ;
%Straight Lines [id:da8999599002414695] 
\draw    (89.77,54.57) -- (105.54,54.57) ;
\draw [shift={(108.54,54.57)}, rotate = 180] [fill={rgb, 255:red, 0; green, 0; blue, 0 }  ][line width=0.08]  [draw opacity=0] (8.93,-4.29) -- (0,0) -- (8.93,4.29) -- cycle    ;
%Straight Lines [id:da7310168362849985] 
\draw    (54.69,1.29) -- (54.76,16.57) ;
\draw [shift={(54.77,19.57)}, rotate = 269.73] [fill={rgb, 255:red, 0; green, 0; blue, 0 }  ][line width=0.08]  [draw opacity=0] (8.93,-4.29) -- (0,0) -- (8.93,4.29) -- cycle    ;
%Straight Lines [id:da34432558763156007] 
\draw  [dash pattern={on 4.5pt off 4.5pt}]  (5,5.69) -- (35.15,33.97) ;
\draw [shift={(37.33,36.03)}, rotate = 223.17000000000002] [fill={rgb, 255:red, 0; green, 0; blue, 0 }  ][line width=0.08]  [draw opacity=0] (8.93,-4.29) -- (0,0) -- (8.93,4.29) -- cycle    ;

% Text Node
\draw (23,45.14) node [anchor=north west][inner sep=0.75pt]   [align=left] {condition};
% Text Node
\draw (0.05,62.72) node [anchor=north west][inner sep=0.75pt]   [align=left] {True};
% Text Node
\draw (61.56,2.45) node [anchor=north west][inner sep=0.75pt]   [align=left] {False};

\end{tikzpicture}} &     A Polygon with generative links connected at vertices, influence links towards the polygon connected at edges, and conditional values nearby generative links pointing towards the polygon        & Node representing a condition selecting one out of two or more possible generative flows from parent nodes. We denote this as "\textit{conditioned generative selection}". 
\end{tabular}
\caption{Semantics of the Generative Flow Graph representation of probabilistic programs.}\label{table:generative_flow_graph_semantics}
\end{table*}

\subsection{Rewriting KL-divergence for Stochastic Message-Passing}\label{app:S-msg}
In this section, we will derive the dual objective for the combination of message-passing and stochastic variational inference presented in \cref{sec:SMP}. Start by considering

\begin{align*}
    p_{\Theta } (X_{G} |X^{\{A\}} ) & =\int p_{\Theta }( X_{G} ,Z^{\{A\}} |X^{\{A\}} )dZ\\
    & =\int p(X_{G} |Z^{\{A\}} ,X^{\{A\}} )p_{\Theta } (Z^{\{A\}} |X^{\{A\}} )dZ\\
    & =\int p(X_{G} |Z^{\{A\}} )p_{\Theta } (Z^{\{A\}} |X^{\{A\}} )dZ\\
    & =\int p(X_{G} |Z^{\{A\}} )\prod _{a\in A} p_{\Theta ^{\{a\}} ,\text{Pa}\breve{\Theta }\left( C^{\{a\}}\right)}\left( Z^{\{a\}} |\text{Pa}\breve{Z}\left( C^{\{a\}}\right) ,X^{\{a\}}\right) dZ
\end{align*}

By replacing $p_{\Theta ^{\{b\}} ,\text{Pa}\breve{\Theta }\left( C^{\{b\}}\right)}\left( Z^{\{b\}} |\text{Pa}\breve{Z}\left( C^{\{b\}}\right) ,X^{\{b\}}\right)$ with their corresponding variational distributions $q_{\breve{\Phi }^{\{b\}}}\left( Z^{\{b\}}\right)$ for $b\in A\setminus a$ we obtain 

\begin{align}
\tilde{p}^{\{a\}} (X_{G} |X^{\{a\}} )= & \int p(X_{G} |Z^{\{A\}} )p_{\Theta ^{\{a\}} ,\text{Pa}\breve{\Theta }\left( C^{\{a\}}\right)}\left( Z^{\{a\}} |\text{Pa}\breve{Z}\left( C^{\{a\}}\right) ,X^{\{a\}}\right)\prod _{b\in A\setminus a} q_{\breve{\Phi }^{\{b\}}}\left( Z^{\{b\}}\right) dZ \label{eq:marginal_likelyhood_approx}
\end{align}

where we have used $\tilde{p}^{\{a\}} (X_{G} |X^{\{a\}} )$ instead of $\tilde{p}^{\{a\}} (X_{G} |X^{\{A\}} )$ to emphasize the conditional independence between $X_{G}$ and $X^{\{b\}}$ given $ Z^{\{b\}}$ for $ b\in A\setminus a$ implicitly assumed by the approximation. Furthermore, also notice that we can rewrite the distribution, $p(X_{G} |Z^{\{A\}} )$, as follows

\begin{align*}
p(X_{G} |Z^{\{A\}} ) & =p(\text{Ch} X_{G}\left( Z^{\{a\}}\right) |Z^{\{A\}} )p(X_{G} \setminus \text{Ch} X_{G}\left( Z^{\{a\}}\right) |Z^{\{A\}} )\\
 & =p\left(\text{Ch} X_{G}\left( Z^{\{a\}}\right) |\text{Pa} Z\left(\text{Ch} X_{G}\left( Z^{\{a\}}\right)\right)\right) p\left( X_{G} \setminus \text{Ch} X_{G}\left( Z^{\{a\}}\right) |Z^{\{A\}} \setminus Z^{\{a\}}\right)
\end{align*}

by separating the global observed variables, $X_G$, into those who are direct children of $Z^{\{a\}}$, and those who are not. 
With the definition above and the ones given in \cref{sec:SMP}, we can rewrite the KL-divergence as follows:

\allowdisplaybreaks
\begin{align}
    & D_{KL}\left[ q_{\Phi ^{\{a\}}} (Z)||\tilde{p}_{\Theta ^{\{a\}}}^{\{a\}} (Z|X)\right] \label{eq:msg_svi_kl_rewritten1}\\
    & \quad =\int _{Z} q_{\Phi ^{\{a\}}} (Z)\log\left(\frac{q_{\Phi ^{\{a\}}} (Z)}{\tilde{p}_{\Theta ^{\{a\}}}^{\{a\}} (Z|X)}\right) dZ\\
    & \quad =\int _{Z} q_{\Phi ^{\{a\}}} (Z)\log\left(\frac{q_{\Phi ^{\{a\}}}\left( Z^{\{a\}}\right)\prod _{b\in A\setminus a} q_{\breve{\Phi }^{\{b\}}}\left( Z^{\{b\}}\right)}{\frac{p(X_{G} |Z^{\{A\}} )}{\tilde{p}^{\{a\}} (X_{G} |X^{\{a\}} )} p_{\Theta ^{\{a\}} ,\text{Pa}\breve{\Theta }\left( C^{\{a\}}\right)}\left( Z^{\{a\}} |\text{Pa}\breve{Z}\left( C^{\{a\}}\right) ,X^{\{a\}}\right)\prod _{b\in A\setminus a} q_{\breve{\Phi }^{\{b\}}}\left( Z^{\{b\}}\right)}\right) dZ\\
    & \quad =\int _{Z} q_{\Phi ^{\{a\}}} (Z)\log\left(\frac{q_{\Phi ^{\{a\}}}\left( Z^{\{a\}}\right)}{\frac{p(X_{G} |Z^{\{A\}} )}{\tilde{p}^{\{a\}} (X_{G} |X^{\{a\}} )} p_{\Theta ^{\{a\}} ,\text{Pa}\breve{\Theta }\left( C^{\{a\}}\right)}\left( Z^{\{a\}} |\text{Pa}\breve{Z}\left( C^{\{a\}}\right) ,X^{\{a\}}\right)}\right) dZ\\
    & \quad =\int _{Z} q_{\Phi ^{\{a\}}} (Z)\log\left(\frac{q_{\Phi ^{\{a\}}}\left( Z^{\{a\}}\right)}{\frac{p(X_{G} |Z^{\{A\}} )}{\tilde{p}^{\{a\}} (X_{G} |X^{\{a\}} )}\frac{p_{\Theta ^{\{a\}} ,\text{Pa}\breve{\Theta }\left( C^{\{a\}}\right)}\left( Z^{\{a\}} ,X^{\{a\}} |\text{Pa}\breve{Z}\left( C^{\{a\}}\right)\right)}{p_{\Theta ^{\{a\}} ,\text{Pa}\breve{\Theta }\left( C^{\{a\}}\right)}\left( X^{\{a\}} |\text{Pa}\breve{Z}\left( C^{\{a\}}\right)\right)}}\right) dZ\\
    & \quad =\int _{Z} q_{\Phi ^{\{a\}}} (Z)\log\left(\frac{q_{\Phi ^{\{a\}}}\left( Z^{\{a\}}\right)}{\frac{p\left(\text{Ch} X_{G}\left( Z^{\{a\}}\right) |\text{Pa} Z\left(\text{Ch} X_{G}\left( Z^{\{a\}}\right)\right)\right) p\left( X_{G} \setminus \text{Ch} X_{G}\left( Z^{\{a\}}\right) |Z^{\{A\}} \setminus Z^{\{a\}}\right)}{\tilde{p}^{\{a\}} (X_{G} |X^{\{a\}} )}\frac{p_{\Theta ^{\{a\}} ,\text{Pa}\breve{\Theta }\left( C^{\{a\}}\right)}\left( Z^{\{a\}} ,X^{\{a\}} |\text{Pa}\breve{Z}\left( C^{\{a\}}\right)\right)}{p_{\Theta ^{\{a\}} ,\text{Pa}\breve{\Theta }\left( C^{\{a\}}\right)}\left( X^{\{a\}} |\text{Pa}\breve{Z}\left( C^{\{a\}}\right)\right)}}\right) dZ\\
    & \quad =\int _{Z} q_{\Phi ^{\{a\}}} (Z)\log\left(\frac{q_{\Phi ^{\{a\}}}\left( Z^{\{a\}}\right)}{p\left(\text{Ch} X_{G}\left( Z^{\{a\}}\right) |\text{Pa} Z\left(\text{Ch} X_{G}\left( Z^{\{a\}}\right)\right)\right) p_{\Theta ^{\{a\}} ,\text{Pa}\breve{\Theta }\left( C^{\{a\}}\right)}\left( Z^{\{a\}} ,X^{\{a\}} |\text{Pa}\breve{Z}\left( C^{\{a\}}\right)\right)}\right) dZ \nonumber\\
    & \quad \quad \quad \quad +\int _{Z} q_{\Phi ^{\{a\}}} (Z)\log\left(\tilde{p}^{\{a\}} (X_{G} |X^{\{a\}} )p_{\Theta ^{\{a\}} ,\text{Pa}\breve{\Theta }\left( C^{\{a\}}\right)}\left( X^{\{a\}} |\text{Pa}\breve{Z}\left( C^{\{a\}}\right)\right)\right) dZ\\
    & \quad \quad \quad \quad -\int _{Z} q_{\Phi ^{\{a\}}} (Z)\log\left( p\left( X_{G} \setminus \text{Ch} X_{G}\left( Z^{\{a\}}\right) |Z^{\{A\}} \setminus Z^{\{a\}}\right)\right) dZ \nonumber\\
    & \quad =\underbrace{E_{Z\sim \tilde{q}_{\Phi ^{\{a\}}}^{\{a\}}}\left[\log\left(\frac{q_{\Phi ^{\{a\}}}\left( Z^{\{a\}}\right)}{p\left(\text{Ch} X_{G}\left( Z^{\{a\}}\right) |\text{Pa} Z\left(\text{Ch} X_{G}\left( Z^{\{a\}}\right)\right)\right) p_{\Theta ^{\{a\}} ,\text{Pa}\breve{\Theta }\left( C^{\{a\}}\right)}\left( Z^{\{a\}} ,X^{\{a\}} |\text{Pa}\breve{Z}\left( C^{\{a\}}\right)\right)}\right)\right]}_{-L_{KL}^{\{a\}}\left( \Theta ^{\{a\}} ,\Phi ^{\{a\}}\right)} \nonumber\\
    & \quad \quad \quad \quad +E_{Z\sim \tilde{q}_{\text{Pa}\breve{Z}}^{\{a\}}}\left[\underbrace{\log\left(\tilde{p}^{\{a\}} (X_{G} |X^{\{a\}} )p_{\Theta ^{\{a\}} ,\text{Pa}\breve{\Theta }\left( C^{\{a\}}\right)}\left( X^{\{a\}} |\text{Pa}\breve{Z}\left( C^{\{a\}}\right)\right)\right)}_{LogEvd_{X_{G} ,\ X^{\{a\}}}^{\{a\}}\left( \Theta ^{\{a\}}\right)}\right]  \label{eq:msg_svi_kl_rewritten2}\\
    & \quad \quad \quad \quad -\underbrace{E_{Z\sim \prod _{b\in A\setminus a} q_{\breve{\Phi }^{\{b\}}}\left( Z^{\{b\}}\right)}\left[\log\left( p\left( X_{G} \setminus \text{Ch} X_{G}\left( Z^{\{a\}}\right) |Z^{\{A\}} \setminus Z^{\{a\}}\right)\right)\right]}_{C\ } \nonumber
\end{align}
where
\begin{align*}
\tilde{q}_{\Phi ^{\{a\}}}^{\{a\}} & =q_{\Phi ^{\{a\}}}\left( Z^{\{a\}}\right)\prod _{Z^{\{b\}} \in \text{Pa}\breve{Z}\left( C^{\{a\}}\right) \cup \text{Pa}\breve{Z}\left(\text{Ch} X_{G}\left( Z^{\{a\}}\right)\right) \setminus Z^{\{a\}}} q_{\breve{\Phi }^{\{b\}}}\left( Z^{\{b\}}\right)
\end{align*}

is the joint variational distribution over the latent variables, $Z^{\{a\}}$, local to the a'th node collection, and the latent variables parent to the a'th node collection, $\text{Pa}\breve{Z}\left( C^{\{a\}}\right)$, or having the same child global observed variables as the a'th node collection, $Z^{\{b\}} \in \text{Pa}\breve{Z}\left(\text{Ch} X_{G}\left( Z^{\{a\}}\right)\right) \setminus Z^{\{a\}}$. Furthermore,
\begin{align*}
\tilde{q}_{\text{Pa}\breve{Z}}^{\{a\}} & =\prod _{Z^{\{b\}} \in \text{Pa}\breve{Z}\left( C^{\{a\}}\right)} q_{\breve{\Phi }^{\{b\}}}\left( Z^{\{b\}}\right)
\end{align*}
is the joint variational distribution over the latent variables parent to the a'th node collection, $\text{Pa}\breve{Z}\left( C^{\{a\}}\right)$. Finally, $LogEvd_{X_{G} ,\ X^{\{a\}}}^{\{a\}}\left( \Theta ^{\{a\}}\right)$ denotes the joint log-evidence for $X_{G}$ and $ X^{\{a\}}$, and $C$ is constant with respect to $ \Theta ^{\{a\}} ,\Phi ^{\{a\}} \ $. By simple rearranging terms we obtain the dual objective

\begin{align*}
L_{KL}^{\{a\}}\left( \Theta ^{\{a\}} ,\Phi ^{\{a\}}\right) = & E_{Z\sim \tilde{q}_{\text{Pa}\breve{Z}}^{\{a\}}}\left[ LogEvd_{X_{G} ,\ X^{\{a\}}}^{\{a\}}\left( \Theta ^{\{a\}}\right)\right] -D_{KL}\left[ q_{\Phi ^{\{a\}}} (Z)||\tilde{p}_{\Theta ^{\{a\}}}^{\{a\}} (Z|X)\right] -C
\end{align*}

\end{document}